\documentclass[11pt]{article}

\usepackage[final]{acl}

\usepackage{times}
\usepackage{latexsym}

\usepackage{todonotes}
\usepackage{amsmath}
\usepackage{amssymb}
\usepackage{acro}
\DeclareAcronym{MMD}{
  short = MMD,
  long  = Maximum Mean Discrepancy
}

\DeclareAcronym{LLM}{
  short = LLM,
  long  = Large Language Model
}

\DeclareAcronym{RKHS}{
  short = RKHS,
  long  = Reproducing Kernel Hilbert Space
}

\DeclareAcronym{WP}{
  short = WP,
  long  = Reddit Writing Prompts
}

\DeclareAcronym{CI}{
  short = CI,
  long  = Confidence Interval
}

\DeclareAcronym{MDA}{
  short = MDA,
  long  = Multi-Dimensional Analysis
}

\usepackage{array}
\usepackage{colortbl}
\usepackage{xcolor}

\usepackage{booktabs}
\usepackage{rotating}

\usepackage{graphicx}
\usepackage{subcaption}
\usepackage{placeins}

\usepackage{xcolor}
\usepackage{xparse}

\usepackage{longtable}
\usepackage{multirow}

\newcommand{\loss}[1]{\textsc{#1}}

\usepackage[T1]{fontenc}

\usepackage[utf8]{inputenc}

\usepackage{microtype}

\usepackage{inconsolata}

\usepackage{graphicx}

\title{How Human-Like Are Large Language Models? \\A Register-Aware Linguistic Evaluation Framework}

\author{
  \textbf{Björn Nieth}\textsuperscript{1,4}, 
  \textbf{Marianna Gracheva}\textsuperscript{2}, 
  \textbf{Michaela Mahlberg}\textsuperscript{2,3},\\
  \textbf{Bjoern Eskofier}\textsuperscript{1,4,5,6}, 
  \textbf{Emmanuelle Salin}\textsuperscript{1} \\[0.3em]
  \small \textsuperscript{1}Department Artificial Intelligence in Biomedical Engineering (AIBE), FAU Erlangen-Nürnberg, Germany\\
  \small \textsuperscript{2}The Text and Language Lab, Department of Digital Humanities and Social Studies (DHSS), FAU Erlangen-Nürnberg, Germany\\
  \small \textsuperscript{3}Department of Linguistics and Communication, University of Birmingham, United Kingdom
  \small \textsuperscript{4}Munich Center for Machine Learning (MCML), Germany\\
  \small \textsuperscript{5}Chair of AI-supported Therapy Decisions, LMU München, Germany
  \small \textsuperscript{6}Institute of AI for Health, Helmholtz Zentrum München, Germany\\[0.2em]
  \small \texttt{\{bjoern.nieth, marianna.gracheva, michaela.mahlberg, bjoern.eskofier, emmanuelle.salin\}@fau.de}
}

\begin{document}
\maketitle
\begin{abstract}
While factual correctness and task-performance have received much attention by \ac{LLM} research, the fundamental question of how human-like generated texts are on a linguistic level has been underexplored. From a corpus-linguistic perspective, language production is inherently context-dependent. Distinct communicative contexts give rise to differences in frequencies and co-occurrence patterns of linguistic features that shape registers of language use. A text failing to adhere to these patterns can be correct content-wise, but still may not appear natural to human readers. In this work, we propose a context-aware evaluation framework in which human-likeness is measured as a two-sample problem between the linguistic feature distribution of a human reference corpus for a given register and a corresponding \ac{LLM}-generated corpus. We implement this framework using the \ac{MMD} measure and the 67 lexico-grammatical features introduced by Biber, which are commonly applied in corpus linguistics. In our experiments, we compare seven instruction-tuned, open-source models and look at  five English-language datasets spanning distinct registers. While across all tested setups, \acp{LLM} deviate from the human baseline, which models are closest to human language depends on the register and is not dictated by model size.
\end{abstract}

\section{Introduction}

Research on \acf{LLM} capabilities has largely focused on task- or domain-oriented benchmarks, specialized metrics such as faithfulness (e.g., grounding with respect to a given input or reference \cite{es_ragas_2024}) or downstream task performance. While such metrics are useful to evaluate specific aspects of \ac{LLM} abilities, they fail to address the fundamental issue of how closely \ac{LLM}-generated language resembles human language use. As the use of \acp{LLM} is becoming increasingly pervasive in people's everyday lives, e.g. in the form of personal chatbots,  the proportion of \ac{LLM}-generated text encountered in various media will only increase. Therefore, it is important to be able to evaluate how well these generated texts reproduce the specific linguistic patterns observed in human language.

\begin{figure}[t]
  \centering
  \includegraphics[width=1\columnwidth]{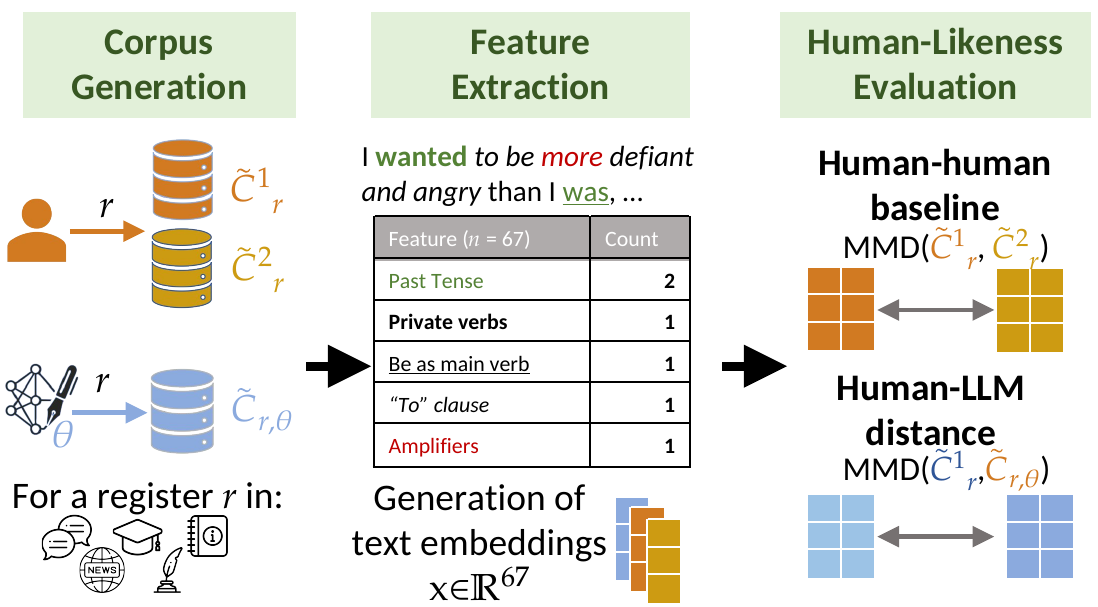}
  \caption{Overview of the proposed evaluation framework. Given a specific register  $r$, we collect a human and LLM-generated corpus. Then we extract a set of 67 linguistic features from the texts. Finally, we calculate the distance between the human sample and the synthetic corpus and resample the distance between human samples as a baseline.}
  \label{fig:method-overview}
\end{figure}

From a linguistic perspective, human language cannot be described in terms of one single model.  Human language production inherently depends on situational factors such as communicative purpose, audience, mode, and production circumstances, among others \cite{biber_register_2019}. These situational factors define a \textit{register}, i.e. variety of language use. Hence, a general analysis of \ac{LLM}-language without attention to register will always disregard more specific information.  The differences between registers (e.g., research articles, news, conversations, etc.) make it difficult to draw general conclusions \cite{douglas_biber_register_2012,veirano_pinto_elena_2023}. Certain features are frequent in a register because they are demanded by a particular situation of use. As such, human language tends to produce common frequencies and co-occurrence patterns of linguistic features for a register. In this paper, we define human-likeness as the ability to produce language that fits the distribution of linguistic features of a target register. An \ac{LLM}-generated text failing to adhere to these register-specific linguistic feature distributions will be perceived as unnatural and affect the human user experience. As a result, the text may miss its  communicative purpose in the context of real-world applications \cite{sardinha_ai-generated_2024}.

Therefore, a growing body of work has focused on analyzing \ac{LLM}-generated texts by examining the linguistic features of these texts. While these studies provide empirical evidence on the differences between AI-generated and human-authored texts, they often fail to explicitly measure register-alignment across different registers and different models. 

To overcome these limitations, we introduce a register-aware evaluation framework focusing on the distributional difference of linguistic features between corpora. An overview of our method is shown in Figure \ref{fig:method-overview}. We publish our code, generation settings and results along with this paper\footnote{Code and results available at: \url{https://github.com/BjoernNieth/Register_Aware_LLMs}}. Our main contributions are: 
\begin{itemize}
    \itemsep 0em 
    \item A method for the assessment of the human-likeness of \ac{LLM}-generated texts based on linguistic features and the \acf{MMD}.
    \item An open-source implementation and analysis of the framework, using the 67 linguistic features introduced in \cite{biber1988variation}. 
    \item A benchmark of five registers with a human baseline and the generation results of seven models in a zero and few-shot setting, together with a further ablation study on prompt stability for one register. 
\end{itemize}

\section{Related Work}
Traditional benchmarks for \ac{LLM} evaluation focus on task performance 
(e.g., MMLU \cite{hendrycks_measuring_2021}) or domain understanding often through multiple choice questions (ChemBench \cite{mirza_framework_2025}, LegalBENCH \cite{guha_legalbench_2023}, MedQA \cite{jinWhatDiseaseDoes2021}) without assessing whether \ac{LLM}-generated texts are appropriate to the underlying register. However, recent work has shown that the performance of an \ac{LLM} on a benchmark is influenced by the presence and absence of registers in the pre-training data \cite{myntti_register_2025}.

The task of document-level \ac{LLM}-generated text detection is closely linked to our setup as the same features allowing a classification can be used to study language on a corpus level. Two common approaches for \ac{LLM}-generated text detection are (i) using \acp{LLM} in a white- or black-box fashion and (ii) building a discriminative model based on linguistic features \cite{wu_survey_2025}. Models like Ghostbuster \cite{verma_ghostbuster_2024} or DetectGPT \cite{mitchell_detectgpt_2023} both use the probabilities emitted by \acp{LLM} to detect \ac{LLM}-generated texts.  Moreover, models using linguistic features have been shown to be sufficient to detect \ac{LLM}-generated texts in a variety of different settings \cite{yadagiri_ai-generated_2025, li_linguistic_2025, zaitsu_stylometry_2025,przystalski_stylometry_2026}. 

While previous research has focused on detecting LLM generated texts at document level, our goal is to measure the differences between \ac{LLM} generated language and human language at corpus level. Closest to our work is the comparison measure MAUVE \cite{pillutla_mauve_2021}. MAUVE uses quantized embeddings of an \ac{LLM} to assess the distance between two corpora by divergence frontiers \cite{djolonga_precision-recall_2020} with the KL-divergence.  MAUVE yields a distance metric, which closely correlates with human judgment. A major drawback of MAUVE is that it relies on model embeddings which are not human-interpretable.

There is a growing number of analytical studies that have analyzed the difference between human and \ac{LLM}-generated texts at corpus level using linguistic features. These works either focus implicitly on a single register \cite{zamaraeva_comparing_2025, bagdasarov_like_2025, georgiou_differentiating_2025} or look at  multiple registers \cite{sardinha_ai-generated_2024, reinhart_llms_2025, milicka_ai_2025, milicka_benchmark_2025}. While these studies all show systematic differences in language production between humans and \acp{LLM}, they come with several limitations.

First, prior studies often assess register-alignment under \textbf{explicit-register conditioning}, typically by asking models to continue a sufficiently large, register-specific human text \cite{reinhart_llms_2025, milicka_ai_2025, milicka_benchmark_2025}. In this setup the linguistic patterns for the specific register are already present in the example and the model is simply required to continue these patterns. Such an approach primarily assesses the model’s pretraining rather than an internal notion of register. In contrast, we define \textbf{implicit-register conditioning} as a way of prompting that requires the model to infer the register from the information about the language production contained in the prompt. While this prompting technique is used in \cite{sardinha_ai-generated_2024}, the author uses only a generic prompt for each register, which can only provide a limited amount of context information.

Second, several analyses compare marginal sample distributions or co-occurrence patterns of linguistic features, without providing a measure for the observed differences. While such results give empirical insights, they lack a clear metric that can be optimized in future work.

We propose a framework for measuring human-likeness of generated texts under (i) implicit-register conditioning: the model has to infer the target register from the situational information provided in the prompt, with every sample being generated with a different prompt to ensure a variety of contexts are taken into account. Our framework (ii) employs controlled and transparent prompting in a zero- and few-shot setting, (iii) compares multiple open-source models across architectures and scales, and (iv) formulates human-likeness as a two-sample problem over multidimensional and human-interpretable linguistic features.

\section{Methodology}
An overview of our method is shown in Figure \ref{fig:method-overview}. The first part consists in corpus generation. Given a target register $r$, we first collect a human reference corpus, subsample a representative subset and use its metadata to construct natural language prompts. These prompts instruct \acp{LLM} to generate texts using similar situational contexts. As the prompts do not contain parts of the original text, they correspond to our definition of implicit-register conditioning. Using different \acp{LLM} in a zero- and few-shot setting, we generate parallel corpora $\tilde{C}_{r,\theta}$ for our register $r$. For each text we extract a set of linguistic features, and encode each corpus into a multidimensional feature space. Finally, in this space, we frame the problem of human-likeness as a two-sample problem quantified by  the measure $\mathrm{MMD}(\tilde{C}^1_r, \tilde{C}^{r,\theta})$. To assess, whether the observed distance indicates a meaningful distance, we compare it against $\mathrm{MMD}(\tilde{C}_r^1, \tilde{C}_r^2)$, where $\tilde{C}_r^1$ and $\tilde{C}_r^1$ are two disjoint subsets repeatedly subsample from the full corpus $C_r$ to get a \ac{CI} for the expected \ac{MMD} between human samples.

\subsection{Data generation}
Let $R=\{r_1,\dots,r_K\}$ denote a set of target registers.
For each register $r\in R$, we define a corpus
$C_r=\{(t_i, m_i)\}_{i=1}^{N_r}$,
where $t_i \in V^*$ is a human-authored text, $V^*$ the set of all finite sequences over a vocabulary 
$V$, $m_i$ is metadata giving specific context about $t_i$ and $N_r$ is the size of the corpus.\\
We express implicit-register conditioning by defining, for each register $r$, a natural-language prompt template  $p_r(m_i)$ that translates metadata into a task-prompt which describes a language generation task. The prompt template provides general information of the register, for example, if the mode is written or spoken or if the language is used in an online space. The metadata then provides the specific setting of the language production, such as information about the people involved, the topics of a conversation or the contents covered in an article. 

Given a model with parameters $\theta$ we generate model outputs $\tilde{t}_i$ by autoregressive sampling with prompt template $p_r(m_i)$ resulting in a register-specific synthetic corpus
$\tilde{C}_{r,\theta}=\{(\tilde{t}_i, m_i)\}_{i=1}^{N_r}$.

Taking a set of human demonstrations, we can define a few-shot prompt template $p_{r,s}(m_i,\tilde{C}_r)$, where $\tilde{C}_r$ is a corpus of human-authored texts of register $r$ with $(t,m)\in \tilde{C}_r \implies (t,m) \notin C_r$ and $s$ the number of "shots", i.e., the number of texts taken from $\tilde{C}_r$ to construct the prompt. Using this prompt template we can generate a few-shot synthetic corpus $\tilde{C}_{r,\theta, s}$ by sampling our model in the same fashion. The few-shot examples are selected with a seeded random function, such that every model uses the same prompt for each instance $\tilde{t}_i$. While the model sees one or multiple examples for register appropriate language in this setting, these examples use a different context than the text the model produces. Therefore, the few-shot setting also counts as implicit-register conditioning, as the model has to learn in-context what appropriate language looks like for the register, but apply this to a new example.
\subsection{Distributional framework}
Given a corpus $C$ as described above, we define a set of vectors $X=\{(x_i)\}_{i=1}^{N_r}$, where $x_i \in \mathbb{R}^d$, as the resulting set when applying a function $f: V^* \mapsto \mathbb{R}^d$ to each element in $C$. Each dimension in $x_i$ describes the frequency of one specific linguistic feature in text $t_i$. In our work, the function $f$ is implemented by a program counting the occurrence of specific linguistic features in a text.

In this framework, we treat the observed sets $X_r$ and $\tilde{X}_{r,\theta}$ as samples from two underlying distributions $P_r$ and $P_{r,\theta}$. In this interpretation, quantifying the discrepancy between a model and the desired human distribution is calculated by $D(X_r, \tilde{X}_{r,\theta})$, where $D(\cdot,\cdot)$ is some function calculating a distance between two samples of two distributions. In other words, we formulate the problem of human likeness as a two-sample problem in a linguistic feature space.

\subsection{Distance Metrics}

We implement $D(X_r, \tilde{X}_{r,\theta})$ using the \ac{MMD}. Let $\mathcal{H}_k$ be a \ac{RKHS} associated with a characteristic kernel $k$. The squared \ac{MMD} between two distributions $P$ and $Q$ is defined as:
\[ 
\begin{aligned} 
    &= \sup_{\|f\|_{\mathcal{H}_k}\le 1} \left(\mathbb{E}_{x\sim P}[f(x)] - \mathbb{E}_{y\sim Q}[f(y)]\right)^2 \\ 
    &= \left\|\mu_P - \mu_Q\right\|_{\mathcal{H}_k}^2, 
\end{aligned} 
\]
where $\mu_P := \mathbb{E}_{x\sim P}[k(x,\cdot)]$ and $\mu_Q := \mathbb{E}_{y\sim Q}[k(y,\cdot)]$ denote the kernel mean embeddings. In other words, \ac{MMD} measures the distance between the mean embeddings of two distributions in the \ac{RKHS}. For characteristic kernels, the mean embedding is injective, implying that $\mathrm{MMD}(P,Q)=0$ if and only if $P=Q$ \cite{gretton_kernel_2012}.

Given two samples $X=\{x_i\}_{i=1}^{m}\sim P$ and $Y=\{y_j\}_{j=1}^{n}\sim Q$, a biased empirical estimate of the $\mathrm{MMD}^2_k(P,Q)$ is given by
\[
\begin{aligned}
\mathrm{MMD}^2_k(X,Y)
&= \frac{1}{m^2}\sum_{i=1}^{m}\sum_{j=1}^{m} k(x_i,x_j) \\
&\quad - \frac{2}{mn}\sum_{i=1}^{m}\sum_{j=1}^{n} k(x_i,y_j)\\
&\quad + \frac{1}{n^2}\sum_{i=1}^{n}\sum_{j=1}^{n} k(y_i,y_j).
\end{aligned}
\]
We use an RBF kernel for $k$. The bandwidth parameter is estimated via the median heuristic, i.e., using the median pairwise distance among pooled samples $X\cup Y$ \cite{gretton_kernel_2012}. This is a standard approach commonly used in the evaluation of generative models \cite{zhu_deep_2021, liang_enhancing_2018, long_learning_2015, dziugaite_training_2015}.

While other distances could be used to implement $D(\cdot,\cdot)$, we selected the \ac{MMD} as it only depends on a single kernel with a single bandwidth parameter. This is a clear advantage over metrics based on the Kullback-Leibler divergence, which require explicit estimation of probability density functions as used in \cite{bagdasarov_like_2025}. 

\subsection{Sample size}
In practice, available corpora are often too large for our method to be computationally feasible. To strike a balance between distributional expressiveness and computational cost, we sample each corpus using the following method.

Given a large corpus $C_{r,full}$ and the according set of linguistic feature vectors $X_{r,full}$ as defined above, where $full$ means we are using the entire dataset available. We resample a \ac{CI} for a sampling size $n<<|C_{r,full}|$ by $B$ times subsampling two exclusive sets $X_1,X_2\subset X_{r,full}$ of size $n$ and calculating $\mathrm{MMD}^2(X_1, X_2)$. From the resulting $B$ distances, we calculate an $m$\% \ac{CI} by taking the upper and lower $m/2$ percentiles. Repeating this for different $n$ we can find a dataset size that is a balanced tradeoff between sampling error and computational effort.

For the selected $n$, the upper limit of our $m\%$ \ac{CI} of the \ac{MMD} gives a good empirical upper bound for the distances we would expect to see between two samples of the human distribution for our sample size. If the observed distance between a model and the human dataset is larger than the upper limit of our \ac{CI} this is a strong indicator that under the selected experiment setup, the model sample is farther away than what we would typically expect between two samples of the human distribution.

While we only use a subsampled version of the full human corpora for our experiments, we use the full corpora of each register to standardize all experiments with the full human mean and standard deviation. Because the \ac{MMD} is only comparable with respect to the same \ac{RKHS}, we use a pooled version of our full human corpora to get the bandwidth of the kernel using the median heuristic for the respective register \cite{gretton_kernel_2012}. This ensures that all distances calculated within one register are comparable.   

\subsection{Prompt stability}
To test the stability of our generation procedure, we define a set of alternative versions of prompt templates $P_r^\prime=\{p_{r,1}^\prime, ..., p_{r,n}^\prime\}$ which differ from $p_r$ in ordering, tone, and formality. By constructing different versions of synthetic corpora $C_{r,i}$, it is possible to test whether the observed effects are stable over different prompting techniques. 
\section{Experimental Setup}
\subsection{Linguistic Framework}
In this study, we use the set of linguistic features introduced in \cite{biber1988variation}. The feature set consists of 67 linguistic features. In his work, \citeauthor{biber1988variation} identified six latent dimensions of linguistic variation on a dataset of 23 registers. These dimensions are based on the co-occurrence of the 67 linguistic features in the corpus of the study. The dimensions are interpreted by the functions that their underlying features commonly fulfill in texts, from which a descriptive dimension name is derived, such as Involved vs. Information Production or Narrative vs. Non-Narrative Concerns. By linearly combining the features using the weights published in \cite{biber1988variation} the \textit{Dimension Score} of a text can be calculated. In corpus-based text-linguistics, comparing these dimension scores between different registers is a standard research method. For the full list of features and their weights on the six dimensions please refer to Table \ref{tab:biber1988-loadings}. 


\subsection{Data}
We apply our framework to five different datasets spanning different communicative purposes and situational contexts:
\begin{itemize}
    \itemsep 0em 
    \item \textbf{Spoken conversation}: \textit{BNC2014Spoken} — Transcribed recordings of naturally occurring conversations with rich metadata on speakers, topics, and situational context \cite{love_spoken_2022}.
    \item \textbf{Academic writing}: \textit{S2ORC\_ACL} — Introductions of ACL main conference papers (2009--2018), parsed via the S2ORC API, with abstracts used as document-level metadata \cite{lo_s2orc_2020}.
    \item \textbf{Instructive online text}: \textit{wikiHow} — Instructional articles describing step-by-step procedures, using article text as content and titles/headlines as metadata \cite{koupaee_wikihow_2018}.
    \item \textbf{Creative writing}: \textit{WritingPrompts} — User-written fictional stories generated in response to prompts posted by other users \cite{fan_hierarchical_2018}.
    \item \textbf{News reporting}: \textit{XSum} — BBC news articles paired with single-sentence summaries \cite{narayan_dont_2018}.
\end{itemize}

We apply standard and dataset-specific preprocessing methods to the data, which are described in more detail in Appendix \ref{sec:data_preprocessing}. The selected multidimensional evaluation framework yields stable linguistic features for texts starting from 400 tokens. For computational efficiency purposes, we limit all our human and generated texts to a length of 400  tokens with a soft limit to the end of the next sentence up to 440 tokens. We apply this both to human and generated texts to avoid adding positional biases to our data. All datasets use data before the release of ChatGPT (November 30, 2022) to avoid \ac{LLM}-generated text entering as human data.  For more detail, refer to Appendix \ref{sec:sample_size_appendix}.

\subsection{Generation Setup}
We evaluate a set of seven open-source \acp{LLM} spanning multiple model families, parameter scales, and training strategies. The evaluated models are Apertus 70B \cite{hernandez-cano_apertus_2026}, Llama 3.3 70B and Llama 3.1 8B \cite{grattafiori_llama_2024}, Qwen 3 32B and Qwen 3 8B \cite{yang_qwen3_2025}, and Gemma 3 27B and Gemma 3 12B \cite{team_gemma_2025}. 
We restrict our scope to instruction-tuned models, as only pretrained models tend not to follow instructions \cite{weiFINETUNEDLANGUAGEMODELS2022, ouyangTrainingLanguageModels}, making them unsuitable for our setup.

The prompt templates are written following general best practice. One example prompt used in our creative writing experiments is shown in Table \ref{tab:prompt-spec}. In this example, the metadata is a writing prompt provided by a user in an online forum. For further detail on the other experiment prompts please see Appendix \ref{sec:prompts_appendix}. 
To keep the computation within reasonable bounds, we only study prompt stability for the BNC2014Spoken dataset. 

\begin{table}[t]
\centering
\small
\begin{tabular}{p{2.3cm} p{4.7cm}}
\hline
\textbf{Component} & \textbf{Content} \\
\hline
System prompt &
You are a participant on an online creative writing forum where users write short stories inspired by prompts from other members. Your task is to write the opening section of a story based on the given prompt. Write the beginning of a story of at least 400 words. You do not need to finish the story. Please output only your story text. \\

\hline
User prompt &
Please write a story for the following prompt: \texttt{\{prompt\}} \\

\hline
Assistant prefix &
Certainly, here is my answer: \\

\hline
Few-shot template &
Certainly, here is my answer: \texttt{\{story\}} \\
\hline
\end{tabular}
\caption{Prompt specification for the creative-writing corpus (WritingPrompts). Curly-braced expressions denote placeholders for the metadata replaced at inference time.}
\label{tab:prompt-spec}
\end{table}

\subsection{Implementation Details}
Depending on the size of the model, we use 1 or 2 A100 GPUs with 80 GB of VRAM each. For inference, we use the vLLM library to load the open-source models from HuggingFace. All generations are done using a temperature=1 and top-p=1. This is equal to sampling the full distribution of the model given the prompt. While sampling parameters have an influence on the lexical diversity of generated texts, they tend not to have a systematic influence in terms of a \ac{MDA} after Biber \cite{milicka_ai_2025}. 
To extract the linguistic features, we use pybiber\footnote{Implementation available at: \url{https://github.com/browndw/pybiber}. \cite{sardinha_ai-generated_2024} who also works within the Biber framework for register features uses the Biber Tagger instead, which is not freely available.} 


\section{Results}
\subsection{MMD stability}
\label{sec:MMD_stability}
\begin{figure}[t]
  \centering
  \includegraphics[width=\columnwidth]{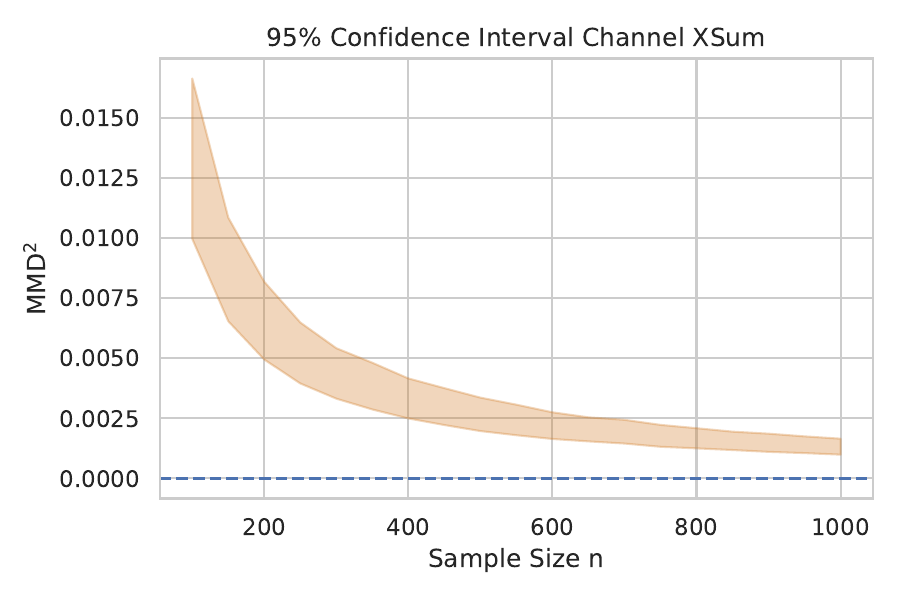}
  \caption{$\mathrm{MMD}^2$ with a resampled confidence interval for different sample sizes on the XSum dataset.}
  \label{fig:MMD_stability}
\end{figure}

Figure \ref{fig:MMD_stability} shows the human-human $\mathrm{MMD}^2$ for the XSum dataset with a 95\% \ac{CI}. Respective plots for the other datasets are in Appendix \ref{sec:sample_size_appendix}. For all datasets, we observe a sharp decrease in mean $\mathrm{MMD}^2$ distance and a tightening of the \ac{CI} when the sample size is increased from $n$=50 to $n$=400. Afterwards, we observe an asymptotic decrease and tightening when further increasing the sample size. For this study, we fix our sample size to n=600 for all experiments. We selected this sample size as a suitable tradeoff between sampling error and computation time.

\begin{figure*}[t]
  \centering
  \includegraphics[width=0.85\textwidth]{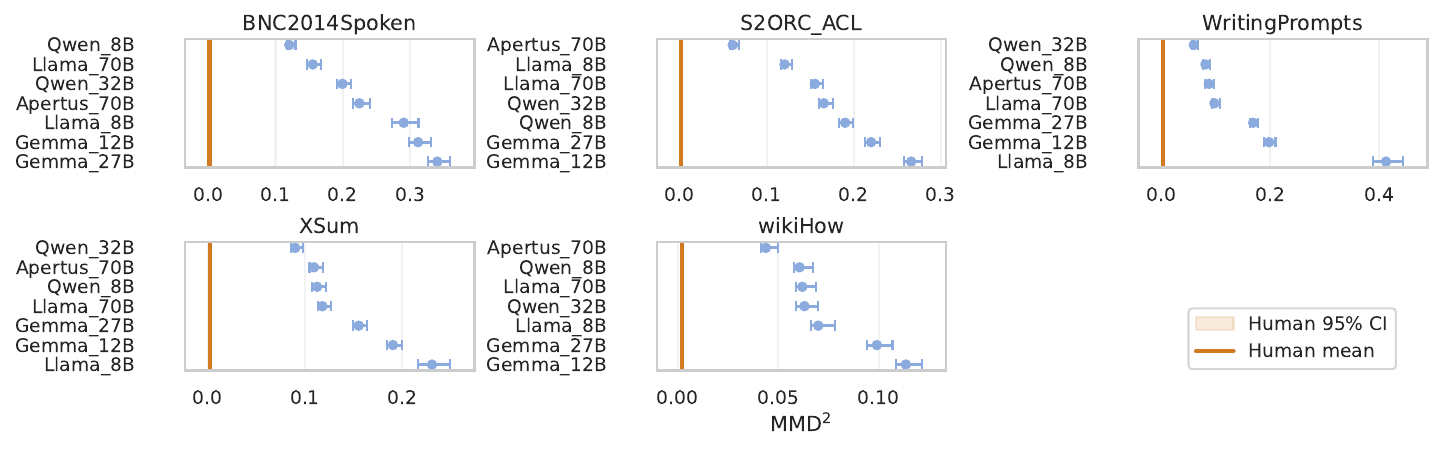}
  \caption{$\mathrm{MMD}^2$ for all datasets and models to the respective human corpus in the zero-shot setting, where the points indicate the observed $\mathrm{MMD}^2$ and the whiskers show the 95\% \ac{CI} resampled on coupled samples from the human and model corpus.  The orange line in each plot gives the respective Human-Human $\mathrm{MMD}^2$ for the respective datasets with the resampled \ac{CI}. The models on the y-axis are sorted by their observed $\mathrm{MMD}^2$ distance. Because the distances between human samples are much smaller than the distances between models and human, the human baseline \ac{CI} is not visible.}
  \label{fig:MMD_ranking}
\end{figure*}

\subsection{Model vs human}
In Figure \ref{fig:MMD_ranking}, we show the $\mathrm{MMD}^2$ distance between the synthetic corpora and the human corpus for each register with an estimated 95\% \ac{CI}. The sampling error between human-human discussed in Section \ref{sec:MMD_stability} is much smaller than the observed human-\ac{LLM} distances, which further justifies the selected sample size for the experiments.

For all datasets, all tested models under our experiment setup produce a distribution of linguistic features that is farther away than what we would expect to see between two human samples of the same size. The distance ordering of the models changes by dataset, thus human-likeness indeed needs to be evaluated by register. The Qwen models produce the most human-like distribution on the BNC2014Spoken, WritingPrompts and XSum datasets, even though they have a relatively few parameters compared to other models tested. On the S2ORC\_ACL and wikiHow datasets, Apertus 70B is the most human-like model tested.

Relative to the human baseline, we can observe that some registers are harder for the models to generate. While the upper limit of the \ac{CI} on the $\mathrm{MMD}^2$ between human samples is about 0.0025 for all registers, we can observe that for Spoken conversation the observed $\mathrm{MMD}^2$ values are generally many times larger than this baseline when compared to the wikiHow dataset.
\subsection{Few-Shot generation}
\begin{figure}[t]
  \centering
  \includegraphics[width=\columnwidth]{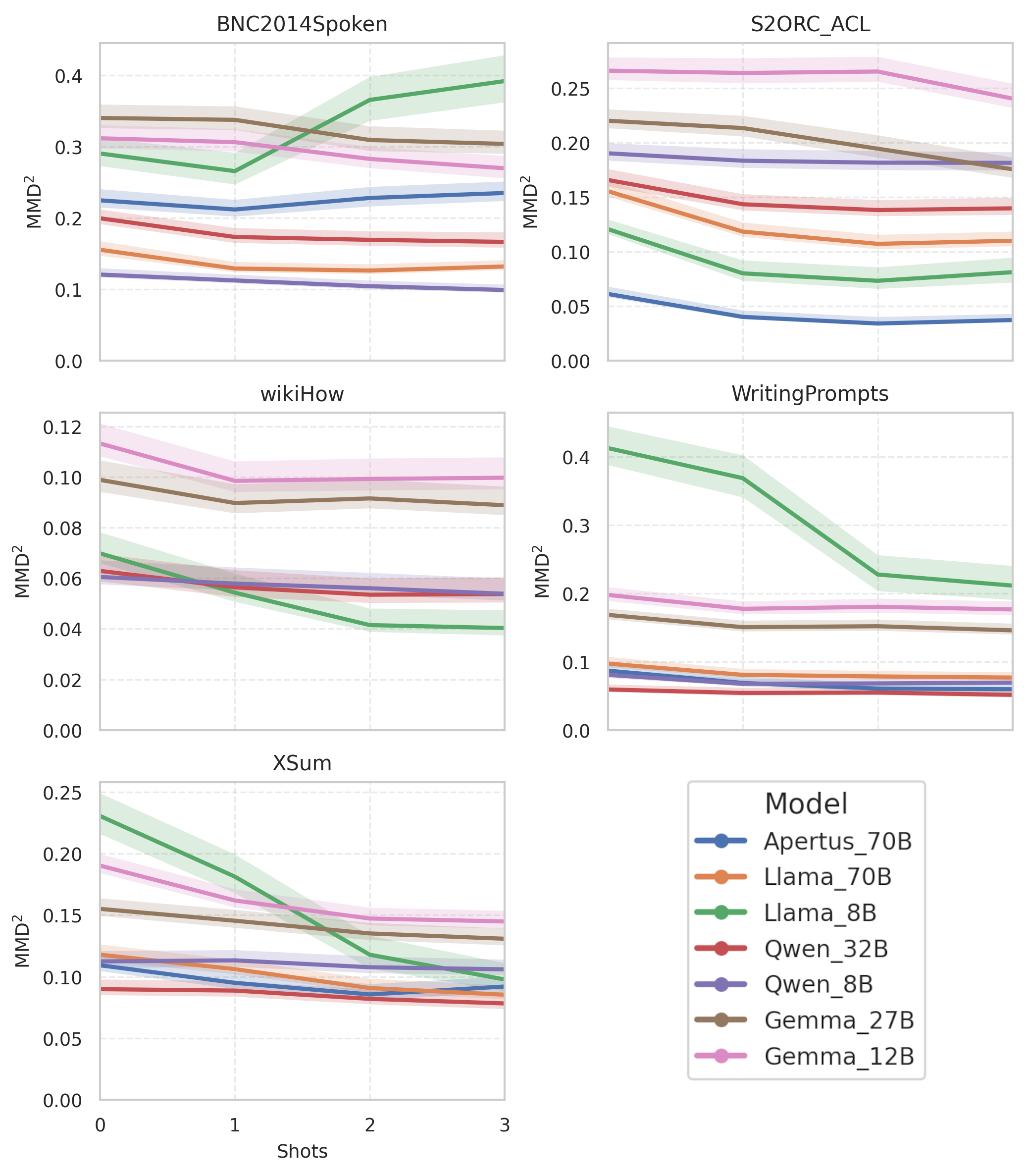}
  \caption{$\mathrm{MMD}^2$ for the n-shot experiments for all models and datasets. The solid line represents the observed $\mathrm{MMD}^2$ values, while the shaded area gives the 95\% \ac{CI} resampled on coupled samples from the human and model corpus.}
  \label{fig:few_shot_mmd}
\end{figure}

Figure \ref{fig:few_shot_mmd} shows the influence of few-shot examples on the $\mathrm{MMD}^2$. In some cases, providing the model with examples in the prompt helps to generate samples closer to the human distribution, the effect however is minor. For the WritingPrompts and the XSum dataset, Llama 8B in the Zero-Shot setting generates samples which are relatively far away from the human distribution. Through demonstrations, the distance drastically decreases. Interestingly, for the BNC2014Spoken dataset, the one-shot setting produces more human-like text than the two- and three-shot setting for Llama 8B generations.

\subsection{Prompt stability}
\begin{figure}[t]
  \centering
  \includegraphics[width=\columnwidth]{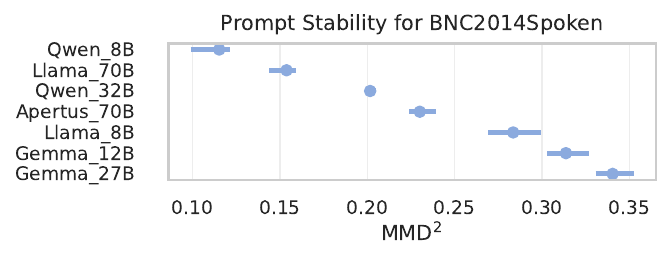}
  \caption{$\mathrm{MMD}^2$ for the zero-shot prompt stability experiments to the human reference sample of the BNC2014Spoken. Dots indicate the mean value over all prompts, while the band shows the minimum and maximum observed distance for the respective model under all prompt variations.}
  \label{fig:stability_mmd}
\end{figure}

Figure \ref{fig:stability_mmd} shows the results of the prompt stability experiments. While a variation in the prompt changes the absolute value of the observed distances, the overall ranking of the model is consistent over all prompt variations. Smaller models appear to be more susceptible to prompt variations. 

In Figure \ref{fig:stability_correlation}, the correlation of the observed $\mathrm{MMD}^2$ between the different prompt variations is shown. All correlation values are $\geq 0.985$, which indicates that all prompt variations are highly correlated.  

\subsection{Biber dimensions}
\label{sec:biber_dimensions}
\begin{figure}[t]
  \centering
  \includegraphics[width=\columnwidth]{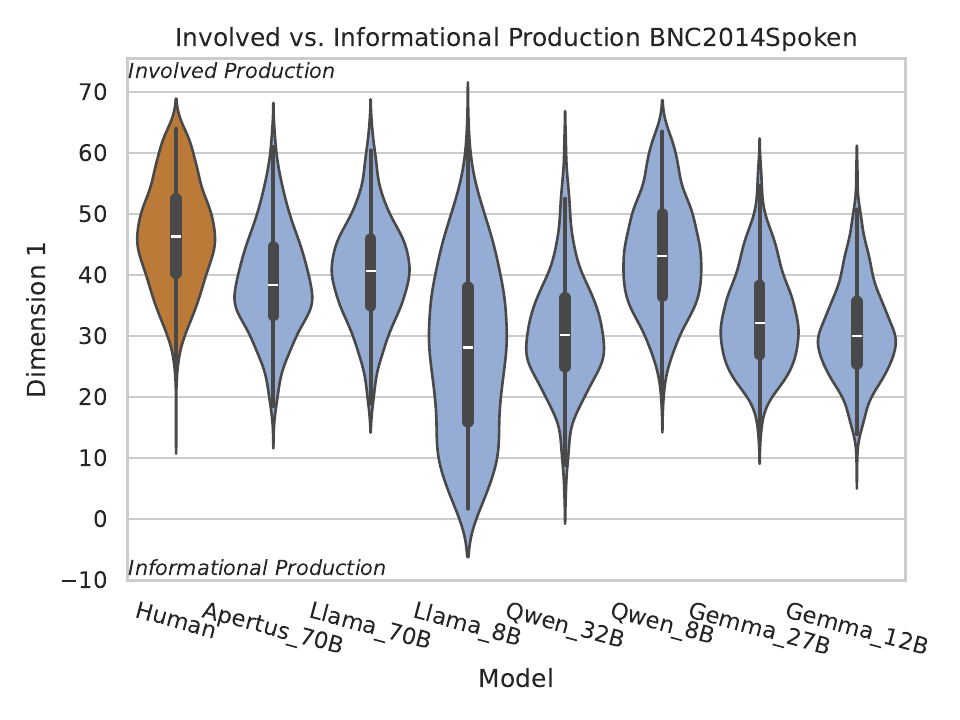}
  \caption{Violinplot of Biber dimension 1 on BNC2014Spoken for human and models in the Zero-Shot setting}
  \label{fig:BNC2014Spoken_Dim1}
\end{figure}

While the \ac{MMD} provides a metric for the overall distance between human and \ac{LLM}-generated texts, it offers no human interpretation of the observed distances. For that purpose, we analyze the six dimensions introduced in \cite{biber1988variation}. 
An example for the first dimension on the BNC2014Spoken dataset is shown in Figure \ref{fig:BNC2014Spoken_Dim1}. This dimension can be interpreted as how involved or informational a text is. For Spoken conversation all tested models produce texts that are less involved than the human baseline.  

The resulting plots for all datasets and dimensions can be found in Figures \ref{fig:app:biber-bnc2014spoken}, \ref{fig:app:biber-S2ORC_ACL}, \ref{fig:app:biber-wikiHow}, \ref{fig:app:biber-WritingPrompts} and \ref{fig:app:biber-XSum}. Some notable observations can be drawn about the behavior of models. Regarding dimension 2, concerned with narrative vs non-narrative discourse, for academic texts, all models except Apertus, tend to generate texts more narrative than the human baseline, while for Creative Writing, some models are less narrative, while others are comparable or more narrative than the human baseline. On dimension 4, Overt Expression of Persuasion, for the wikiHow dataset and Creative Writing, human texts tend to be more persuasive than the tested models.  
\subsection{Features}
\label{sec:features}
Looking at the marginal feature distributions can already show simple patterns of difference in human-understandable terms. These results are further detailed in Appendix \ref{sec:appendix_features}.

For all registers, all models tend to use 'Present Participle Clauses' and 'Nominalizations' more frequently than humans, except for the Qwen models on the WritingPrompts dataset. This is consistent with previous research \cite{reinhart_llms_2025}. On all datasets, all models tend to produce longer words than humans. Except for the Qwen models on the WritingPrompts dataset, all models tend to use 'Attributive Adjectives' more frequently, and, except for Llama 8B and Gemma 12B on the WritingPrompts dataset, use 'Past-Tense' less frequently. The feature 'Other Nouns' occurs more frequently in spoken conversation, Creative Writing, and wikiHow, but less frequently in Academic writing. 

Apart from these global trends, interesting patterns emerge for single registers. For Spoken conversation, all models show increased rates of 'Pied Piping', while for Creative Writing, 'Past Participle Clauses' are frequently used more often than in human texts. For Academic Writing, 'Split Infinitives' and 'Phrasal Coordinations' occur more frequently across all models, and in XSum, '\textit{That} as a Subject' is generally produced more often. These patterns show that certain biases for single features exist across all model families and registers, while some biases are register-specific.

\section{Ablation}
\subsection{Model-Model-Distances}
Using the same bandwidth and normalization as for the human-model distance, the distance between models for one register can be calculated. The results are shown in Figures \ref{fig:Model_Model_MMD_BNC2014Spoken},  \ref{fig:Model_Model_MMD_S2ORC_ACL}, \ref{fig:Model_Model_MMD_wikiHow}, \ref{fig:Model_Model_MMD_WritingPrompts}, \ref{fig:Model_Model_MMD_XSum}. It can be observed that models of the same family tend to cluster linguistically together. Models of the Gemma family show in general big differences to all other models. 

\subsection{Linguistic feature variability}
We compute the sum of the variances of all 67 linguistic features after standardization. We compare them to a human baseline, for which we resampled a \ac{CI} for the variance on the full human sample.

The full results are shown in Figure \ref{fig:variances_models}. For all datasets except the BNC2014Spoken, the models tend to show less variability than the human baseline. For the BNC2014Spoken, the models show more variability than humans. Interestingly, for academic writing, both Apertus 70B and Llama 8B are within the human \ac{CI}. From this, we can conclude that for four out of five registers, the tested models tend to underproduce the variability observed in humans. 

\section{Discussion}
In our experimental setup, all models, for all prompting strategies, fail to produce a linguistic-feature distribution that is within the distance we would expect between human samples of the same size. The relative ordering between the models changes by register. These results highlight the necessity of a register-aware evaluation method for studying \ac{LLM} language. Aggregating results across registers could obscure systematic differences, as a model ranked among the most human-like for one register, does not necessarily rank among the most human-like on another register. This is further supported by our classifier ablation in Appendix \ref{sec:classifier}, where a classifier jointly trained on all registers achieves lower accuracy than a register-specific classifier. 
Overall, our results are in line with register research emphasizing that linguistic patterns reflect the situational context of language use (see e.g. \cite{biber1988variation},  \cite{biber_register_2019}). In our implicit-register conditioning model,  generations tend to improve with human examples, but this does not hold for every model and register. Even when provided with up to three human examples, no model tested under our experiment setup was able to produce a human-like distribution of linguistic features. 

\ac{LLM} behavior is primarily influenced by model architecture, training strategy, and training data. We summarize the available information about the tested models in Appendix \ref{sec:model_details}. Based on this information, we formulate three hypotheses to explain some of the observed behaviors, focusing on training data, model family, and instruction-tuning.

First, distillation and synthetic data in model training do not necessarily alter the register-alignment of models: Across all registers, Apertus 70B and the Qwen variants consistently rank among the models closest to the human reference distribution. The Qwen models are pretrained on large-scale mixtures of public and synthetic data and involve explicit distillation from teacher LLMs, whereas Apertus is trained from scratch on non-distilled data. 

Second, family-specific training and architectures imprint stable stylistic priors on the linguistic feature distribution for our observed registers: Models of the same family tend to cluster closer together than models of different families. This even holds for models of vastly different parameter sizes. The question remains whether these stylistic priors are introduced from differences in the training data or from other model hyperparameters such as the pretraining loss, the alignment strategy or even differences in activation functions. Fully open-source models like Apertus can be used in future research to study where these priors are introduced.

Third, instruction tuning introduces priors on the linguistic feature distribution of generated texts: Apertus was instruction-tuned on \textasciitilde 4 million samples \cite{hernandez-cano_apertus_2026}, while Llama 3.3 70B was instruction-tuned "on over 10 million" samples\footnote{Number taken from the official HuggingFace blog post \url{https://huggingface.co/blog/llama3}}. Except for the BNC2014Spoken dataset, Apertus is consistently closer to the human baseline than Llama 3.3. Also in explicit register-conditioning experiments, the non-instruction-tuned version of Llama is much closer to the human baseline \cite{reinhart_llms_2025}.

The \ac{MMD} metric introduced in this work can be optimized in future work to improve the language production of \acp{LLM}. Minimizing the \ac{MMD} between \ac{LLM}-produced texts and representative register samples will encourage \acp{LLM} to produce texts that are (1) linguistically appropriate for the purpose they fulfill and (2) the \ac{MMD} as a metric encourages models to produce texts from the full human distribution, rather than mode collapsing to a narrower distribution of linguistic features. Incorporating such a metric into the instruction tuning of a model can be used to track the deviation from human language over the training or even to test if the instruction tuning data itself follows a human-like distribution of linguistic features.

The data from this study can be used in subsequent work to further compare \ac{LLM} and human language, by for example using different or additional linguistic features. Furthermore, a more in-depth prompt sensitivity analysis would be an  interesting direction for future work. In this work, seven instruction tuned models on five different registers are tested. In future work, this could be extended to include more models and datasets by following the method outlined in this paper. A detailed guide how to repeat the experiments for other models is shown in Appendix \ref{app:benchmark}.

\section{Conclusion}
We propose a framework for evaluating the human-likeness of LLM-generated texts across different registers. To implement this framework, we used the classic linguistic features introduced in \cite{biber1988variation}. We pose human-likeness as a two-sample problem between a synthetic and human corpus measured by the \ac{MMD}. As such, this work takes a first step toward combining register-functional text linguistics with \ac{LLM} benchmarking. 

Our empirical analysis across five datasets and seven open-source \acp{LLM} demonstrates that \ac{LLM} generated texts differ under our setup in their distribution of linguistic features from a human baseline. These results are consistent for all tested models and are stable across prompt variations. Human-likeness as formulated by our framework is not dictated by model size, with the smaller Qwen 8B and 27B models performing relatively well across nearly all datasets. Overall, our results show that human-likeness is register-dependent and as such should always be evaluated in a register-sensitive manner.  

\section*{Limitations}
There are three limitations of this study that are worth mentioning. First, our analysis only used English language datasets and a fixed set of linguistic features from one specific corpus linguistic framework. While these features are well established in corpus linguistics, they are not exhaustive and likely do not cover all aspects of systematic deviation between \ac{LLM} and Human language.

Second, our experiments tested generation only under naive zero-shot and few-shot generation setups. While these setups yielded consistent results under our prompt stability analysis, n-shot prompting covers only a small part of prompting techniques. This leads to follow-up questions such as the influence of reasoning or further model tuning on the linguistic properties of model output.

Third, with the proposed framework, we only capture these particular linguistic properties of the generated texts. Any other properties of  \acp{LLM} and generated texts are not assessed. There is no guarantee that the texts closest to the human distribution are actually texts that fulfill other properties, such as fluency, factual correctness, or telling a convincing story.

\section*{Acknowledgments}
The authors gratefully acknowledge the scientific support and HPC resources provided by the Erlangen National High Performance Computing Center (NHR@FAU) of the Friedrich-Alexander-Universität Erlangen-Nürnberg (FAU). The hardware is partially funded by the German Research Foundation (DFG). The authors further acknowledge support from the Alexander von Humboldt Foundation as part of the Alexander von Humboldt Professorship endowed by the German Federal Ministry of  Research, Technology and Space (BMFTR).

\bibliography{custom}

\clearpage

\appendix
\renewcommand{\thefigure}{\Alph{section}.\arabic{figure}}
\renewcommand{\thetable}{\Alph{section}.\arabic{table}}

\section{Subsampling strategy}
\label{sec:subsampling_appendix}
To draw our evaluation and few-shot dataset out of the bigger human dataset we employ the following strategy. First, we exclude the 5\% samples with the longest metadata in terms of tokens for the Llama models from being selected as samples for these sets. This step is necessary due to outliers in terms of metadata for online registers, such as the WritingPrompts dataset, where some unusually long story prompts occur. Because of the relative small size and hand-curated nature of the BNC2014Spoken corpus, we exclude only the longest 3\%. This exclusion is necessary in order to make the generation step more efficient by providing an upper bound on the maximum context length required per prompt.

For the subsampling, we 1000 times draw a random dataset from the remaining human distribution and compute the aggregate marginal Wasserstein distance between the six Biber dimensions of the full dataset (including the samples with a long context) and the selected samples. We take the samples with the overall lowest Wasserstein distance as our corpus. We repeat the same procedure for the few-shot dataset, but additionally exclude all samples already selected for the first dataset from occurring in the few-shot dataset.

While it would have been possible to use the \ac{MMD} to validate the subsampling, the calculation grows quadratically with the dataset size. Using the differences between the subsample and the full human sample with the aggregated Wasserstein distance on the six dimensions of Biber serves as a simpler metric and ensures that the selected dataset is representative of the overall dataset for subsequent linguistic studies using the data. Additionally, we calculate Cohen's D between the marginal feature distributions of the selected subsample for the evaluation set and the full human distribution. For all datasets, the absolute average Cohen's D is <0.06, indicating very small effect sizes. \
\section{Prompts used in experiments}
\label{sec:prompts_appendix}
\begin{table}[t]
\centering
\small
\begin{tabular}{p{2.3cm} p{4.7cm}}
\hline
\textbf{Component} & \textbf{Content} \\
\hline
System prompt &
You are tasked with writing a conversation between 2 or more people given context about the speakers and the conversation. Write a conversation of at least 400 words. You do not need to finish the conversation or cover all topics mentioned. You can start with an already ongoing conversation. Please indicate each speaker with "Speaker\_1:", "Speaker\_2:", etc. \\

\hline
User prompt &
Please write a conversation given the following context: \texttt{\{Speaker\_Metadata\}} \texttt{\{Conversation\_Context\}} \\

\hline
Assistant prefix &
Certainly, here is my answer: \\

\hline
Few-shot template &
Certainly, here is my answer: \texttt{\{Conversation\}} \\
\hline
\end{tabular}
\caption{Prompt specification for the BNC2014Spoken corpus. Curly-braced expressions denote placeholders replaced at inference time.}
\label{tab:prompt-spec_BNC2014Spoken}
\end{table}

\begin{table}[t]
\centering
\small
\begin{tabular}{p{2.3cm} p{4.7cm}}
\hline
\textbf{Component} & \textbf{Content} \\
\hline
System prompt &
You are an author of ACL papers. Your task is to write the introduction of a paper given its title and abstract. Write an introduction of at least 400 words. Output only the introduction text. \\

\hline
User prompt &
Please write a paper given the following title and abstract: \texttt{\{title\}}\texttt{\{abstract\}} \\

\hline
Assistant prefix &
Certainly, here is my answer: \\

\hline
Few-shot template &
Certainly, here is my answer: \texttt{\{introduction\}} \\
\hline
\end{tabular}
\caption{Prompt specification for the S2ORC\_ACL corpus. Curly-braced expressions denote placeholders replaced at inference time.}
\label{tab:prompt-spec_S2ORC_ACL}
\end{table}

\begin{table}[t]
\centering
\small
\begin{tabular}{p{2.3cm} p{4.7cm}}
\hline
\textbf{Component} & \textbf{Content} \\
\hline
System prompt &
You are the author of a wikiHow article. Your task is to write the full article given the title and headline. Write an article of at least 400 words. Please only output the article text.\\

\hline
User prompt &
Please write an article given the following title and headline: \texttt{\{title\}}\texttt{\{headline\}} \\

\hline
Assistant prefix &
Certainly, here is my answer: \\

\hline
Few-shot template &
Certainly, here is my answer: \texttt{\{text\}} \\
\hline
\end{tabular}
\caption{Prompt specification for the wikiHow corpus. Curly-braced expressions denote placeholders replaced at inference time.}
\label{tab:prompt-spec_wikiHow}
\end{table}

\begin{table}[t]
\centering
\small
\begin{tabular}{p{2.3cm} p{4.7cm}}
\hline
\textbf{Component} & \textbf{Content} \\
\hline
System prompt &
You are a writer for a British newspaper. Your task is to write the beginning of an article based on a short summary of the content. If necessary, you can add names and other facts to the story. Write a beginning of at least 400 words. You do not need to finish the article. Please output only the article text.\\

\hline
User prompt &
"Please write an article given the following summary: \texttt{\{summary\}} \\

\hline
Assistant prefix &
Certainly, here is my answer: \\

\hline
Few-shot template &
Certainly, here is my answer: \texttt{\{document\}} \\
\hline
\end{tabular}
\caption{Prompt specification for the XSum corpus. Curly-braced expressions denote placeholders replaced at inference time.}
\label{tab:prompt-spec_XSum}
\end{table}

\begin{table}[t]
\centering
\small
\begin{tabular}{p{2.3cm} p{4.7cm}}
\hline
\textbf{Component} & \textbf{Content} \\
\hline
System prompt &
Your task is to write a multi-speaker conversation based on the provided context information about the speakers and situation. The conversation should be at least 400 words long. It may begin in the middle of an interaction. Label turns as 'Speaker\_1:', 'Speaker\_2:', and so on. \\
\hline
User prompt &
Generate a conversation using the following information: \texttt{\{Speaker\_Metadata\}} \texttt{\{Conversation\_Context\}} \\
\hline
Assistant prefix &
Sure, here is my answer: \\
\hline
Few-shot template &
Sure, here is my answer: \texttt{\{Conversation\}} \\
\hline
\end{tabular}
\caption{Prompt ablation 1 for the BNC2014Spoken corpus. Curly-braced expressions denote placeholders replaced at inference time.}
\label{tab:prompt-ablation_BNC2014Spoken_1}
\end{table}

\begin{figure}[!h]
  \centering
  \includegraphics[width=0.9\linewidth]{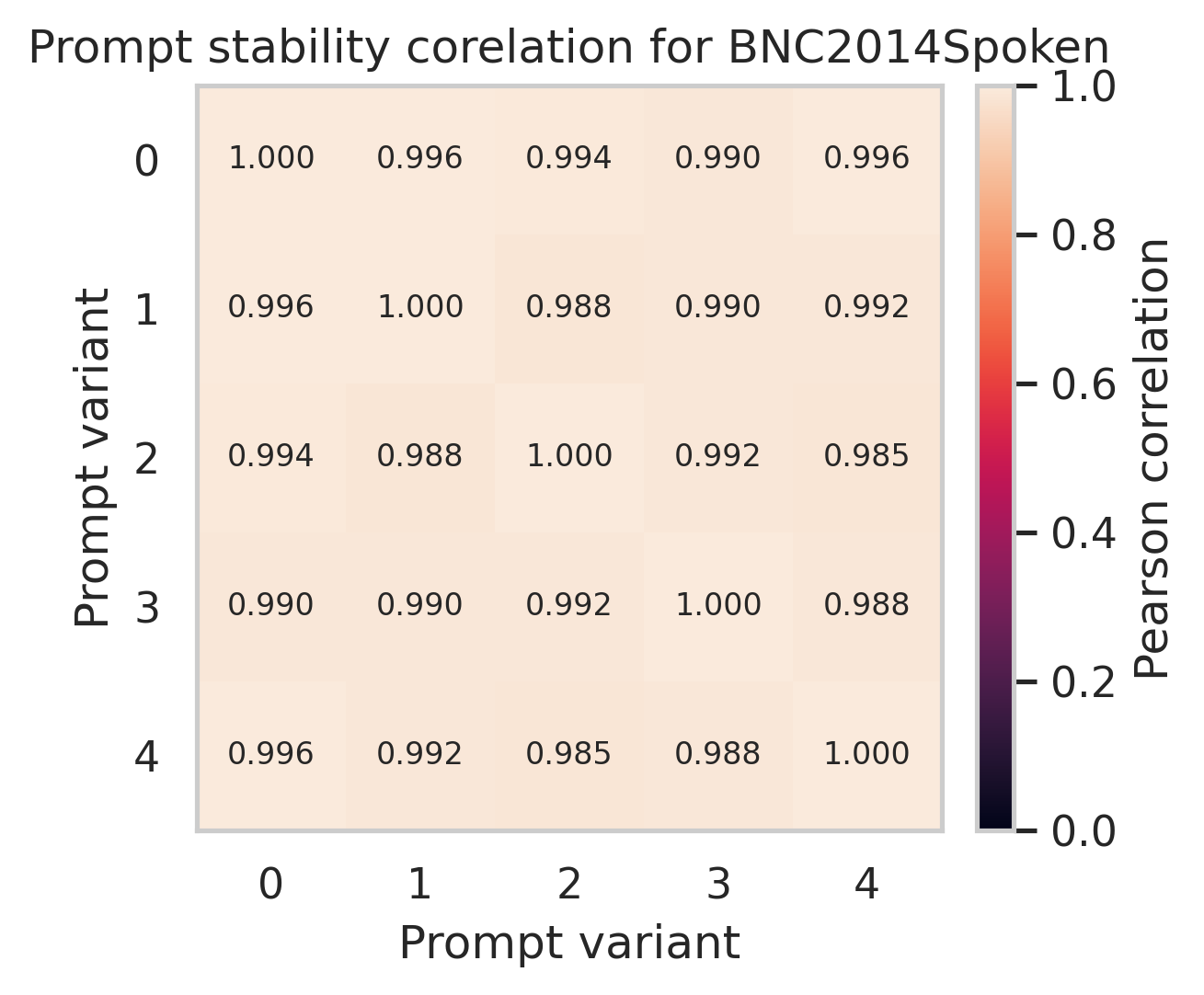}
  \caption{Correlation heatmap between the $\mathrm{MMD}^2$ between human and AI for the BNC2014Spoken between different prompt variants in the Zero-Shot setting.}
  \label{fig:stability_correlation}
\end{figure}
The prompts used for all experiments within one register are the same, except for the ablation protocol explained next. For the Qwen3 model we appended "\textbackslash nothink" to the prompt to avoid the model using its thinking mode. The  prompts are shown in Tables \ref{tab:prompt-spec_BNC2014Spoken}, \ref{tab:prompt-spec_S2ORC_ACL}, \ref{tab:prompt-spec_wikiHow} and \ref{tab:prompt-spec_XSum}. The metadata provided changes by register. For example in the BNC2014Spoken dataset conversation context contains the location, the relation of the speakers, the topics in the conversation, the activities of the people during the conversation and the type of conversation like discussion or telling anecdotes. Speaker information gives the age ranges, gender, nationality, dialect, linguistic origin, birthplace, occupation, education level and soci-economic status of the speakers. While in the WritingPrompts dataset only the writing instruction prompt of a user is given.

To test the prompt stability, we use the following protocol. We write four additional prompt templates where we vary the ordering of the instructions, tone, and phrasing of the instructions. With this setup, we want to estimate the bias introduced through stylistic variations of the prompt. The correlation between the different prompt variants across models is shown in Figure \ref{fig:stability_correlation}.
\section{Sample Size}
\label{sec:sample_size_appendix}
\begin{figure*}
    \centering
    \begin{subfigure}{0.49\linewidth}
        \includegraphics[width=\linewidth]{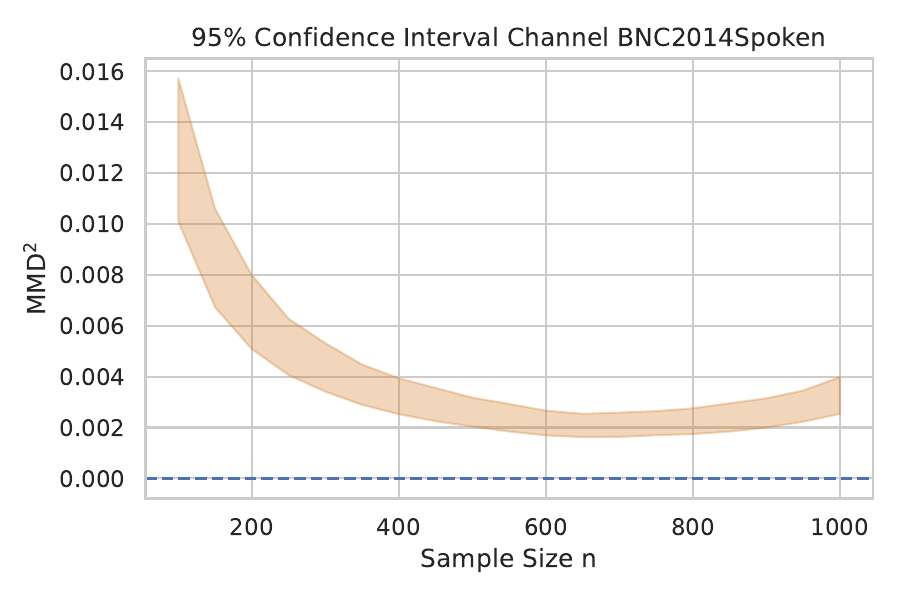}
        \caption{BNC2014Spoken}
    \end{subfigure}
    \begin{subfigure}{0.49\linewidth}
        \includegraphics[width=\linewidth]{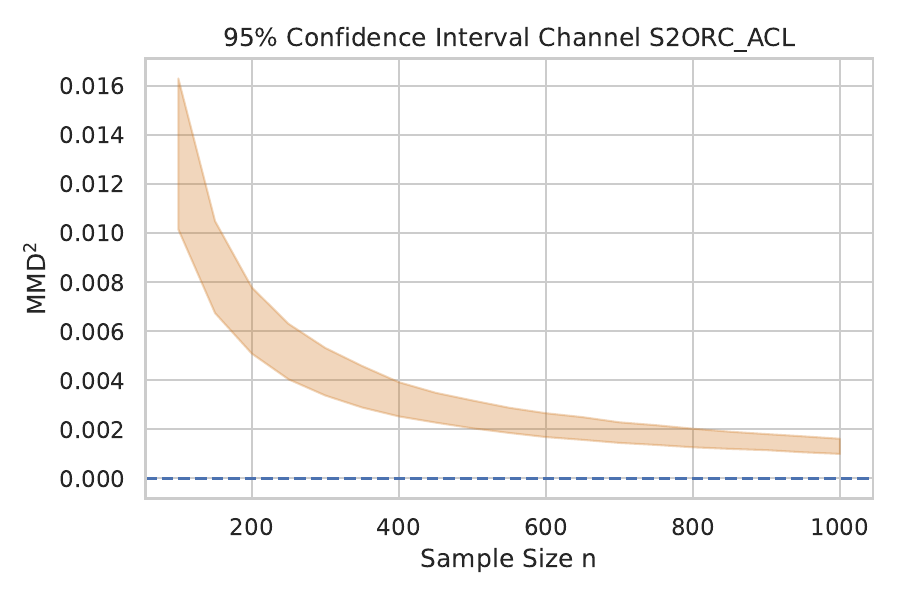}
        \caption{S2ORC\_ACL}
    \end{subfigure}
    \begin{subfigure}{0.49\linewidth}
        \includegraphics[width=\linewidth]{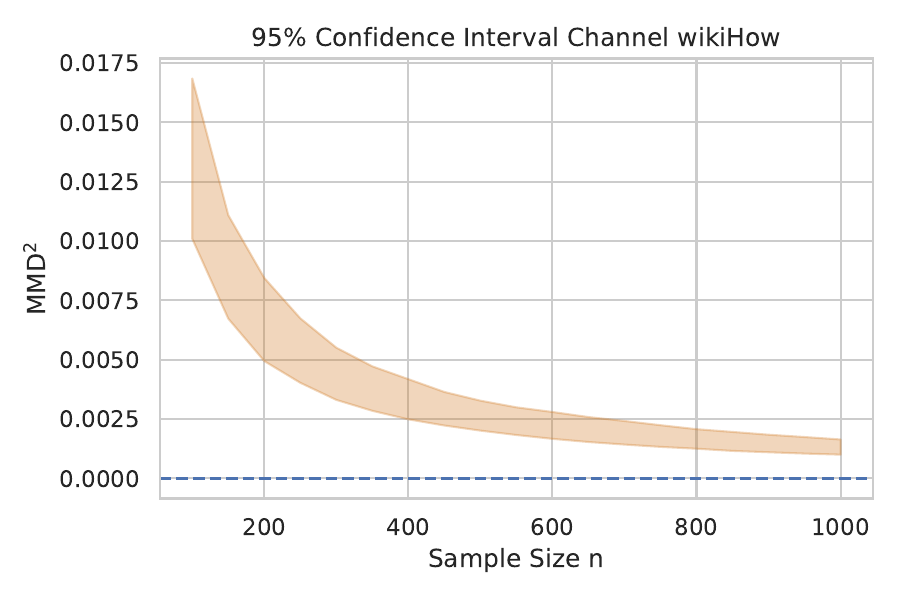}
        \caption{wikiHow}
    \end{subfigure}
    \begin{subfigure}{0.49\linewidth}
        \includegraphics[width=\linewidth]{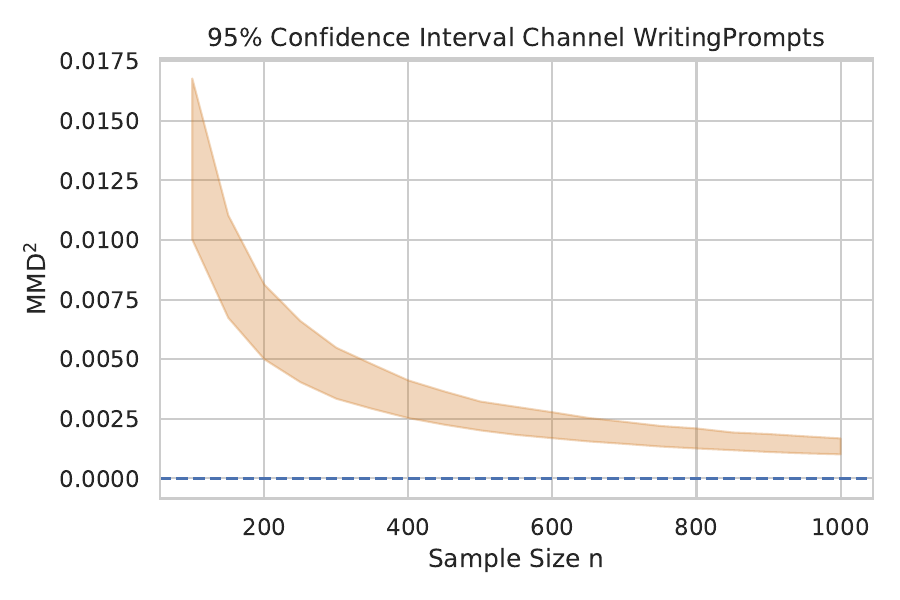}
        \caption{WritingPrompts}
    \end{subfigure}
    \begin{subfigure}{0.49\linewidth}
        \includegraphics[width=\linewidth]{figures/Sample_Sizes/MMD_Stability_Human_Human_XSum.pdf}
        \caption{XSum}
    \end{subfigure}
    \caption{$\mathrm{MMD}^2$ with bootstrapped confidence interval for different sample sizes on all datasets. For BNC2014Spoken error is increasing, since dataset has only ~1200 samples, thus a sample size larger 600 will lead to a smaller and larger subset.}
    \label{fig:subsampling_stability}
\end{figure*}

The results for the subsampling stability of the \ac{MMD} for all datasets are shown in Figure \ref{fig:subsampling_stability}.

\section{Data preprocessing}
\label{sec:data_preprocessing}
All datasets used in this study were obtained from publicly available resources and processed into a unified format consisting of a main text field and associated metadata. 

Spoken conversation data is drawn from the BNC2014 Spoken corpus, which contains transcribed recordings of naturally occurring conversations with detailed speaker- and context-level metadata; access to the corpus can be requested at \url{http://corpora.lancs.ac.uk/bnc2014/} \cite{love_spoken_2022}. 

Academic writing is derived from the ACL Anthology using the S2ORC API \cite{lo_s2orc_2020}, which is available through the Semantic Scholar API at \url{https://www.semanticscholar.org/product/api}; we extract ACL main conference papers published between 2009 and 2018. For each paper, we analyze the introduction section and use the abstract as document-level metadata. 

Instructive online text is sourced from the wikiHow dataset (\url{https://www.wikihow.com/Main-Page}), using the publicly available summarization version at \url{https://www.kaggle.com/datasets/varunucl/wikihow-summarization}, where article bodies constitute the main text and titles and headlines are treated as metadata \cite{koupaee_wikihow_2018}. 

Creative writing data is taken from the Reddit Writing Prompts corpus (\url{https://www.reddit.com/r/WritingPrompts/}), using the dataset released at \url{https://www.kaggle.com/datasets/ratthachat/writing-prompts}, which consists of user-authored stories paired with prompts provided by other users \cite{fan_hierarchical_2018}. 

Finally, we use the XSum dataset, available at \url{https://huggingface.co/datasets/EdinburghNLP/xsum}, which contains BBC news articles (\url{https://www.bbc.com/}) paired with single-sentence summaries \cite{narayan_dont_2018}.

Since all datasets except the BNC2014Spoken dataset are online scraped datasets, we employ a string cleaning method to ensure consistent encodings. This involves replacing all linebreaks, indentation and multiple whitespace with single whitespace. Further, we remove all whitespaces before punctuations and normalize all text into Unicode Normalization Form Compatibility Composition.

To count lexical tokens, we use the Python package Spacy and the "en\_core\_web\_sm" model with disabled Named Entity Recognition. We count all tokens that are not punctuation or a space. To get the sentence boundaries, we use the sentences detected by Spacy.

For the following datasets, we employ special data preparation schemes:

\textbf{Writing Prompts}: Since the Writing Prompt dataset is directly scraped form a social media platform, text can contain unusual formats or characters. To catch this we calculate the punctuation-words ratio. We manually check the texts with the highest punctuation-words ratio and define a threshold to exclude them. We manually defined a threshold of 0.2. The excluded texts are published together with the results of this paper.  

\textbf{BNC2014Spoken}: The BNC2014Spoken corpus is anonymized, meaning that all names, places and other personal information are excluded from the dataset. While the dataset comes pre-annotated with POS-tags, we have to use the same Spacy model as for the other datasets, as a different POS-model will introduce a bias. Therefore, we replace all anonymized information where possible with matching pseudo information. The most common pseudonymized information is names. Therefore, we scrape the Wikipedia page for common English male\footnote{\url{https://en.wikipedia.org/wiki/Category:English_masculine_given_names}} and female\footnote{\url{https://en.wikipedia.org/wiki/Category:English_feminine_given_names}} names and insert them according to the provided metadata into the text, so gender distribution is kept. For other, less frequent personal such as places, telephone numbers, addresses, emails and social media names we use the Faker\footnote{\url{https://faker.readthedocs.io/en/master/}} library. 
\section{Classifier}
\label{sec:classifier}
To add further empirical evidence to the selection of the framework of Biber, we train a classifier to distinguish between human-written and \ac{LLM} generated texts solely by their linguistic features. Following the logic of Biber, registers are distinguishable by their co-occurrence patterns of linguistic features \cite{biber1988variation}. As these co-occurrence patterns are given as linear combinations of the features, we should be able to train a linear decision boundary to distinguish between the human and \ac{LLM} generated texts if the results of our study hold.

We construct our dataset by taking for all registers all human and \ac{LLM} examples, and randomly subsample the large class to the size of the smaller class. We train a classifier on each register and on the whole dataset. We use an 80/20 train-test split and fit a Logistic Regression model with a 5-fold cross-validation to the data.

The accuracy and ROC accuracy are shown in Table \ref{tab:ai_human_register_results}. For all registers, we achieve a ROC AUC > 0.9 on the test set. This indicates that the framework of Biber gives a useful feature set to detect \ac{LLM} generated texts and that \ac{LLM} generated and human-produced texts differ substantially under our tested generation setup. 
\begin{table}[t]
\centering
\small
\rowcolors{2}{gray!10}{white}
\setlength{\arrayrulewidth}{0.8pt}
\renewcommand{\arraystretch}{1.2}

\begin{tabular}{|>{\bfseries}l|c|c|}
\hline
\rowcolor{gray!25}
Register & Accuracy & ROC AUC \\
\hline
All Registers     & 0.881 & 0.949 \\
BNC2014Spoken     & 0.988 & 0.999 \\
S2ORC\_ACL        & 0.951 & 0.989 \\
wikiHow           & 0.920 & 0.981 \\
WritingPrompts    & 0.923 & 0.970 \\
XSum               & 0.954 & 0.992 \\
\hline
\end{tabular}

\caption{Classification performance of the AI vs.\ Human register classifier across datasets.}
\label{tab:ai_human_register_results}
\end{table}
\section{Biber framework and MMD}

While the framework of \citeauthor{biber1988variation} is common in corpus linguistics, the use of the \ac{MMD} is not. Therefore, we want to highlight in the following why the \ac{MMD} can be beneficial for register studies.

Taking the \ac{MMD} with an RBF-kernel as our distance metric, according to Theorem 5 in \cite{gretton_kernel_2012} the \ac{MMD} will only be 0 if and only if the underlying distributions are the same. Thus, a \ac{MMD} within the variability observed between humans indicates (among higher order relationships) that the underlying marginal feature distributions and co-occurrence patterns are within what we would expect between two human samples. In other words, if the observed \ac{MMD} is within the variability we would expect between human samples, this indicates that an analysis of the two samples would yield roughly the same results as analyzing two human samples.

\section{Marginal Feature differences}
\label{sec:appendix_features}
\begin{figure*}[p]
    \includegraphics[height=0.9\textheight]{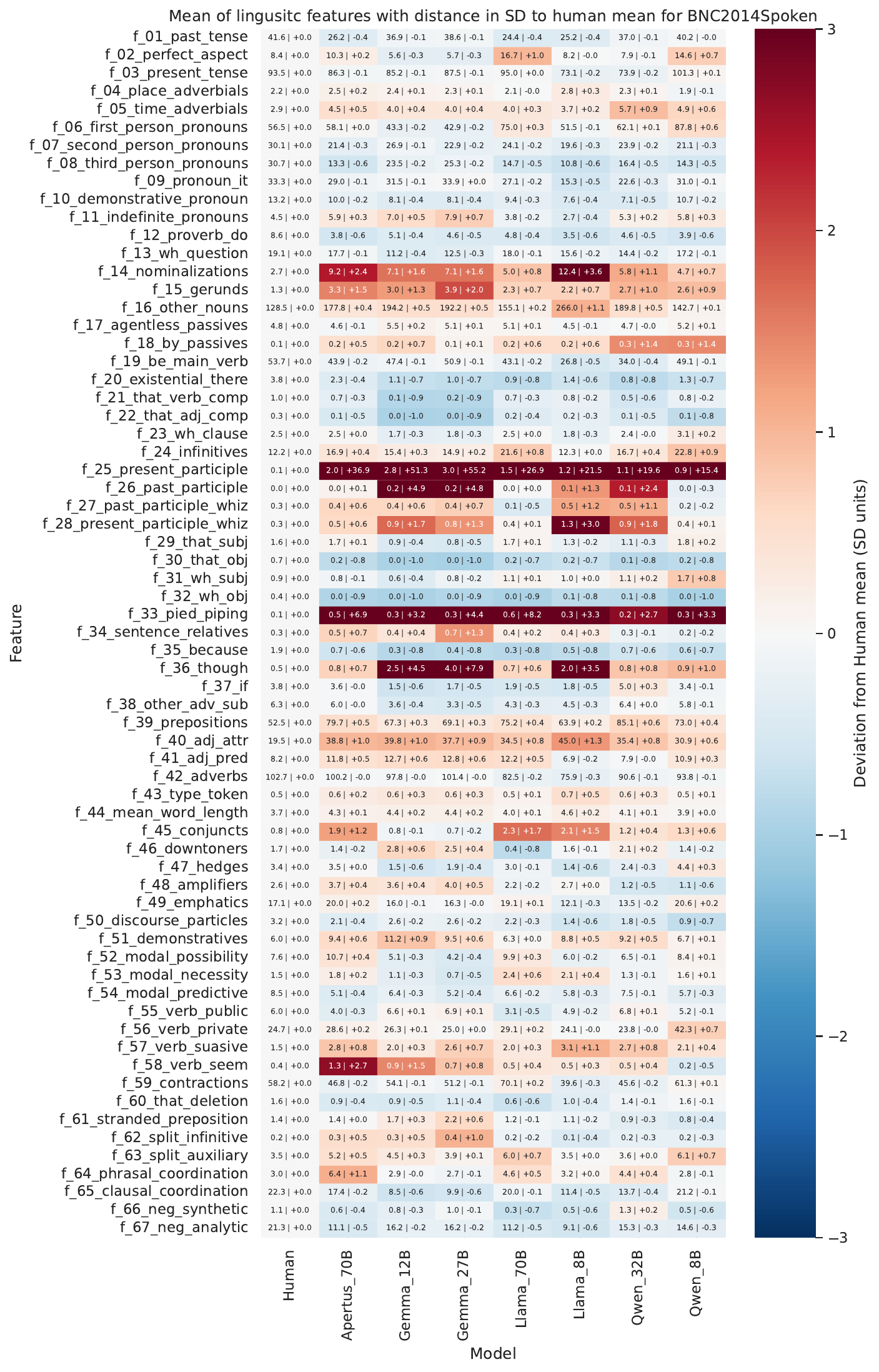}
    \caption{Mean of the normalized linguistic features without standardization to the full human dataset, with the difference in standard deviation to the mean of the human reference sample for the \textbf{BNC2014Spoken} in the Zero-Shot setting}
    \label{fig:AI_Feature_Level_Mean_BNC2014Spoken}
\end{figure*}

\begin{figure*}[p]
    \includegraphics[height=0.9\textheight]{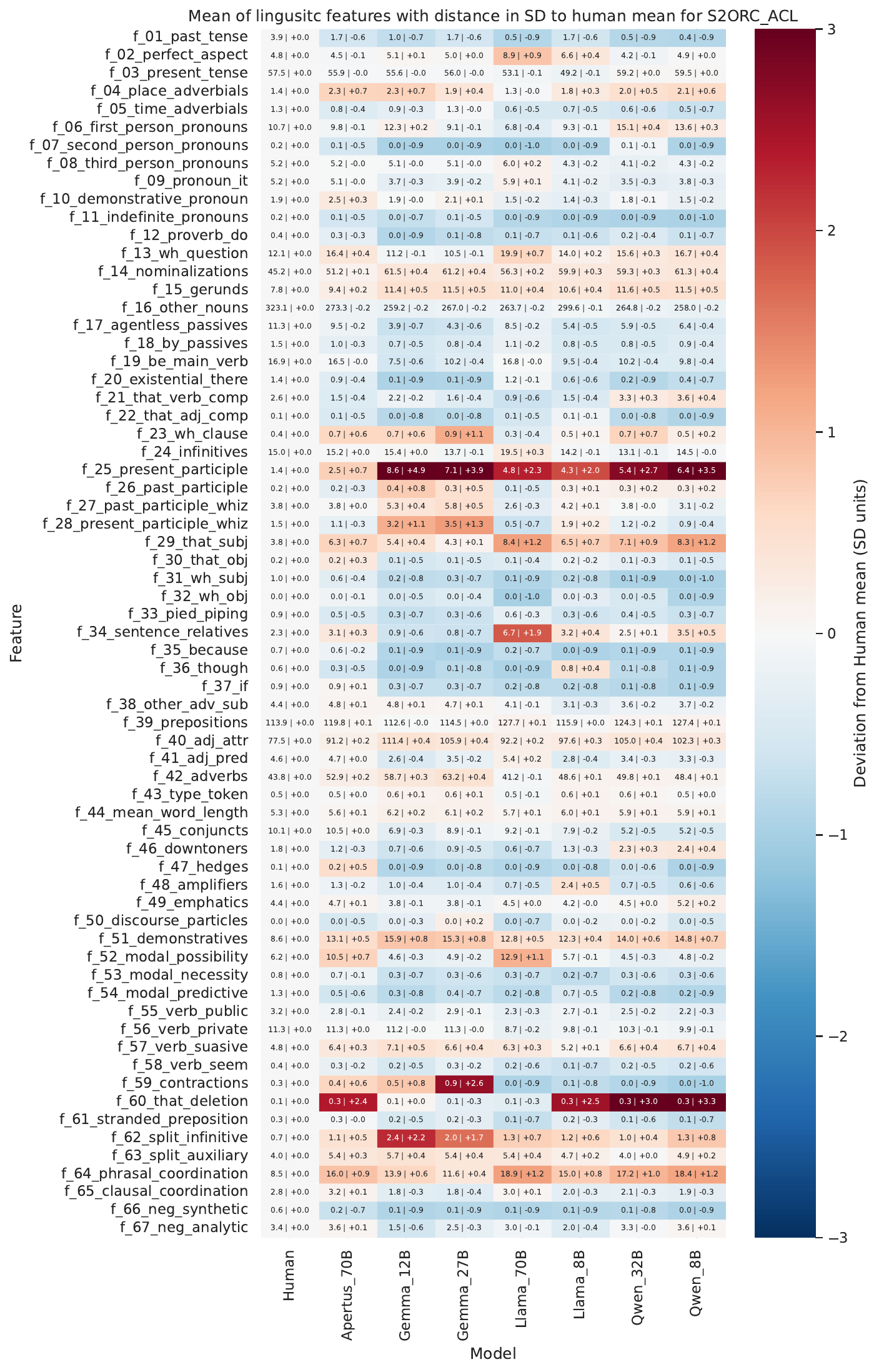}
    \caption{Mean of the normalized linguistic features without standardization to the full human dataset, with the difference in standard deviation to the mean of the human reference sample for the \textbf{S2ORC\_ACL} in the Zero-Shot setting}
    \label{fig:AI_Feature_Level_Mean_S2ORC_ACL}
\end{figure*}

\begin{figure*}[p]
    \includegraphics[height=0.9\textheight]{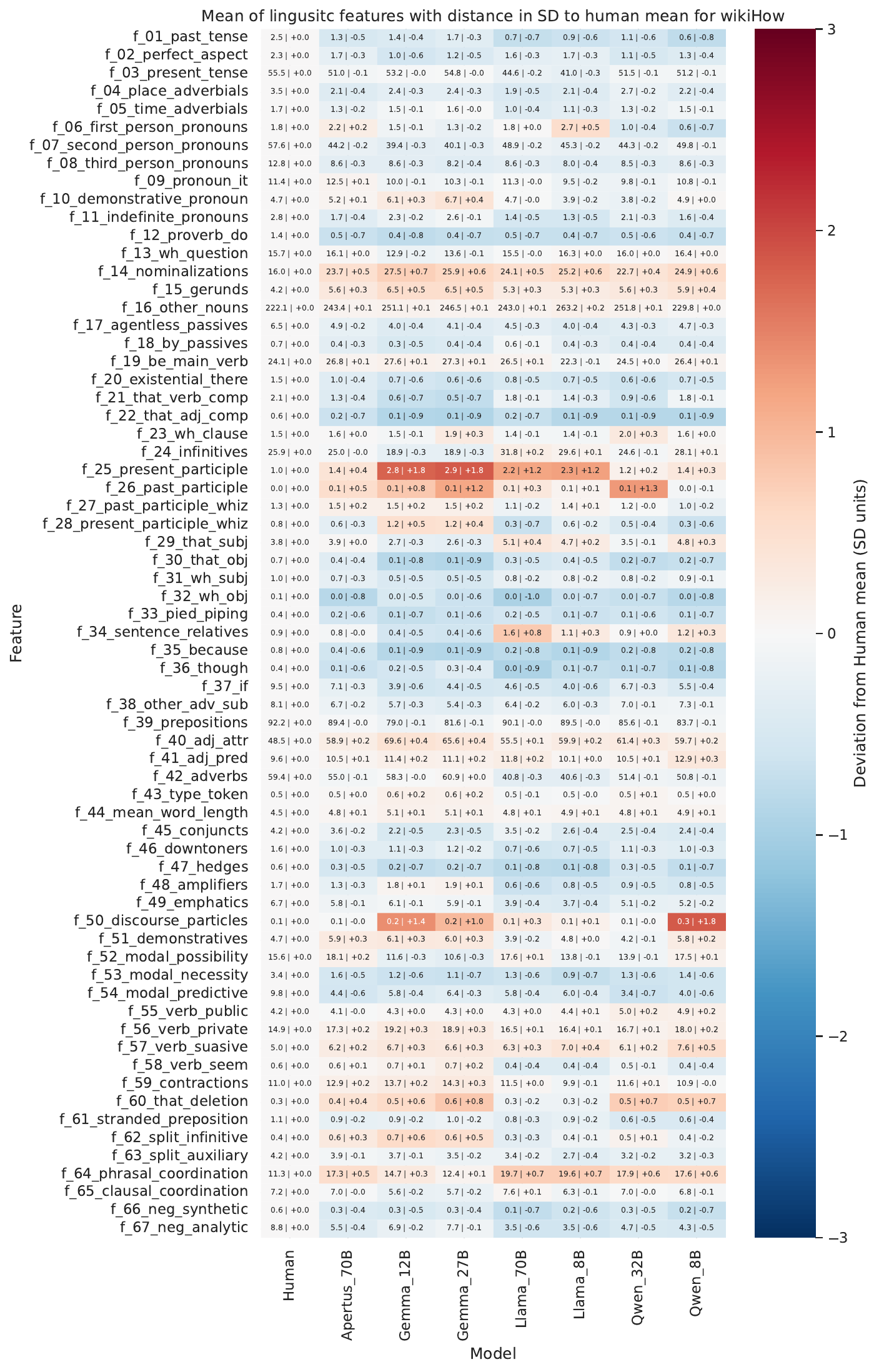}
    \caption{Mean of the normalized linguistic features without standardization to the full human dataset, with the difference in standard deviation to the mean of the human reference sample for the wikiHow in the Zero-Shot setting}
    \label{fig:AI_Feature_Level_Mean_wikiHow}
\end{figure*}

\begin{figure*}[p]
    \includegraphics[height=0.9\textheight]{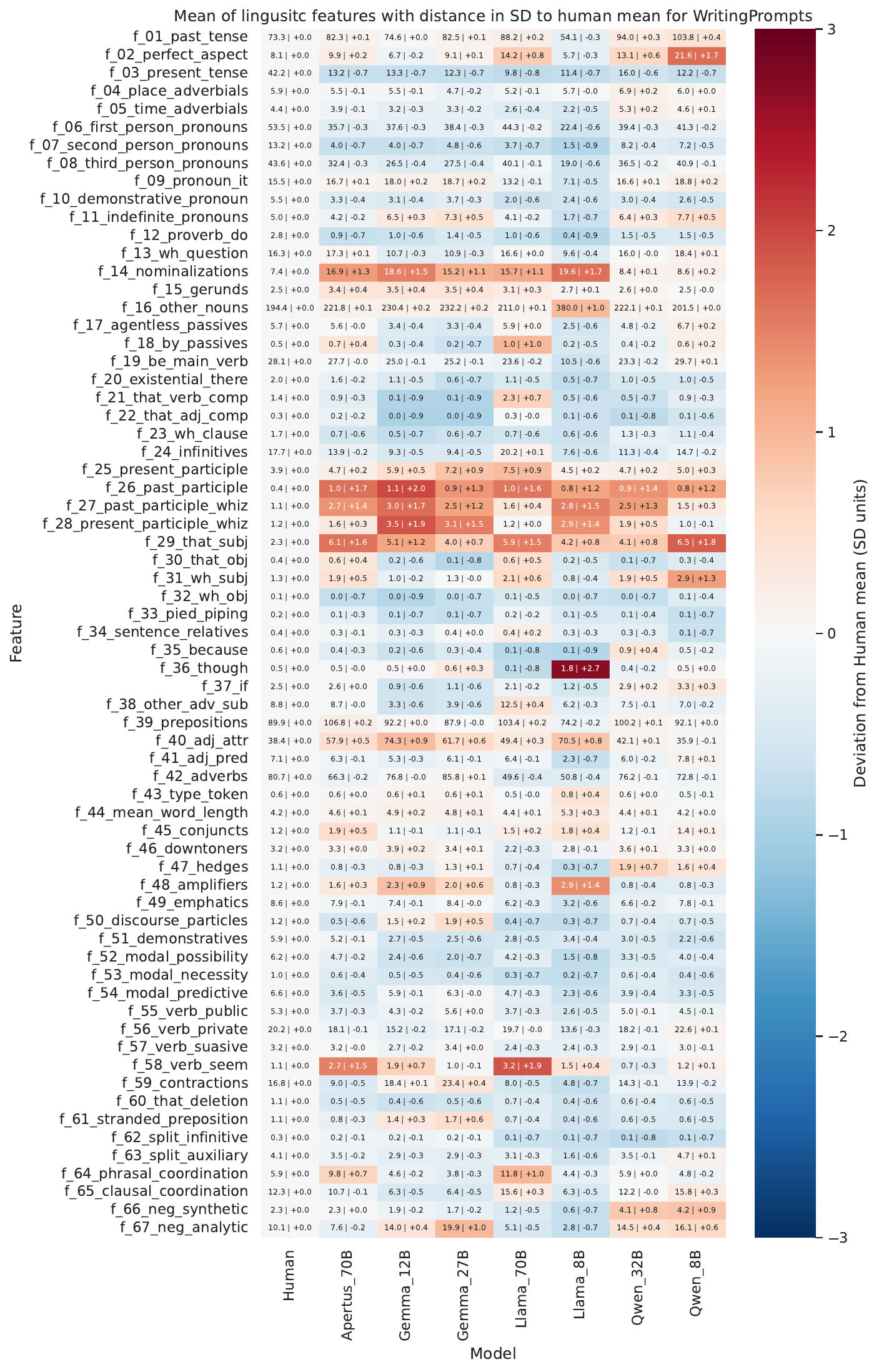}
    \caption{Mean of the normalized linguistic features without standardization to the full human dataset, with the difference in standard deviation to the mean of the human reference sample for the \textbf{WritingPrompts} in the Zero-Shot setting}
    \label{fig:AI_Feature_Level_Mean_WritingPrompts}
\end{figure*}

\begin{figure*}[p]
    \includegraphics[height=0.9\textheight]{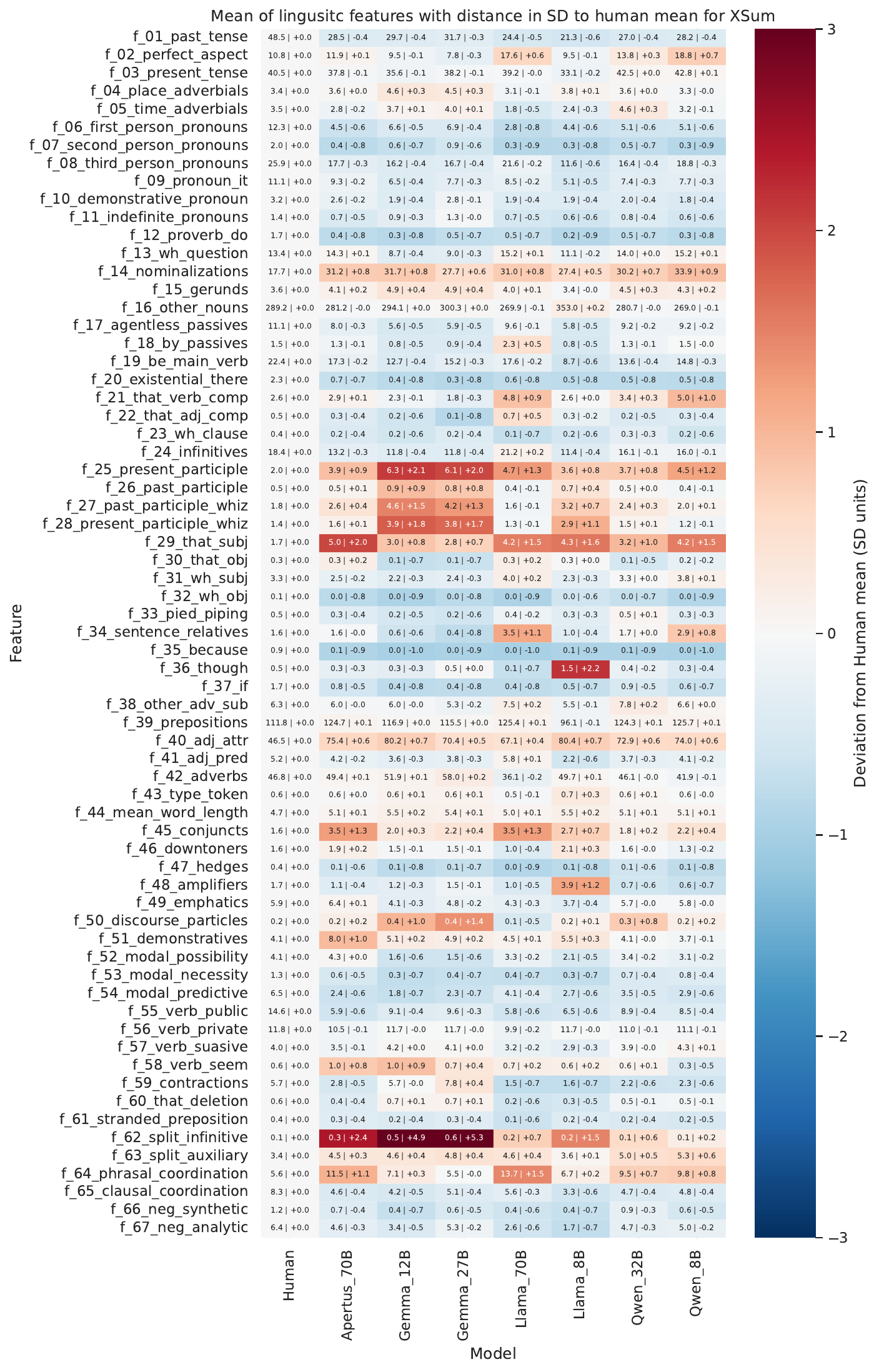}
    \caption{Mean of the normalized linguistic features without standardization to the full human dataset, with the difference in standard deviation to the mean of the human reference sample for the \textbf{XSum} in the Zero-Shot setting}
    \label{fig:AI_Feature_Level_Mean_XSum}
\end{figure*}

\begin{figure*}[p]
    \includegraphics[height=0.9\textheight]{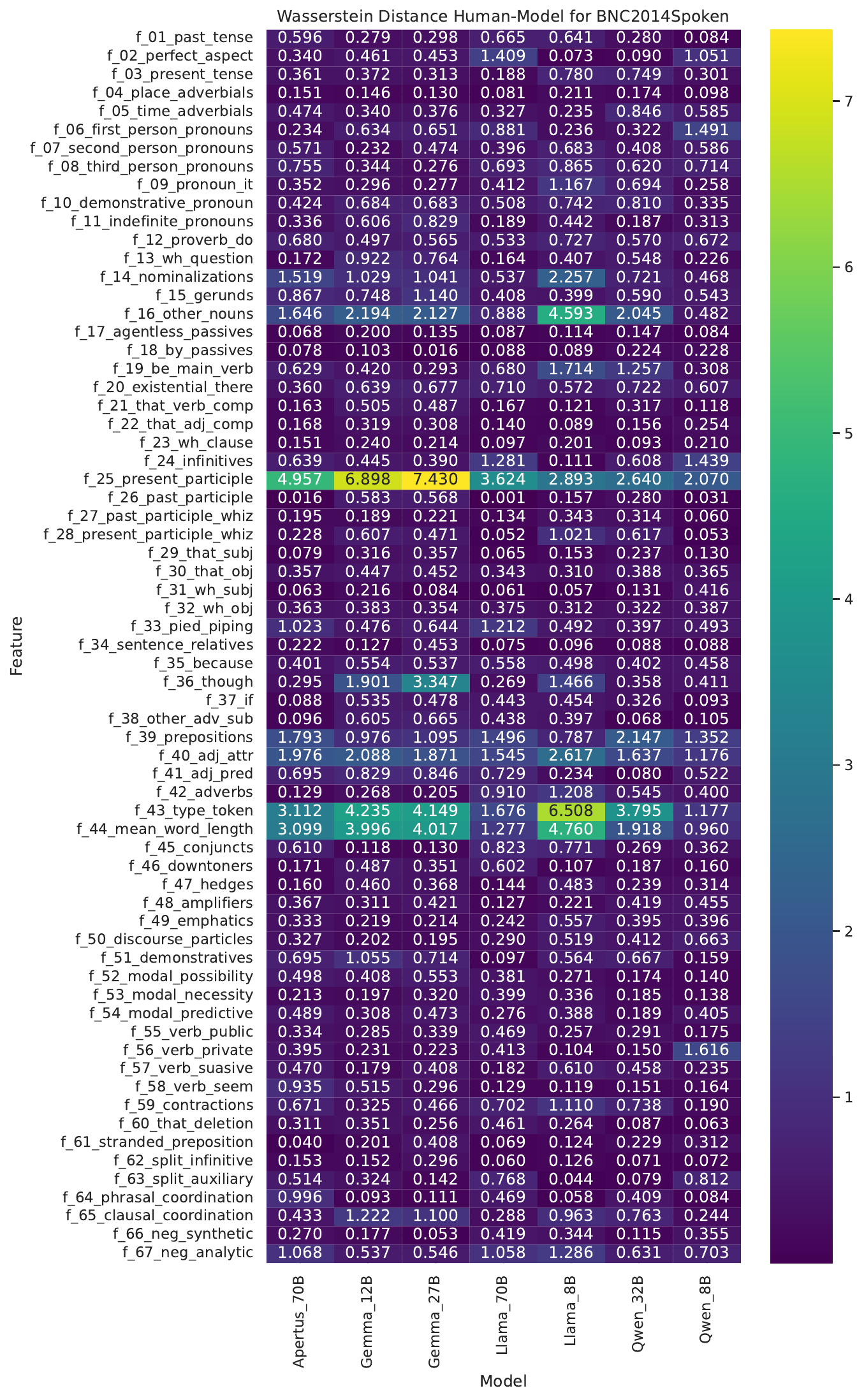}
    \caption{Wasserstein distance for marginal feature distributions between model and human for \textbf{BNC2014Spoken} in the Zero-Shot setting}
    \label{fig:AI_Feature_Level_BNC2014Spoken}
\end{figure*}

\begin{figure*}[p]
    \includegraphics[height=0.9\textheight]{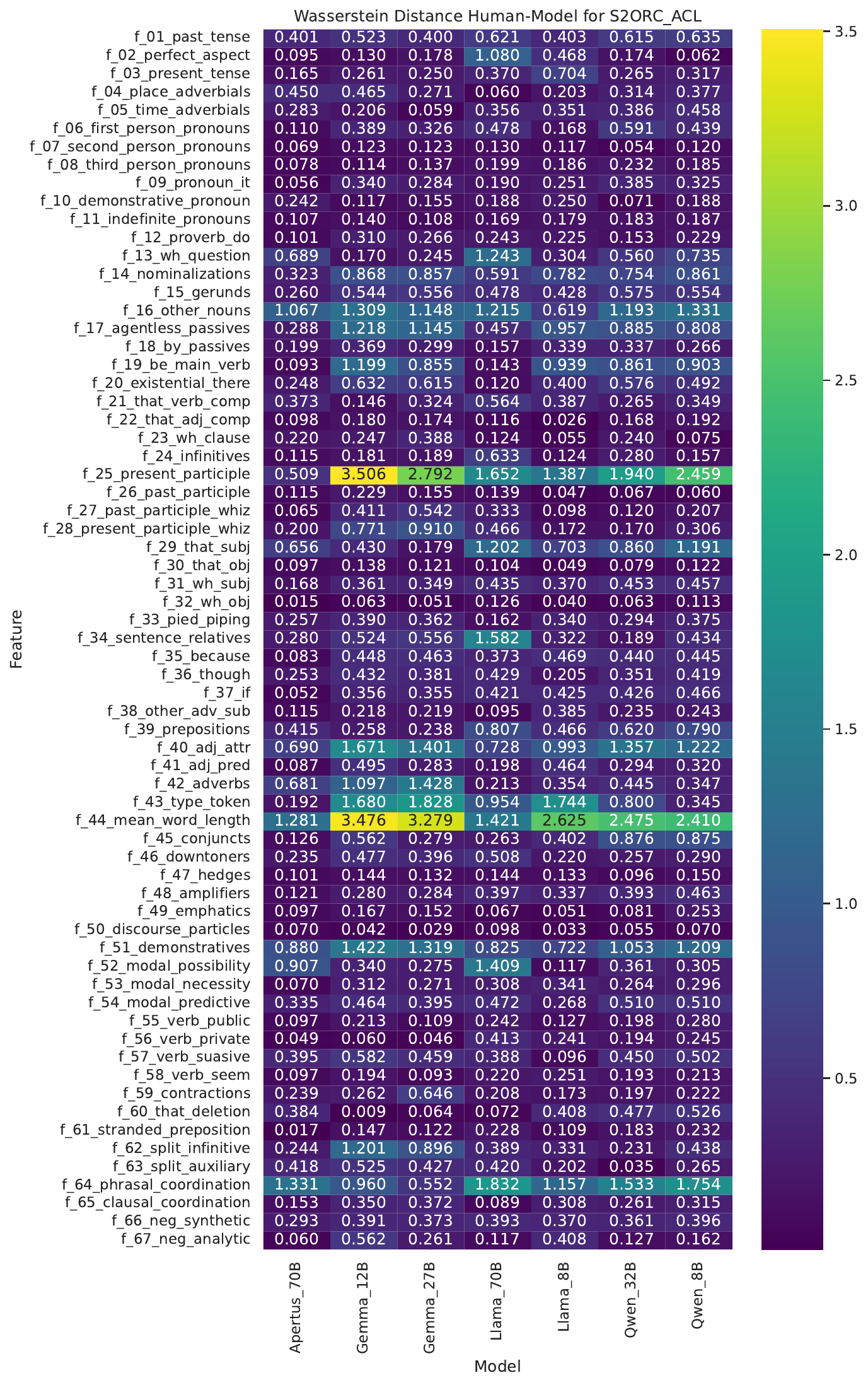}
    \caption{Wasserstein distance for marginal feature distributions between model and human for \textbf{S2ORC\_ACL} in the Zero-Shot setting}
    \label{fig:AI_Feature_Level_S2ORC_ACL}
\end{figure*}

\begin{figure*}[p]
    \includegraphics[height=0.9\textheight]{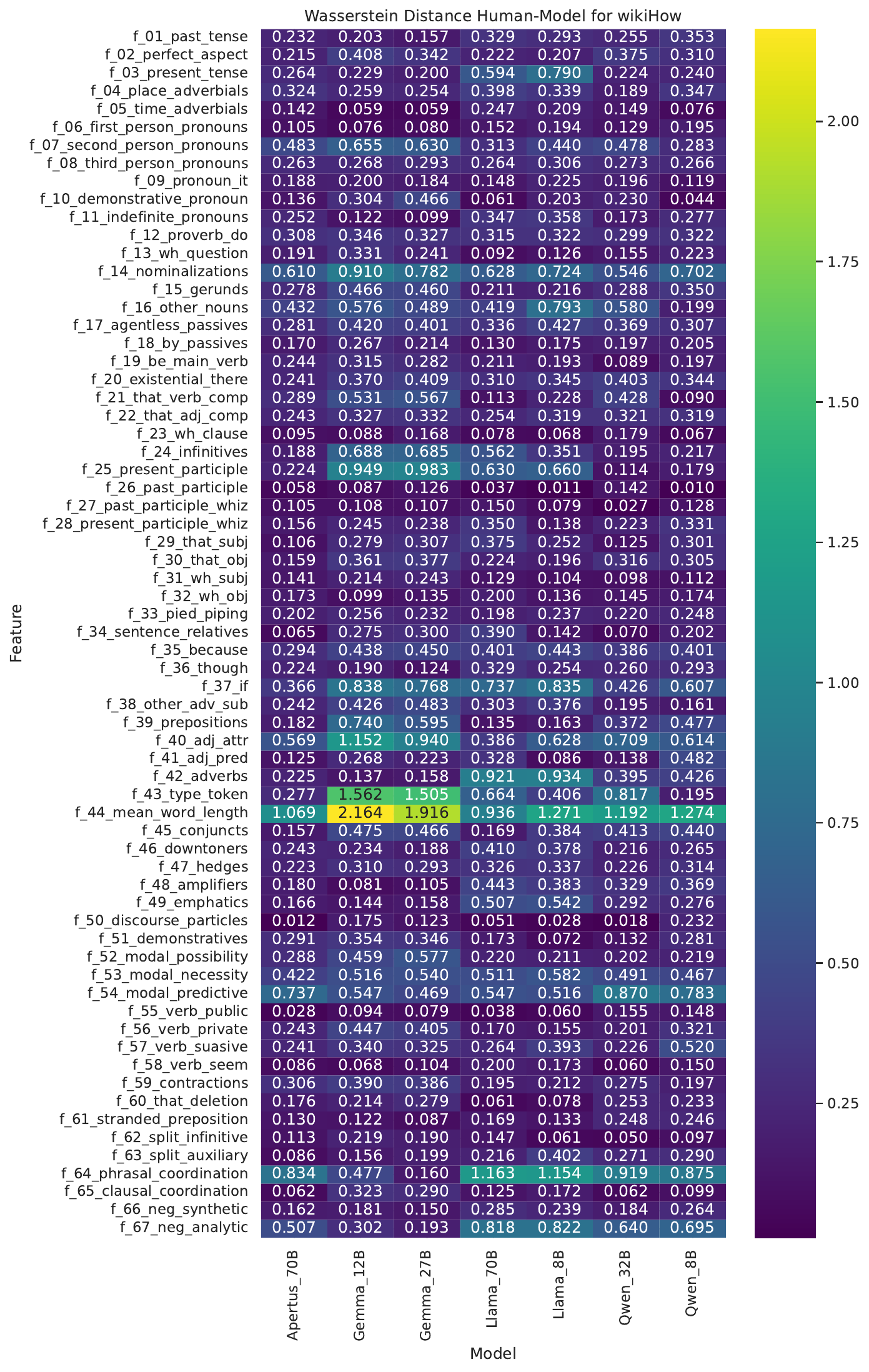}
    \caption{Wasserstein distance for marginal feature distributions between model and human for \textbf{wikiHow} in the Zero-Shot setting}
    \label{fig:AI_Feature_Level_wikiHow}
\end{figure*}

\begin{figure*}[p]
    \includegraphics[height=0.9\textheight]{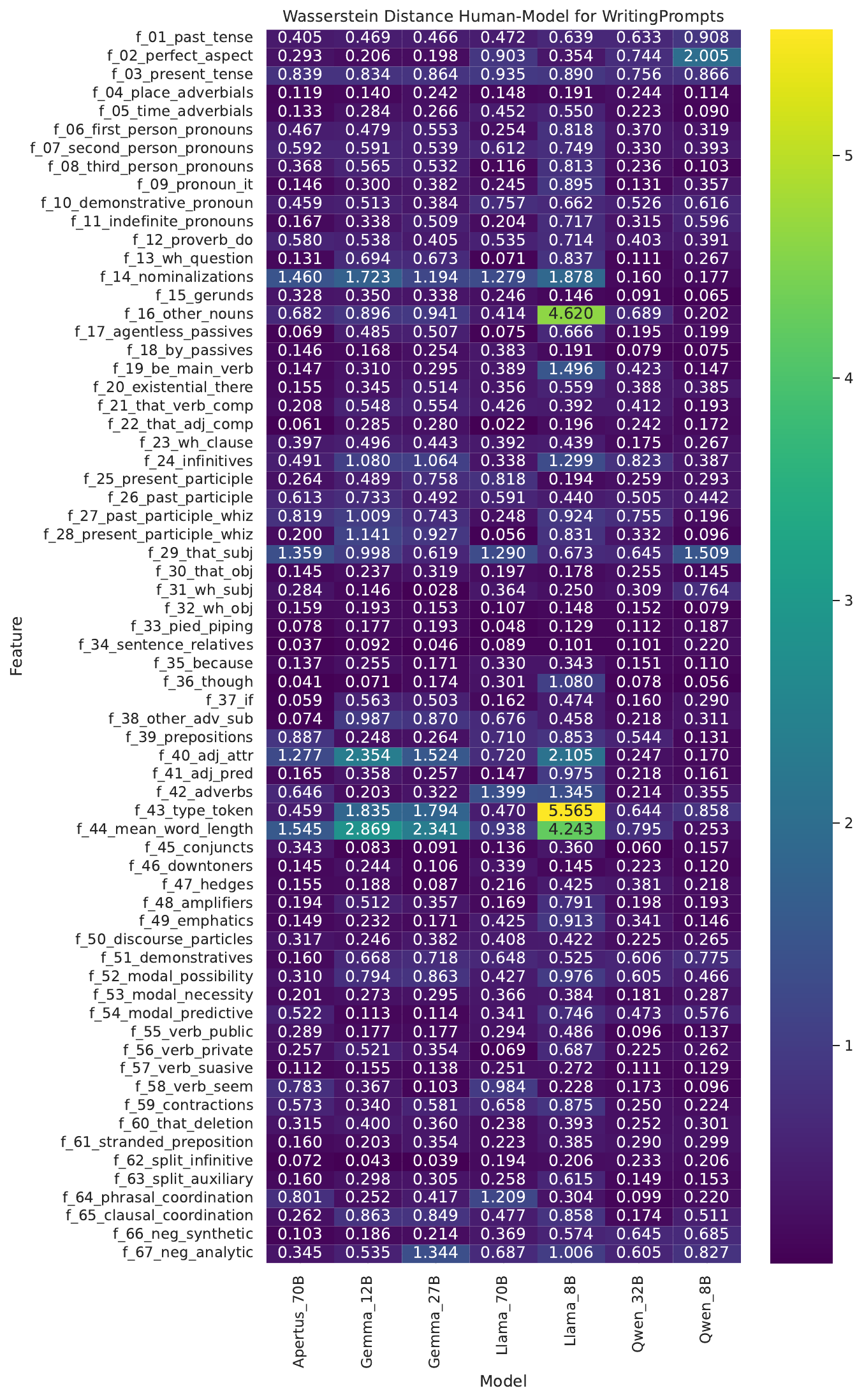}
    \caption{Wasserstein distance for marginal feature distributions between model and human for \textbf{WritingPrompts} in the Zero-Shot setting}
    \label{fig:AI_Feature_Level_WritingPrompts}
\end{figure*}

\begin{figure*}[p]
    \includegraphics[height=0.9\textheight]{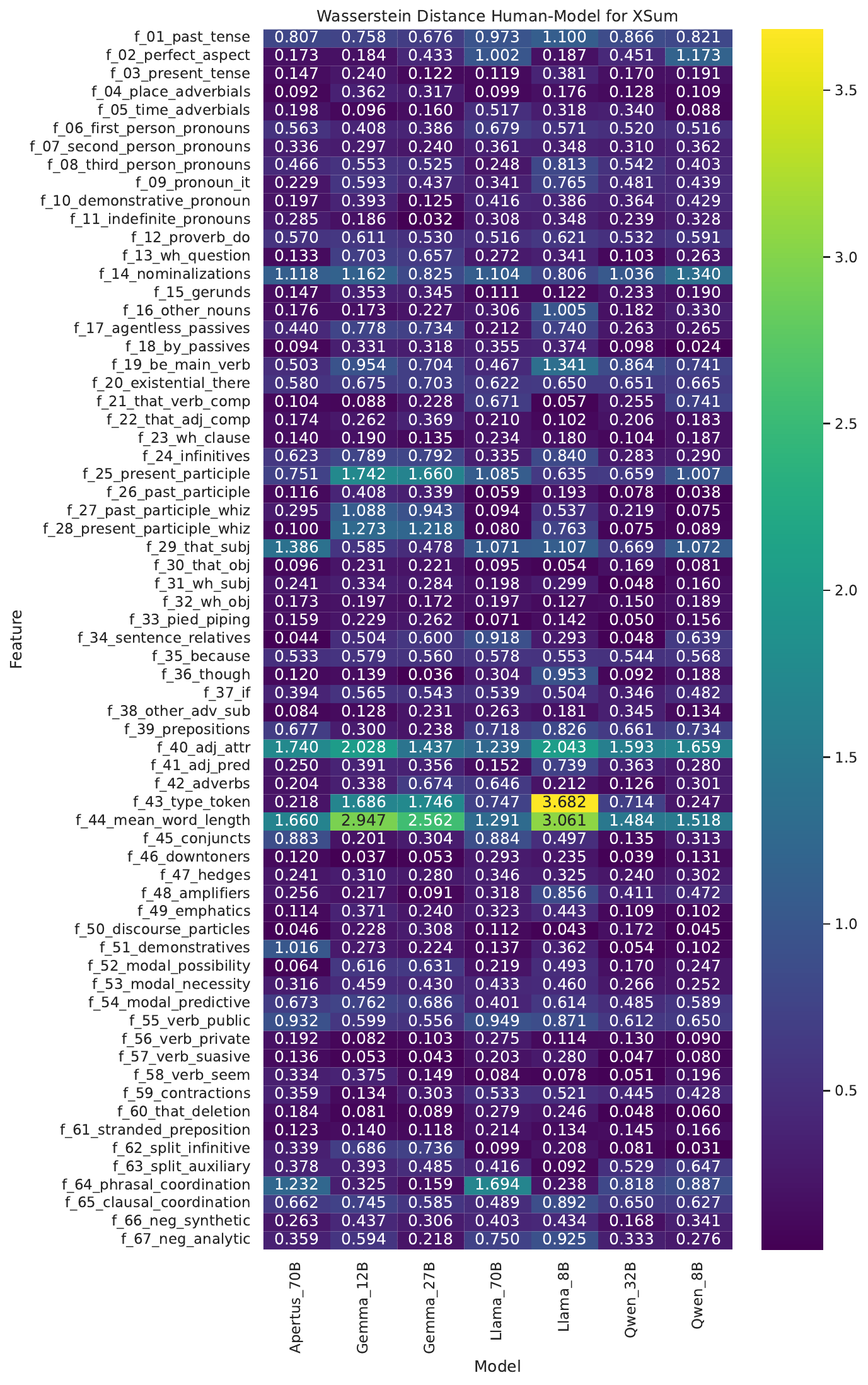}
    \caption{Wasserstein distance for marginal feature distributions between model and human for \textbf{XSum} in the Zero-Shot setting}
    \label{fig:AI_Feature_Level_XSum}
\end{figure*}

To analyse the difference for the marginal feature distributions, we use the following plots. Figures \ref{fig:AI_Feature_Level_Mean_BNC2014Spoken}, \ref{fig:AI_Feature_Level_Mean_S2ORC_ACL}, \ref{fig:AI_Feature_Level_Mean_wikiHow}, \ref{fig:AI_Feature_Level_Mean_WritingPrompts} and \ref{fig:AI_Feature_Level_Mean_XSum} show the mean of the each feature for each model together with the human sample. The heatmap highlights the difference in standard deviations of the human reference for this feature. Further, in Figures \ref{fig:AI_Feature_Level_BNC2014Spoken}, \ref{fig:AI_Feature_Level_S2ORC_ACL}, \ref{fig:AI_Feature_Level_wikiHow}, \ref{fig:AI_Feature_Level_WritingPrompts} and \ref{fig:AI_Feature_Level_XSum} we show the Wasserstein distance of the marginal feature distribution to the human reference sample both standardized with the full human sample. The difference in means is a good way to visualize differences in a human-interpretable way, while the Wasserstein distance is suitable to highlight differences in distributions. 

A comparison between a human text and three models for one writing prompt is shown in Table \ref{tab:qualitative_example}. Annotated are the features discussed in Section \ref{sec:features}. In the text of Gemma 12B appear two nominalizations and 10 attributive adjectives compared to zero nominalizations and two attributive adjectives in the human text. While the use of nouns and past tense is not too different from the human samples, all three generated samples use present participle clauses, while none occurs in the human sample.

\section{Using the Framework to Benchmark New Models}
\label{app:benchmark}

The experiment setup used in this work can be directly repeated to evaluate any new \ac{LLM} against the models reported in this study. To do this, we publish the code used in this study, the prompt templates for each register, the per-register human reference subsamples, the per-register bandwidth values for the \ac{MMD}, and the per-register feature means and standard deviations used for standardization. This appendix outlines the procedure for evaluating a new model.

\textbf{Procedure:} Given a new model with parameters $\theta'$ that one wishes to evaluate on a register $r$:
\begin{enumerate}
    \item Use the published prompt template $p_r(m_i)$ to sample a synthetic corpus $\tilde{C}_{r,\theta'} = \{(\tilde{t}_i, m_i)\}_{i=1}^{N_r}$ from the new model, conditioning on the same metadata items $m_i$ used in this study.
    \item Extract the 67 linguistic features of \cite{biber1988variation} for every text in $\tilde{C}_{r,\theta'}$, mapping each text to a vector in $\mathbb{R}^{67}$.
    \item Standardize the resulting feature vectors using the per-register means and standard deviations published with this study.
    \item Compute the \ac{MMD} + \ac{CI} between the new model's standardized synthetic corpus and the published human subsample for register $r$, using the published bandwidth parameter and an RBF kernel.
    \item Compare the resulting \ac{MMD} value against the values reported in Figure~\ref{fig:MMD_ranking}.
\end{enumerate}

Because both the human subsample and the bandwidth parameter for each register are fixed, the computed \ac{MMD}-values are directly comparable to the ones in this study. The results in this paper were obtained with temperature $1$ and top-$p$ $1$, i.e.\ sampling the full distribution induced by the model. Other sampling parameters can be used to study the influence different sampling parameter have on human-likeness as studied here.

If a researcher wishes to evaluate models on a register not covered by this study, both the human reference subsample and the bandwidth parameter must be re-estimated for that register following the procedure in §3.4 (sample-size selection) and §3.3 (median-heuristic bandwidth). The code to do this is published with this paper (\url{https://github.com/BjoernNieth/Register_Aware_LLMs}).

\section{Model details}
\label{sec:model_details}
\begin{sidewaystable*}[t]
\centering
\small
\setlength{\tabcolsep}{6pt}
\renewcommand{\arraystretch}{1.15}
\begin{tabular}{lcccccccccc}
\toprule
 & \multicolumn{3}{c}{\textbf{Architecture}} 
 & \multicolumn{5}{c}{\textbf{Training \& Alignment}} 
 & \multicolumn{2}{c}{\textbf{Training Data Characteristics}} \\
\cmidrule(lr){2-4} \cmidrule(lr){5-9} \cmidrule(lr){10-11}
\textbf{Model} &
\textbf{Params} &
\textbf{Context} &
\textbf{Activation} &
\textbf{Pretraining Loss} &
\textbf{Optimizer} &
\textbf{Instr.} &
\textbf{Align.} &
\textbf{Distill.} &
\textbf{Pretraining} &
\textbf{Type}  \\
\midrule
Apertus-70B      
& 70B & 65k  & xIELU   
& \loss{Goldfish Loss} 
& AdEMAMix 
& SFT 
& QRPO 
& None 
& 15T     
& Public online data \\

LLaMA-3.1-8B     
& 8B  & 128k & SwiGLU  
& NLL                 
& AdamW    
& SFT 
& DPO + NLL   
& None
& 15T  
& Public online data \\

LLaMA-3.3-70B    
& 70B & 128k & SwiGLU  
& NLL                 
& AdamW    
& SFT 
& DPO + NLL   
& None
& 15T  
& Public online data \\

Qwen-3-8B        
& 8B  & 32k  & SwiGLU  
& NLL                 
& ?        
& SFT 
& Mixed       
& Teacher LLM
& 36T  
& ``Various domains'' + synthetic data \\

Qwen-3-32B       
& 32B & 32k  & SwiGLU  
& NLL                 
& ?        
& SFT 
& Mixed       
& Teacher LLM
& 36T  
& ``Various domains'' + synthetic data \\

Gemma-3-12B      
& 12B & 128k & GELU    
& ?                 
& ?    
& SFT 
& DPO           
& Teacher LLM
& 12T  
& ? \\

Gemma-3-27B      
& 27B & 128k & GELU    
& ?                 
& ?    
& SFT 
& DPO           
& Teacher LLM
& 14T  
& ? \\
\bottomrule
\end{tabular}
\caption{
Model characteristics for all evaluated LLMs. The following acronyms are used: Supervised Fine-tuning (SFT), Quantile Reward Policy Optimization (QRPO), Direct Preference Optimization (DPO) and  negative log-likelihood (NLL)
}
\label{tab:model_metadata}
\end{sidewaystable*}
A detailed overview of the differences between the employed model families is shown in Table \ref{tab:model_metadata}.

\section{Biber Dimensions}

Table \ref{tab:biber1988-loadings} details the list of Biber framework features and their weights on the six dimensions.

\begin{figure*}[p]
\centering
\begin{subfigure}{0.49\linewidth}
  \includegraphics[width=\linewidth]{figures/Biber_Dimensions/Human_AI_BNC2014Spoken_Dimension1_violinplot.pdf}
  \caption{Dimension 1}
\end{subfigure}\hfill
\begin{subfigure}{0.49\linewidth}
  \includegraphics[width=\linewidth]{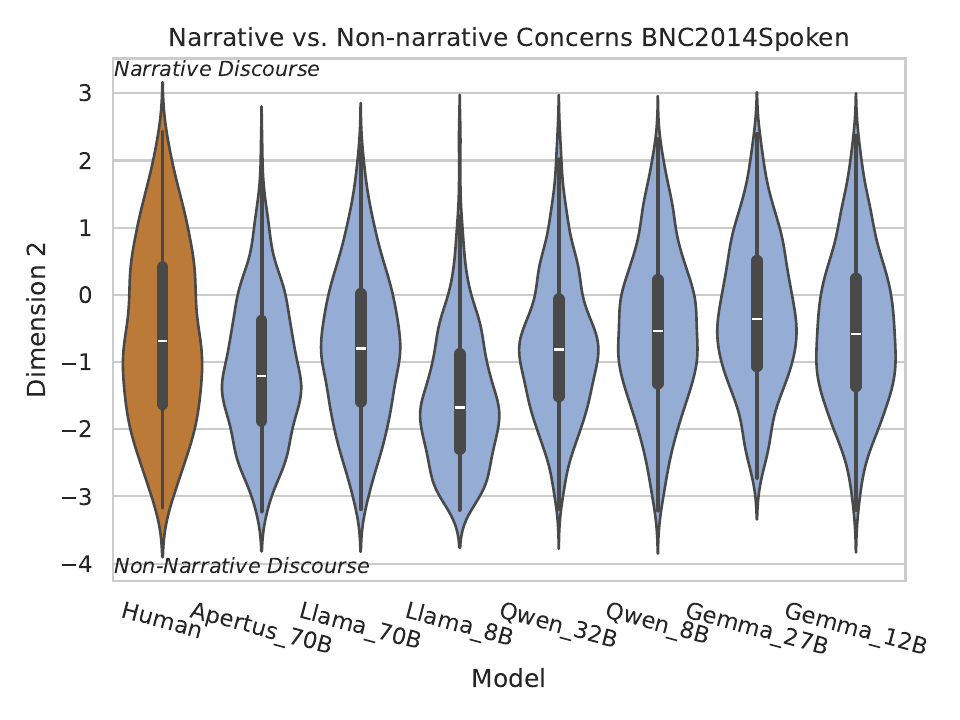}
  \caption{Dimension 2}
\end{subfigure}

\begin{subfigure}{0.49\linewidth}
  \includegraphics[width=\linewidth]{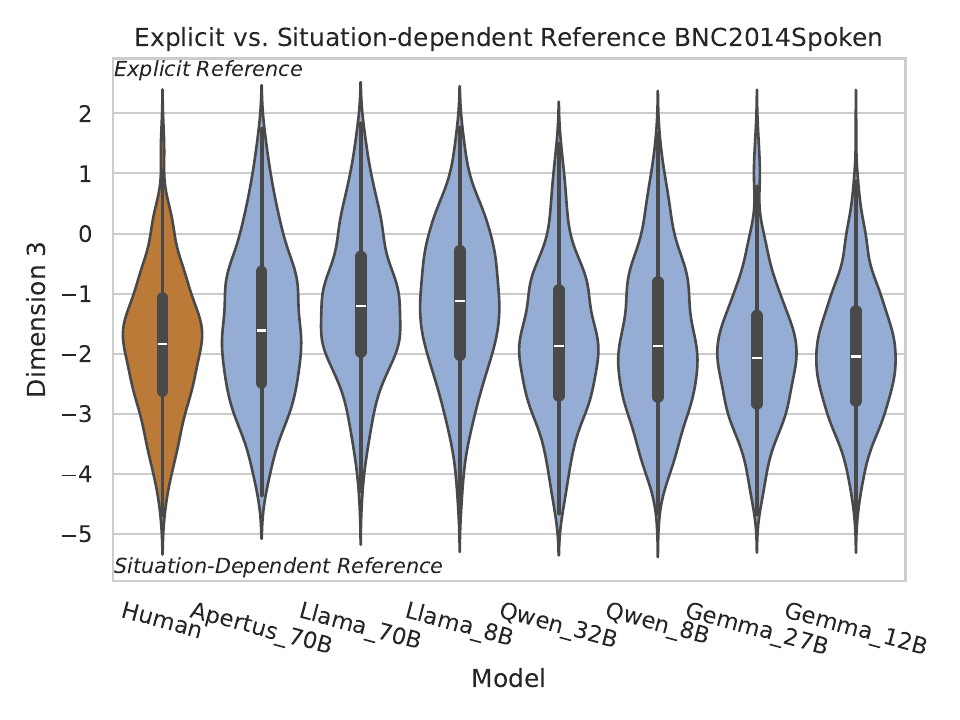}
  \caption{Dimension 3}
\end{subfigure}\hfill
\begin{subfigure}{0.49\linewidth}
  \includegraphics[width=\linewidth]{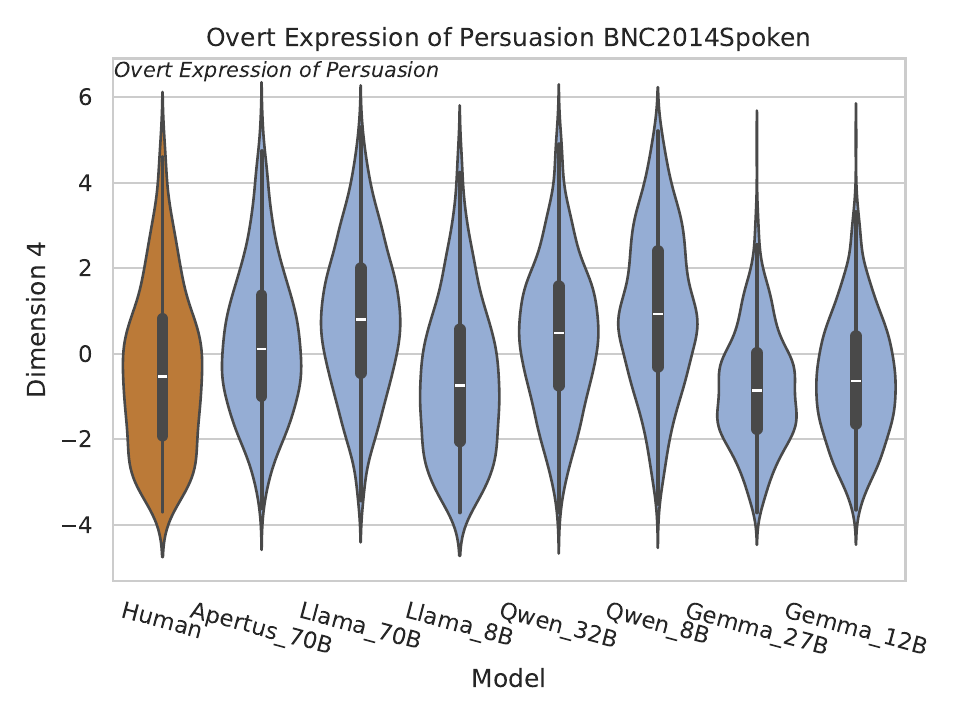}
  \caption{Dimension 4}
\end{subfigure}

\begin{subfigure}{0.49\linewidth}
  \includegraphics[width=\linewidth]{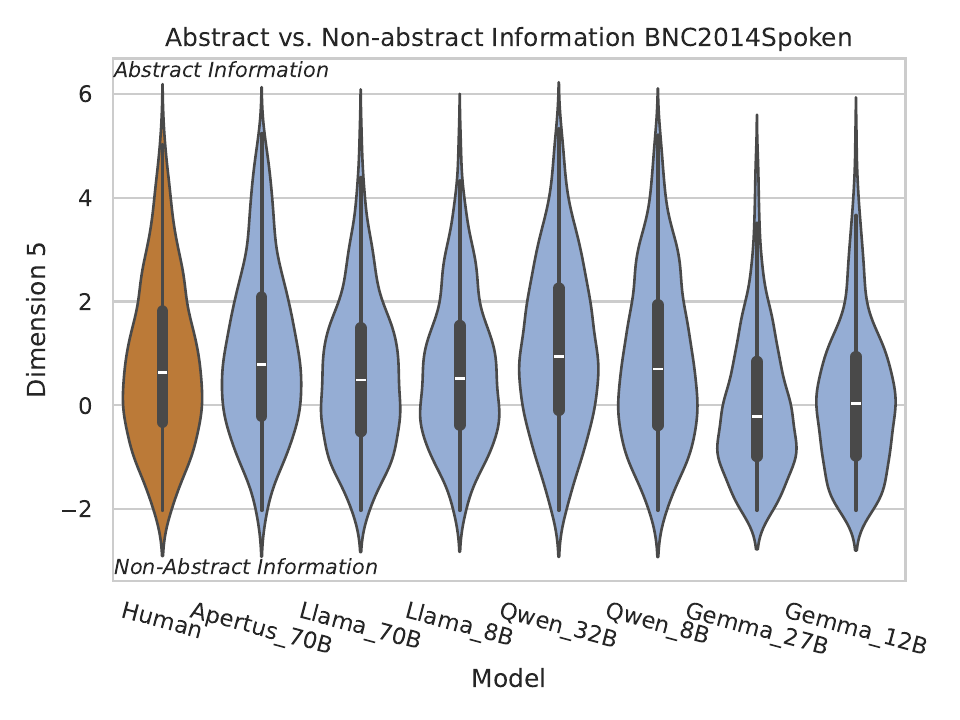}
  \caption{Dimension 5}
\end{subfigure}\hfill
\begin{subfigure}{0.49\linewidth}
  \includegraphics[width=\linewidth]{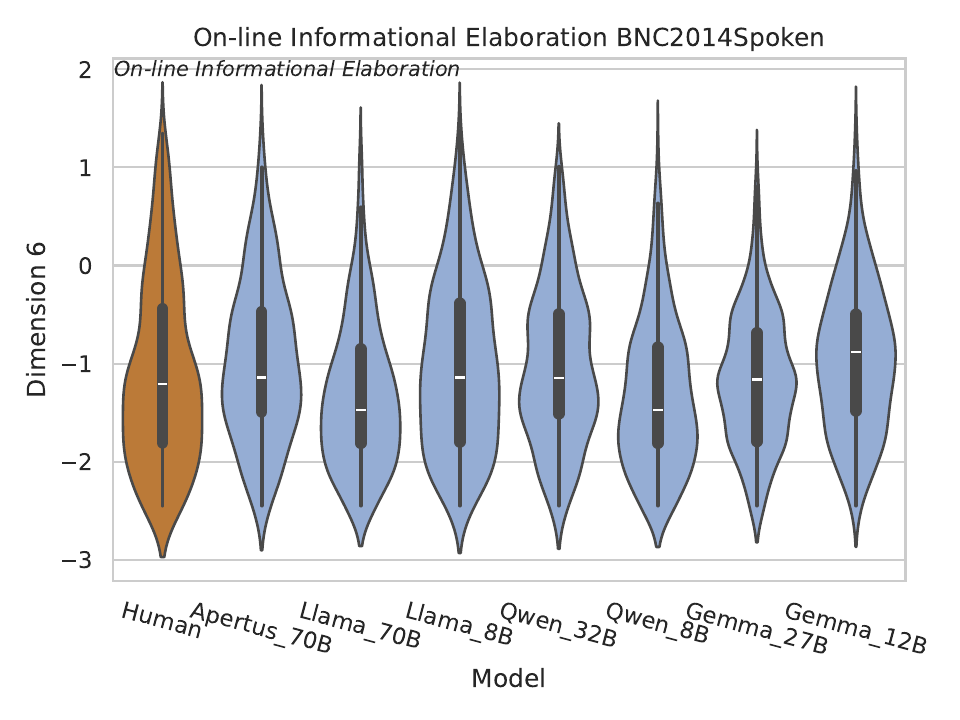}
  \caption{Dimension 6}
\end{subfigure}

\caption{Human and model distributions for Biber dimensions in the Zero-Shot setting (\textbf{BNC2014Spoken}).}
\label{fig:app:biber-bnc2014spoken}
\end{figure*}

\begin{figure*}[p]
\centering
\begin{subfigure}{0.49\linewidth}
  \includegraphics[width=\linewidth]{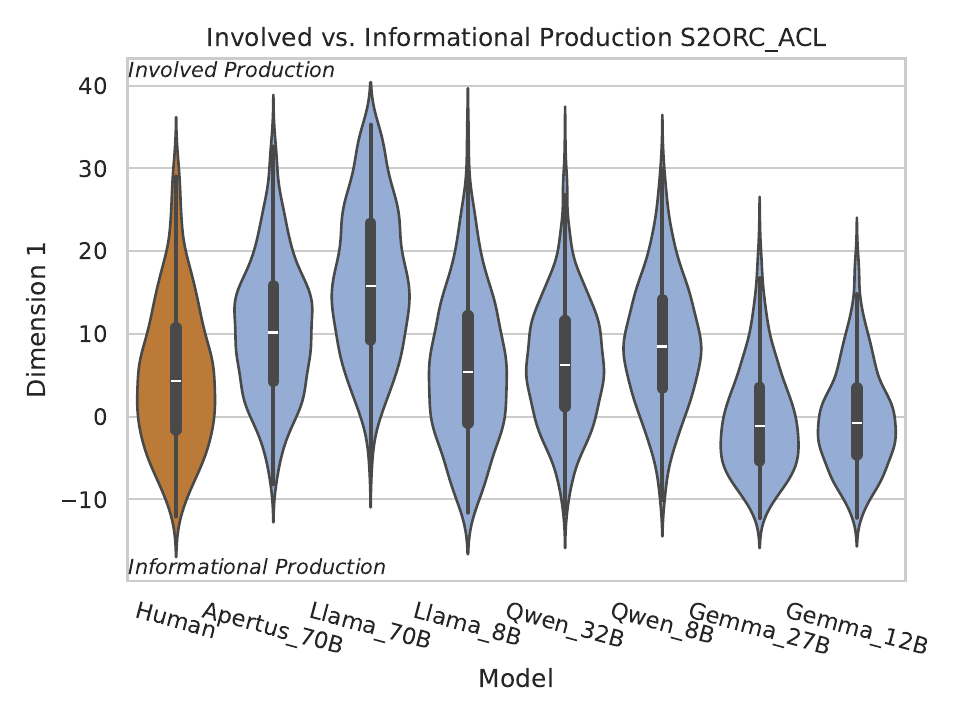}
  \caption{Dimension 1}
\end{subfigure}\hfill
\begin{subfigure}{0.49\linewidth}
  \includegraphics[width=\linewidth]{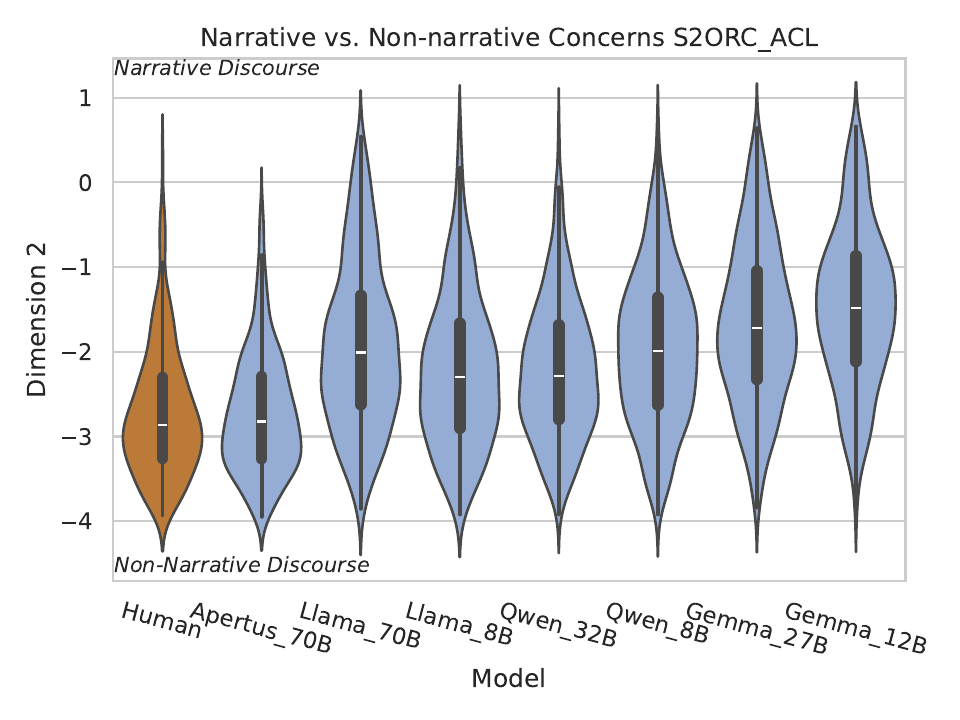}
  \caption{Dimension 2}
\end{subfigure}

\begin{subfigure}{0.49\linewidth}
  \includegraphics[width=\linewidth]{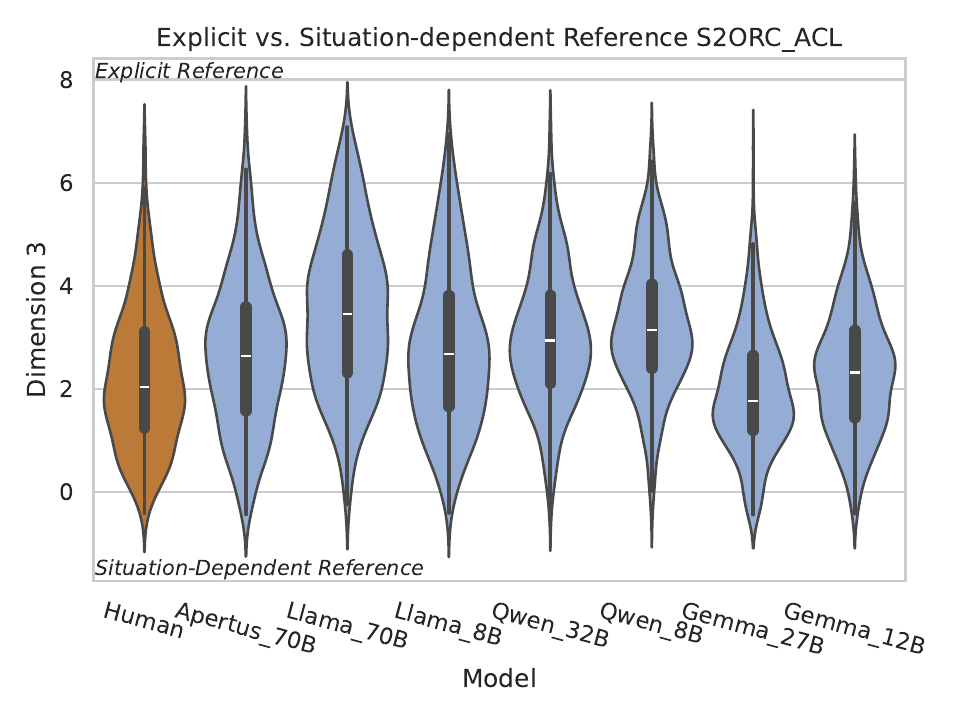}
  \caption{Dimension 3}
\end{subfigure}\hfill
\begin{subfigure}{0.49\linewidth}
  \includegraphics[width=\linewidth]{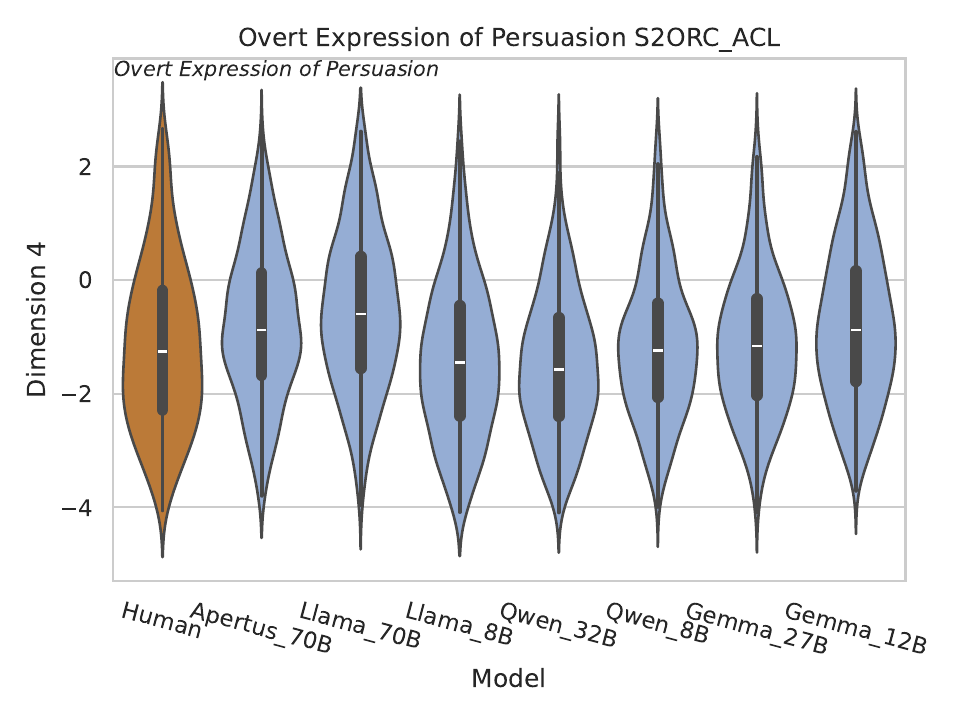}
  \caption{Dimension 4}
\end{subfigure}

\begin{subfigure}{0.49\linewidth}
  \includegraphics[width=\linewidth]{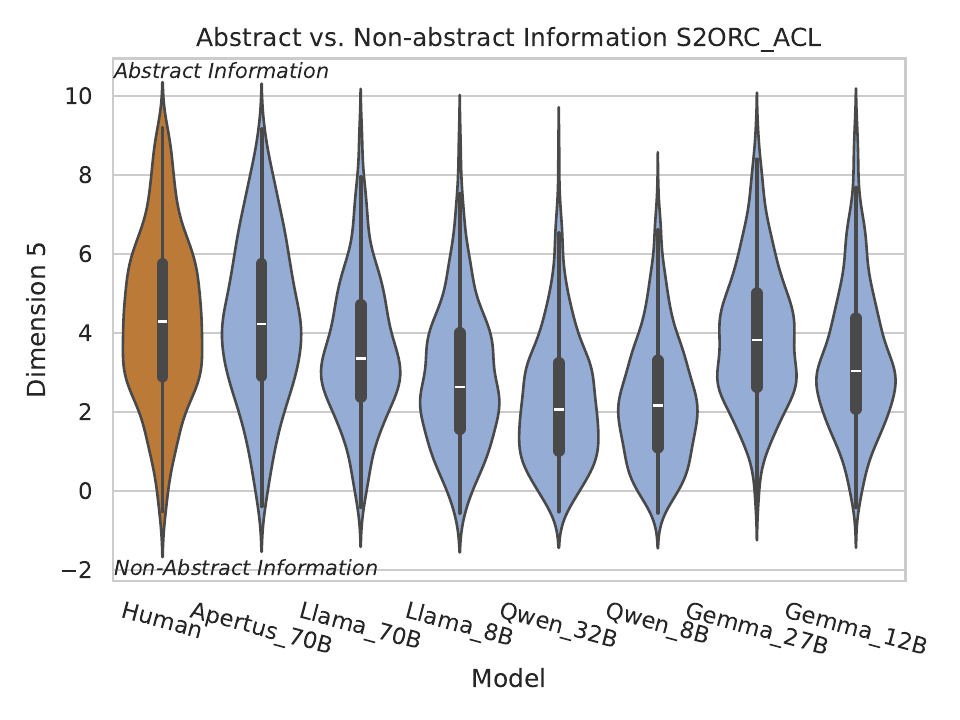}
  \caption{Dimension 5}
\end{subfigure}\hfill
\begin{subfigure}{0.49\linewidth}
  \includegraphics[width=\linewidth]{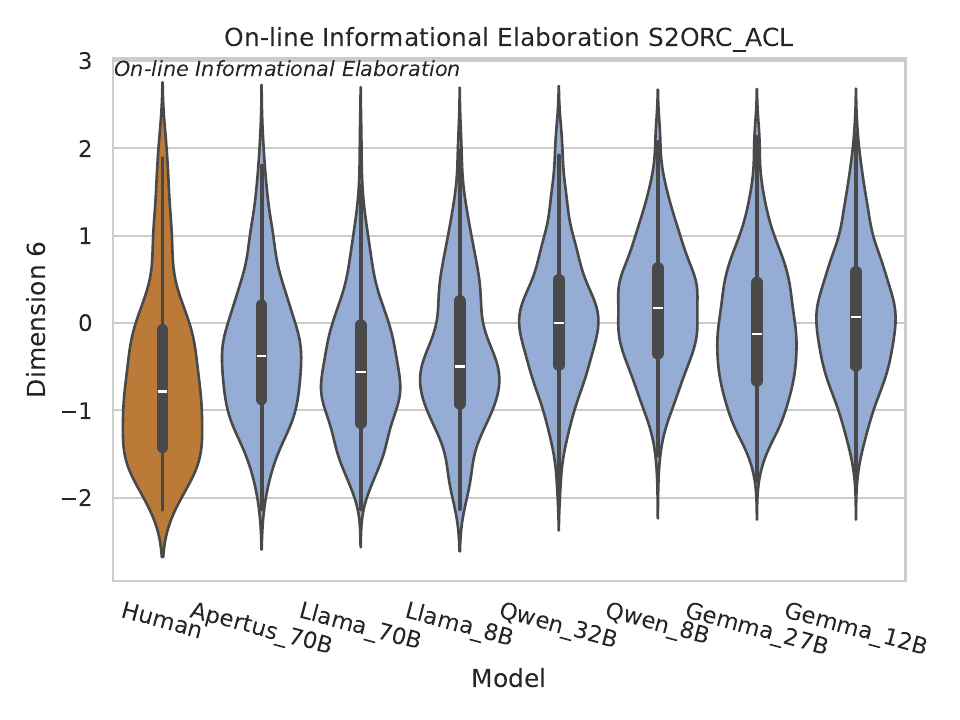}
  \caption{Dimension 6}
\end{subfigure}

\caption{Human and model distributions for Biber dimensions in the Zero-Shot setting (\textbf{S2ORC\_ACL}).}
\label{fig:app:biber-S2ORC_ACL}
\end{figure*}

\begin{figure*}[p]
\centering
\begin{subfigure}{0.49\linewidth}
  \includegraphics[width=\linewidth]{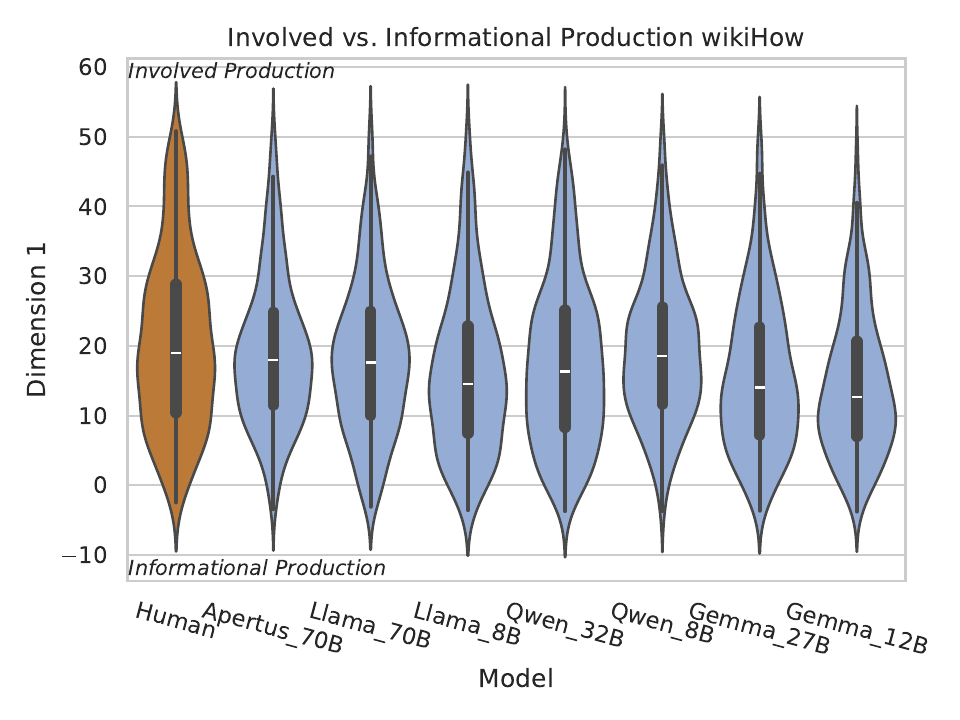}
  \caption{Dimension 1}
\end{subfigure}\hfill
\begin{subfigure}{0.49\linewidth}
  \includegraphics[width=\linewidth]{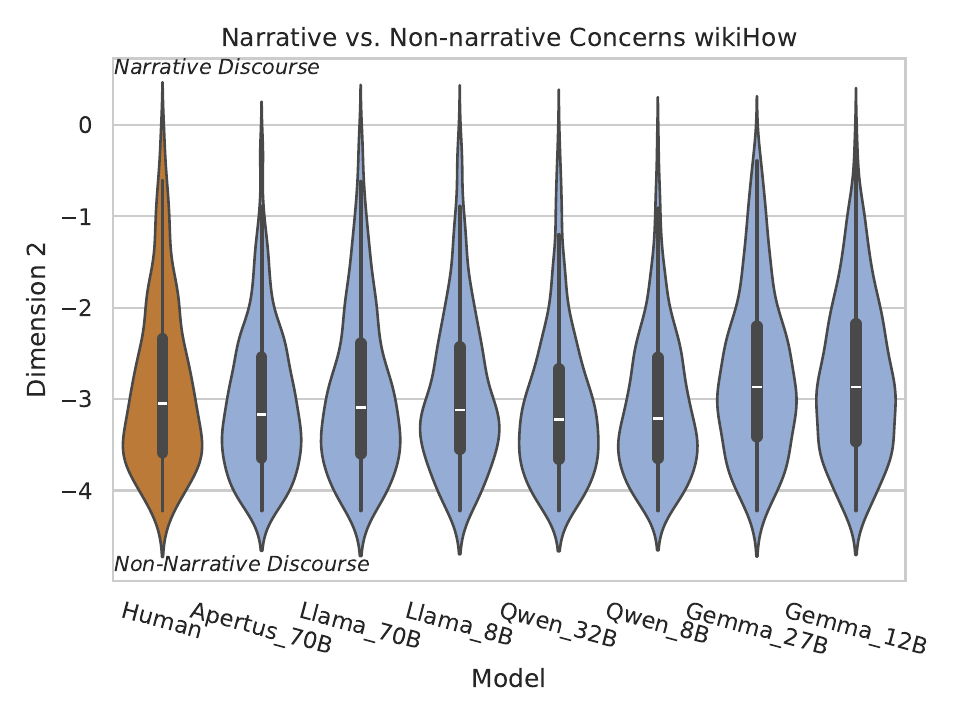}
  \caption{Dimension 2}
\end{subfigure}

\begin{subfigure}{0.49\linewidth}
  \includegraphics[width=\linewidth]{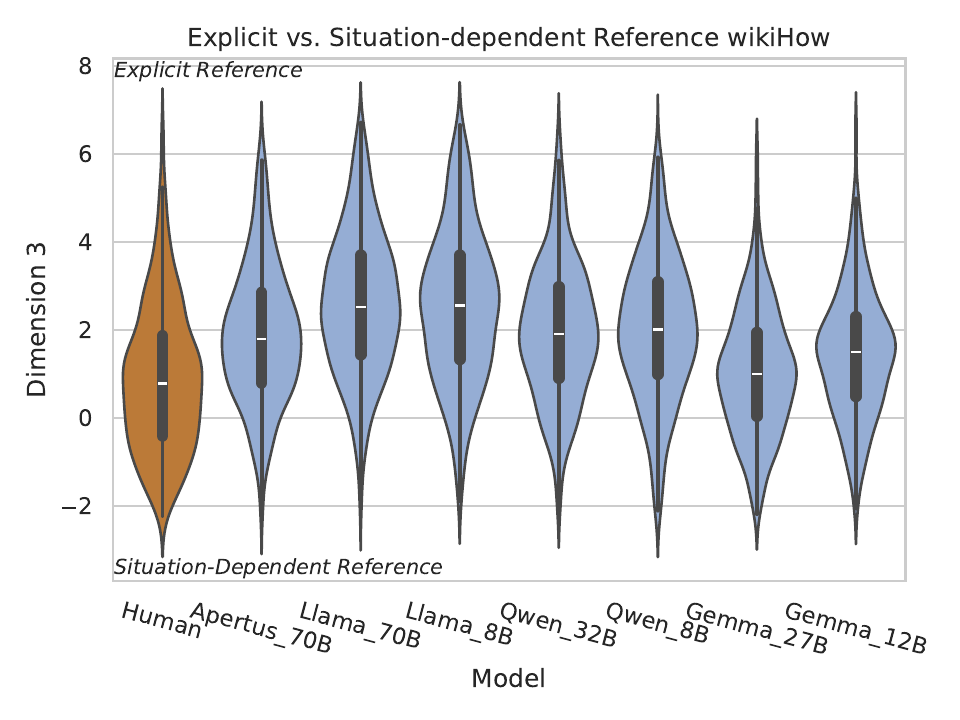}
  \caption{Dimension 3}
\end{subfigure}\hfill
\begin{subfigure}{0.49\linewidth}
  \includegraphics[width=\linewidth]{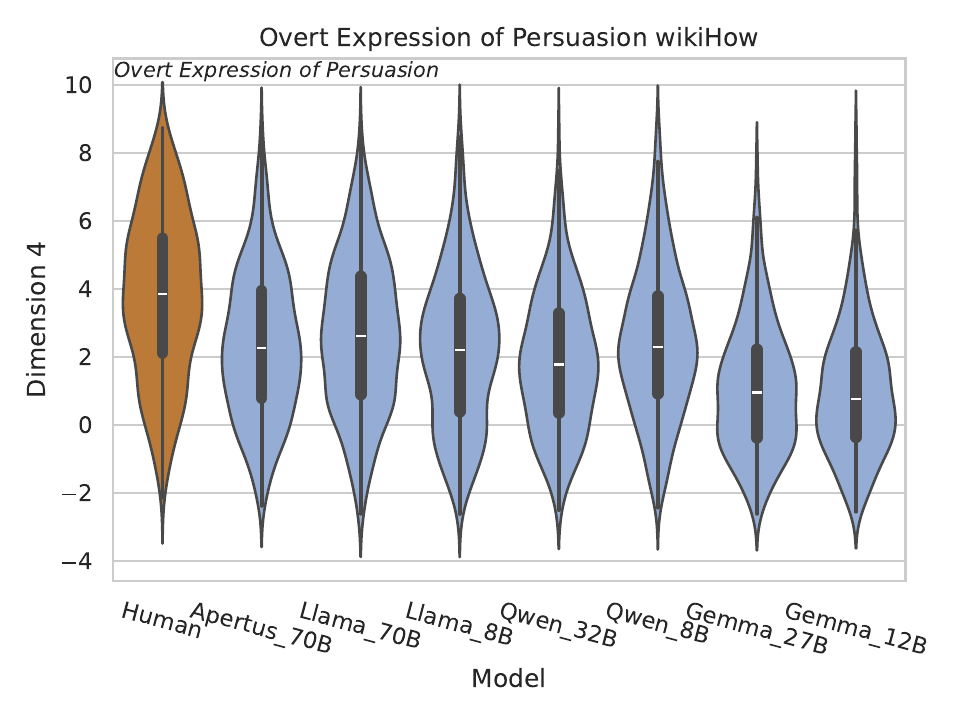}
  \caption{Dimension 4}
\end{subfigure}

\begin{subfigure}{0.49\linewidth}
  \includegraphics[width=\linewidth]{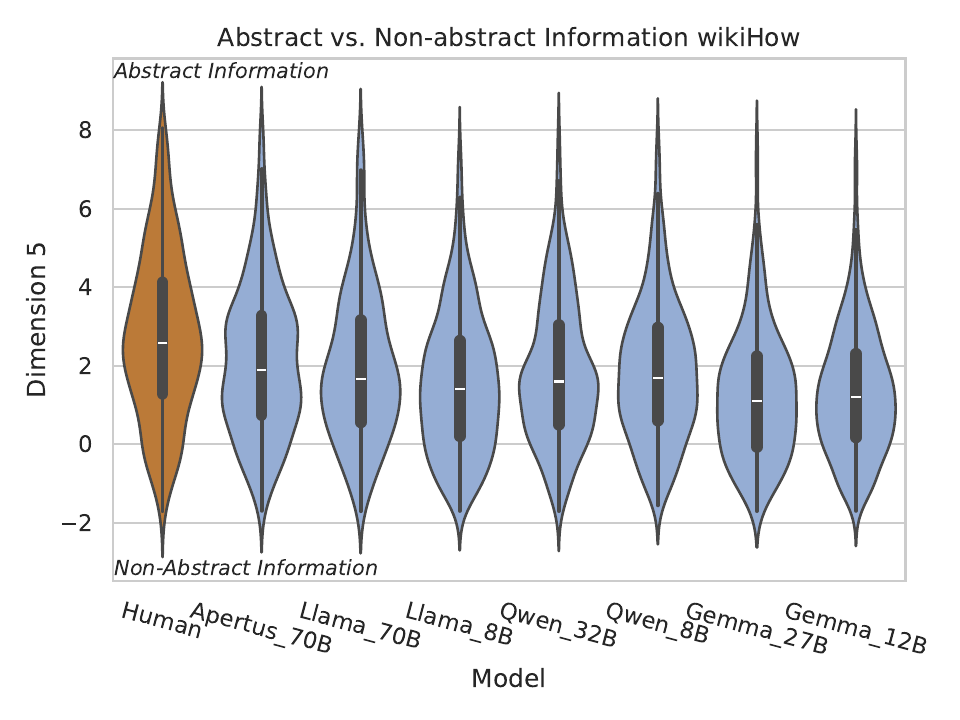}
  \caption{Dimension 5}
\end{subfigure}\hfill
\begin{subfigure}{0.49\linewidth}
  \includegraphics[width=\linewidth]{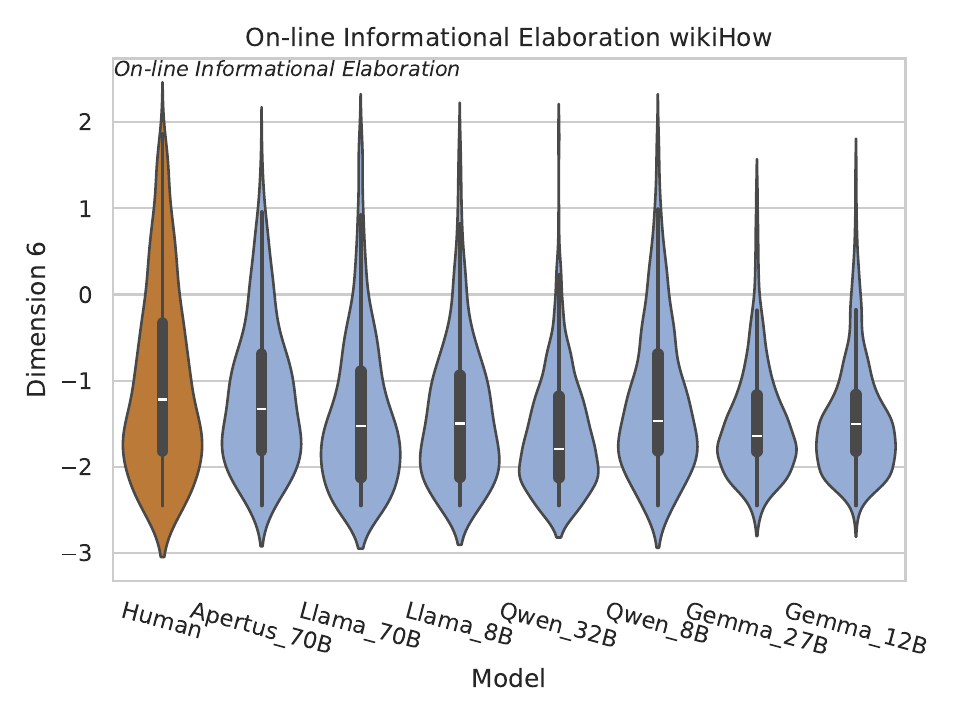}
  \caption{Dimension 6}
\end{subfigure}

\caption{Human and model distributions for Biber dimensions in the Zero-Shot setting (\textbf{wikiHow}).}
\label{fig:app:biber-wikiHow}
\end{figure*}

\begin{figure*}[p]
\centering
\begin{subfigure}{0.49\linewidth}
  \includegraphics[width=\linewidth]{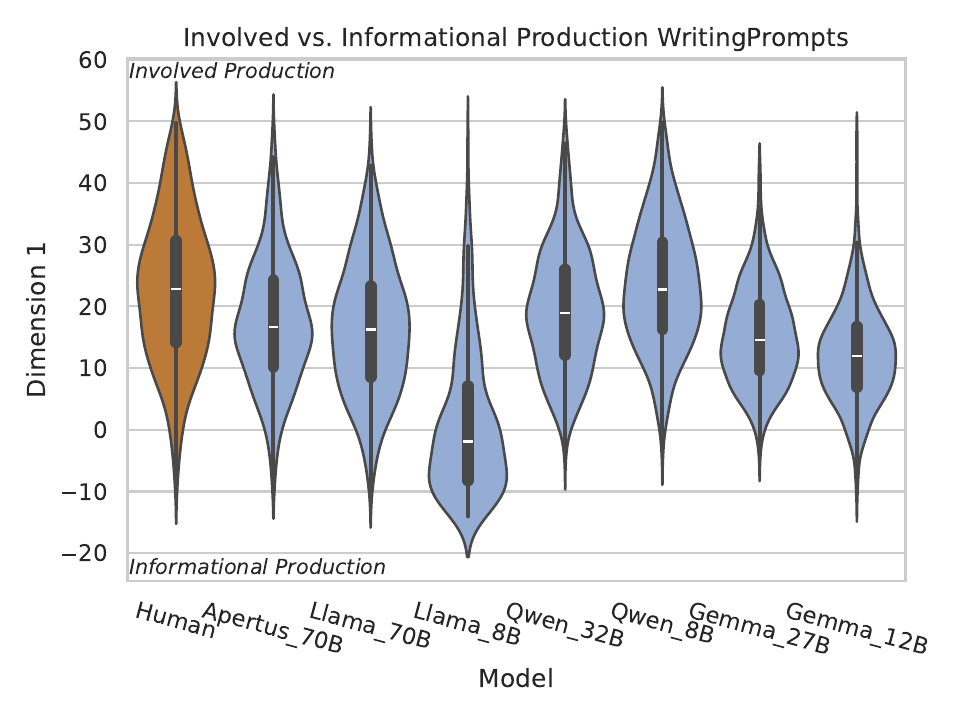}
  \caption{Dimension 1}
\end{subfigure}\hfill
\begin{subfigure}{0.49\linewidth}
  \includegraphics[width=\linewidth]{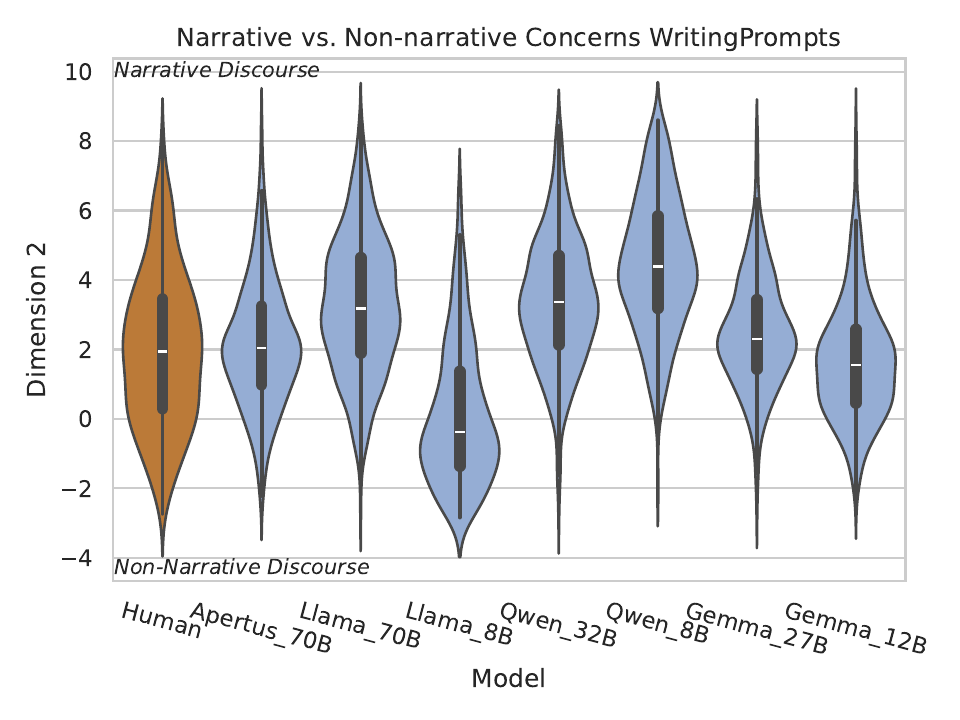}
  \caption{Dimension 2}
\end{subfigure}

\begin{subfigure}{0.49\linewidth}
  \includegraphics[width=\linewidth]{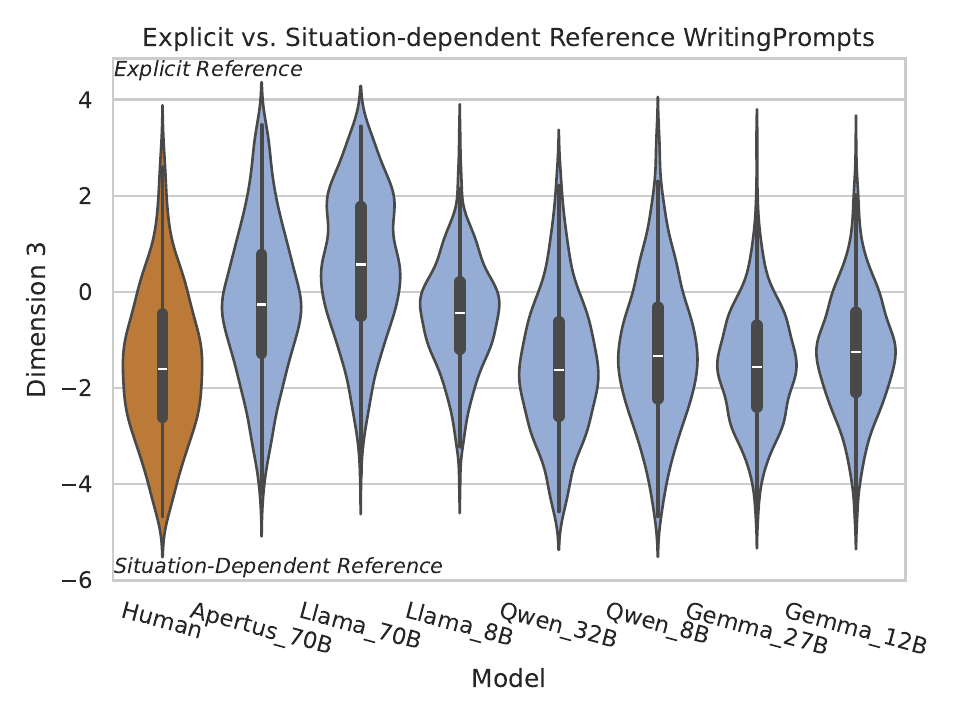}
  \caption{Dimension 3}
\end{subfigure}\hfill
\begin{subfigure}{0.49\linewidth}
  \includegraphics[width=\linewidth]{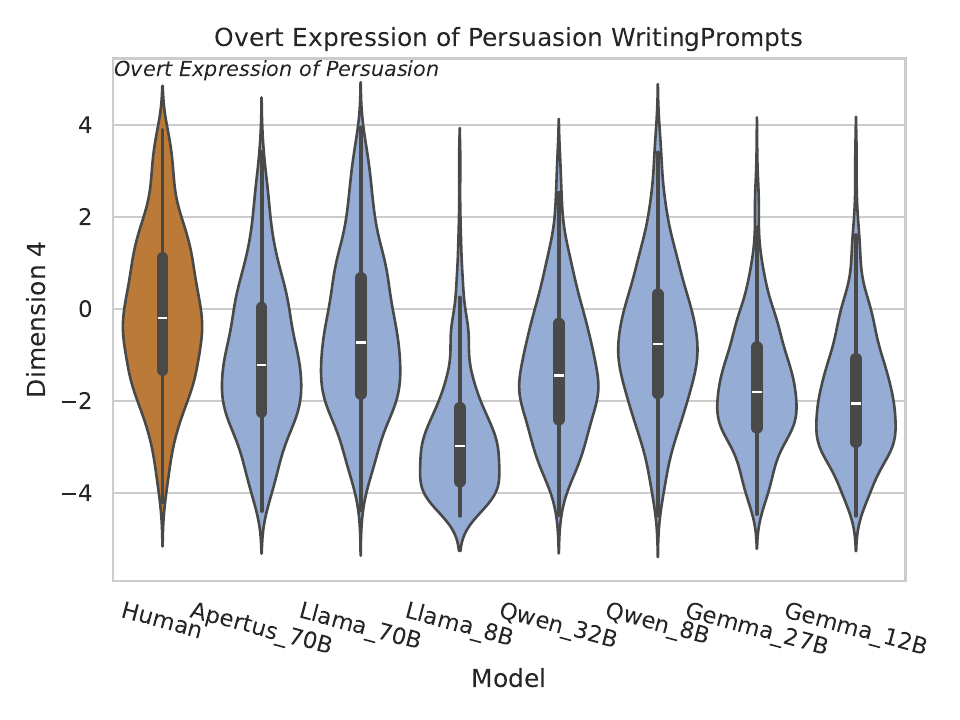}
  \caption{Dimension 4}
\end{subfigure}

\begin{subfigure}{0.49\linewidth}
  \includegraphics[width=\linewidth]{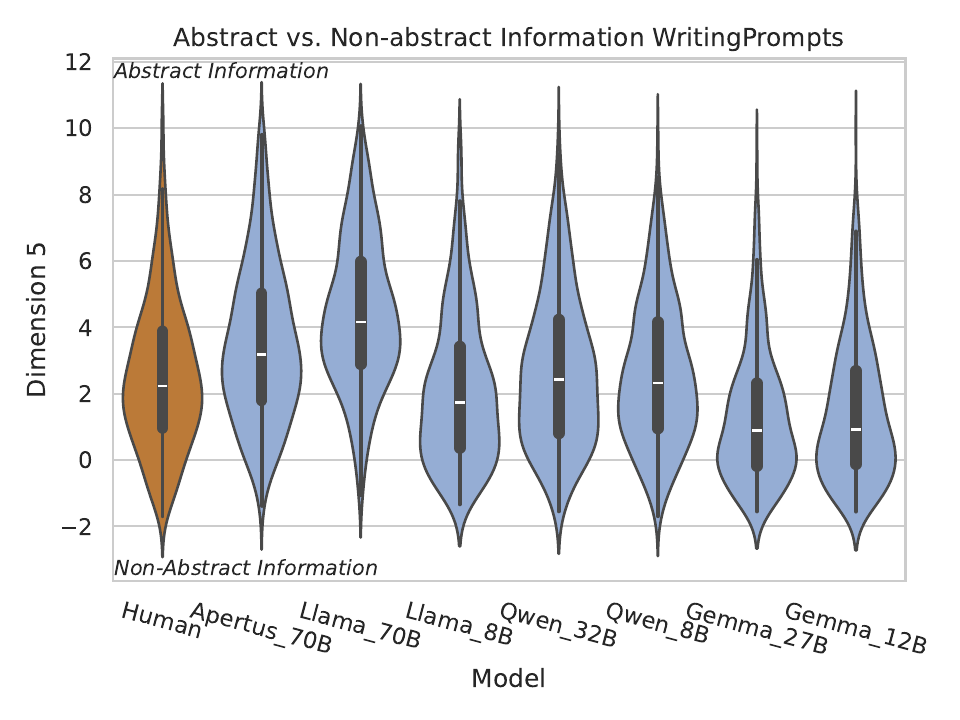}
  \caption{Dimension 5}
\end{subfigure}\hfill
\begin{subfigure}{0.49\linewidth}
  \includegraphics[width=\linewidth]{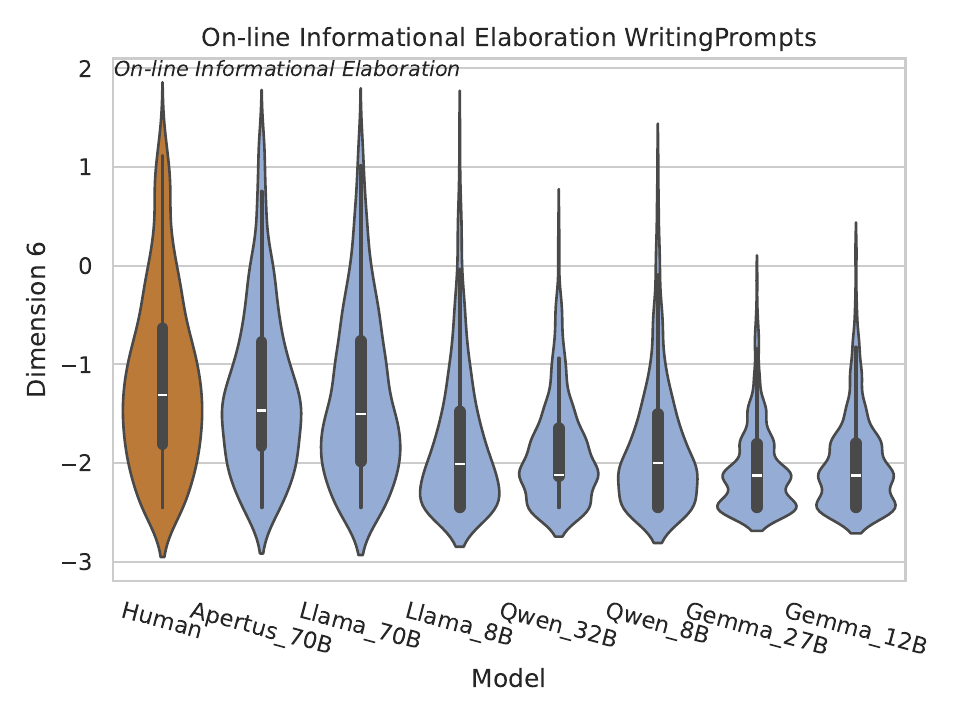}
  \caption{Dimension 6}
\end{subfigure}

\caption{Human and model distributions for Biber dimensions in the Zero-Shot setting (\textbf{WritingPrompts}).}
\label{fig:app:biber-WritingPrompts}
\end{figure*}

\begin{figure*}[p]
\centering
\begin{subfigure}{0.49\linewidth}
  \includegraphics[width=\linewidth]{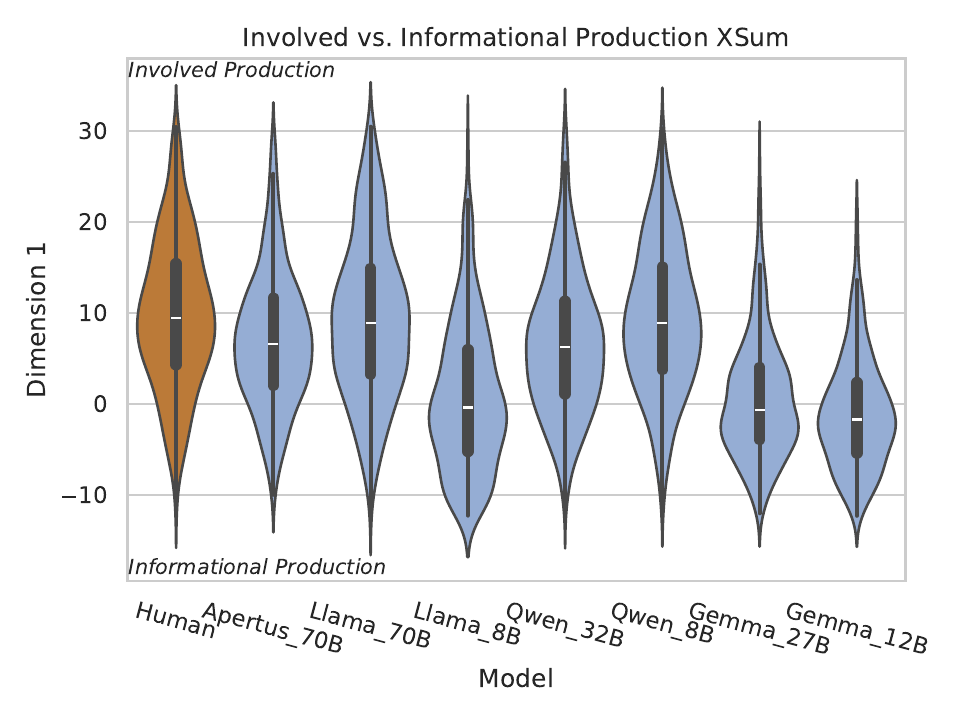}
  \caption{Dimension 1}
\end{subfigure}\hfill
\begin{subfigure}{0.49\linewidth}
  \includegraphics[width=\linewidth]{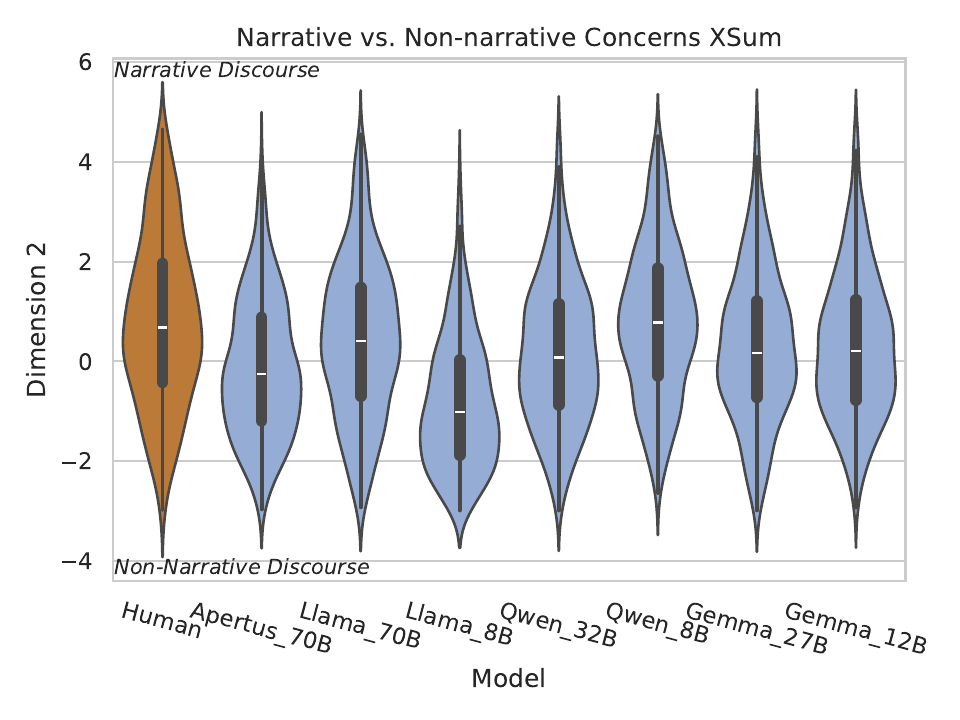}
  \caption{Dimension 2}
\end{subfigure}

\begin{subfigure}{0.49\linewidth}
  \includegraphics[width=\linewidth]{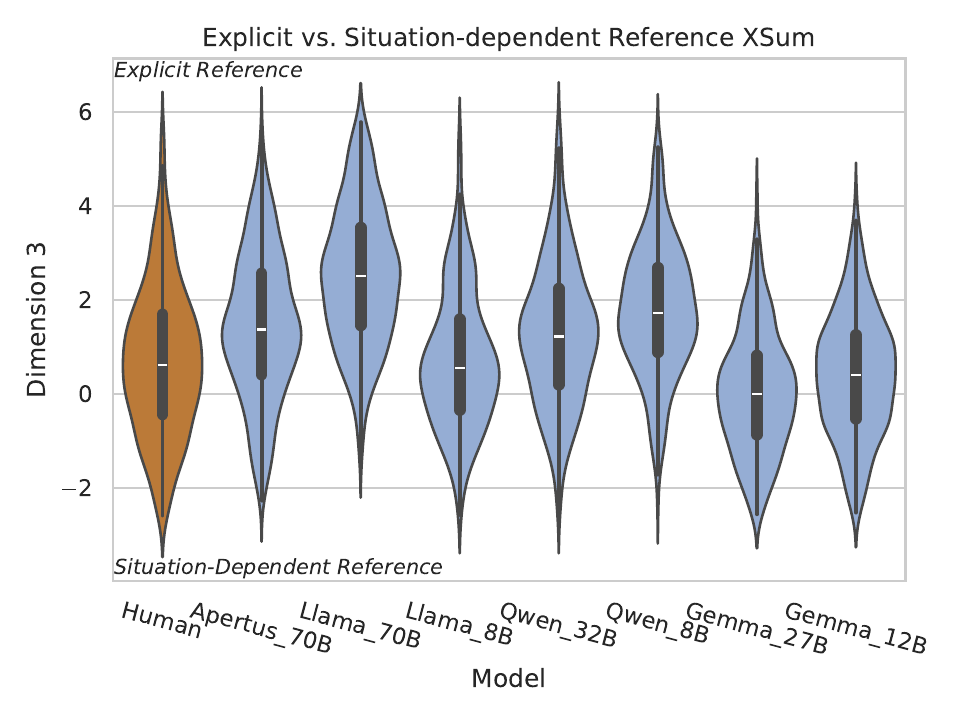}
  \caption{Dimension 3}
\end{subfigure}\hfill
\begin{subfigure}{0.49\linewidth}
  \includegraphics[width=\linewidth]{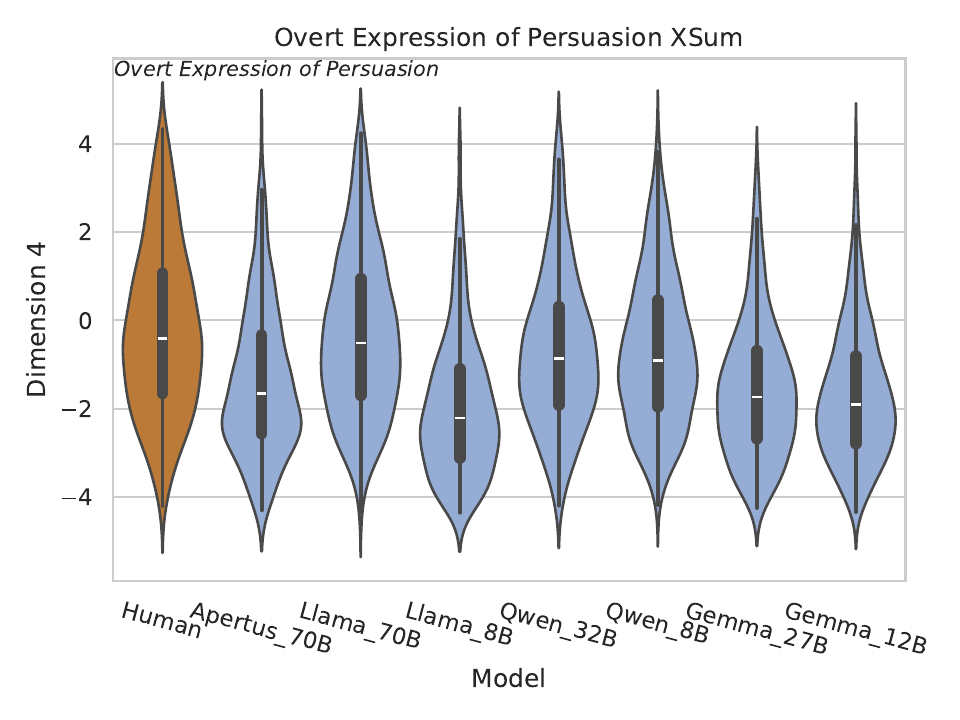}
  \caption{Dimension 4}
\end{subfigure}

\begin{subfigure}{0.49\linewidth}
  \includegraphics[width=\linewidth]{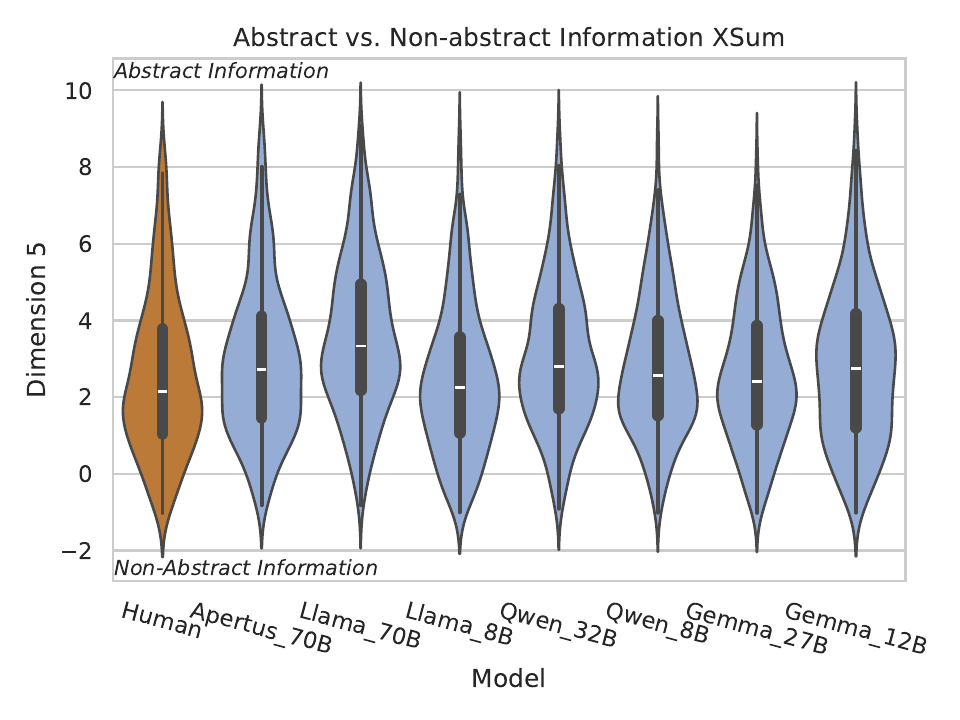}
  \caption{Dimension 5}
\end{subfigure}\hfill
\begin{subfigure}{0.49\linewidth}
  \includegraphics[width=\linewidth]{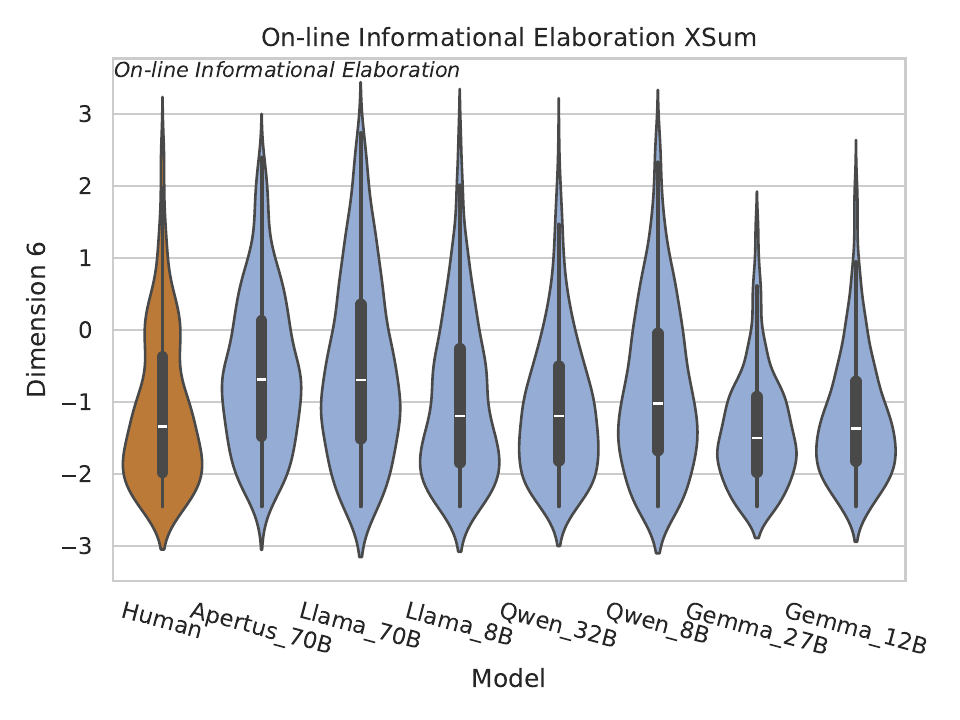}
  \caption{Dimension 6}
\end{subfigure}

\caption{Human and model distributions for Biber dimensions in the Zero-Shot setting (\textbf{XSum}).}
\label{fig:app:biber-XSum}
\end{figure*}

Figures \ref{fig:app:biber-bnc2014spoken}, \ref{fig:app:biber-S2ORC_ACL}, \ref{fig:app:biber-wikiHow}, \ref{fig:app:biber-WritingPrompts} and \ref{fig:app:biber-XSum} show the human and model distributions for all datasets and dimensions. 
\definecolor{nouncolor}{RGB}{65, 180, 211}
\definecolor{pastcolor}{RGB}{162, 55, 13}
\definecolor{attrcolor}{RGB}{0,120,110}
\definecolor{nomcolor}{RGB}{120,0,140}
\definecolor{prespartcolor}{RGB}{200,120,0}
\definecolor{pastpartcolor}{RGB}{100,120,0}

\NewDocumentCommand{\NOUN}{m}{\textcolor{nouncolor}{#1}}
\NewDocumentCommand{\PAST}{m}{\textcolor{pastcolor}{\textbf{#1}}}
\NewDocumentCommand{\ATTR}{m}{\textcolor{attrcolor}{#1}}
\NewDocumentCommand{\NOM}{m}{\textcolor{nomcolor}{\textit{#1}}}
\NewDocumentCommand{\PRESPART}{m}{\textcolor{prespartcolor}{#1}}
\NewDocumentCommand{\PASTPART}{m}{\textcolor{pastpartcolor}{#1}}

\begin{table*}[t]
\centering
\small
\setlength{\tabcolsep}{6pt}
\renewcommand{\arraystretch}{1.2}

\begin{tabular}{p{0.32\linewidth} p{0.32\linewidth} p{0.32\linewidth}}
\toprule

\multicolumn{3}{p{\textwidth}}{
\textbf{Writing Prompt:} ``So how did the Jeep end up in my rooftop office again?''
} \\
\midrule

\multicolumn{3}{p{\textwidth}}{\textbf{Human:} 
My \NOUN{roommate} \PAST{had} some \ATTR{leftover} \NOUN{pizza} in the \NOUN{fridge}. 
Me, being— 
“Have you been playing with \NOUN{rockets} again? \PAST{Was} it a \NOUN{prank}?” 
“No, \NOUN{officer}. This is a completely different \NOUN{chain} of \NOUN{events}.” 
“What does your \NOUN{roommate} having \NOUN{pizza} in the \NOUN{fridge} have anything to do with anything?” 
Jim \PAST{took} a \ATTR{deep} \NOUN{breath}. 
“Well, as any \NOUN{college} \NOUN{student} is at noon, I \PAST{was} hungry. 
So I \PAST{proceeded} to eat the \NOUN{pizza}, thinking it was mine and I \PAST{had} forgotten about it. 
I \PAST{finished} the \NOUN{pizza}, to realize the last \NOUN{time} I \PAST{had} eaten \NOUN{pizza} was at a \NOUN{party} three \NOUN{weeks} ago. 
It \PAST{wasn't} mine.” 
“\PAST{Could} you cut to the \NOUN{chase}, sir?”} \\
\midrule

\textbf{Gemma 12B} & \textbf{Qwen 32B} & \textbf{Llama 70B} \\
\midrule
So how \PAST{did} the \NOUN{Jeep} end up in my \NOUN{rooftop} \NOUN{office} again?

The \NOUN{question} \PAST{echoed} in the \ATTR{cavernous} \NOUN{space},
\PRESPART{bouncing} off the \NOUN{glass} \NOUN{walls} and
the surprisingly \ATTR{resilient} \NOUN{foliage}
\PRESPART{clinging} to the \NOUN{building}’s \NOUN{edge}.

It \PAST{wasn’t} a \ATTR{rhetorical} \NOUN{question},
despite the \ATTR{utter} \NOM{absurdity} of the \NOUN{situation}.

I genuinely \PAST{wanted} to know.

My \NOUN{brain}, normally a finely \ATTR{tuned} \NOUN{engine}
of \ATTR{legal} \NOUN{arguments} and \NOM{negotiations},
\PAST{felt} like a \ATTR{rusty} \NOUN{tricycle},
\PRESPART{spinning} uselessly in the face of this automotive anomaly.

It \PAST{was} a 1987 \NOUN{Jeep} \NOUN{Cherokee},
\NOUN{forest} \ATTR{green},
with the kind of \ATTR{faded} \NOUN{stickers}
that \PAST{hinted} at a \NOUN{life}
\PASTPART{lived} somewhere far removed
from the meticulously \ATTR{manicured} \NOUN{world}
of \NOUN{Meridian Tower}...&
I \PAST{muttered},
\PRESPART{rubbing} my \NOUN{temples}
as I \PAST{stared} at the \ATTR{rusted} \NOUN{hood}
of the \ATTR{old} \NOUN{Wrangler}
\PRESPART{dominating} the center of the \NOUN{space}.

I’d \PAST{asked} myself that same \NOUN{question}
dozens of \NOUN{times} over the \ATTR{past} \NOUN{year},
and honestly, I \PAST{had} as many \ATTR{conflicting} \NOUN{answers}.

But it \PAST{was} my \NOUN{office} now—well, technically my \NOUN{apartment}—
and I \PAST{was} beginning to feel like it \PAST{was} part of the \NOUN{furniture},
a slightly \ATTR{ambitious} \NOUN{coffee} \NOUN{table}
with four \NOUN{doors} and a \ATTR{stubby} \NOUN{antenna}.

I \PAST{sighed} and \PAST{hoisted} a \NOUN{duffel} \NOUN{bag}
onto one \NOUN{shoulder}
before \PRESPART{heading} down the \ATTR{spiral} \NOUN{staircase}
that \PAST{served} as the only \NOUN{access} to the \NOUN{rooftop}.

I \PAST{had} to remember to get more \NOUN{light} \NOUN{bulbs}...
&
So how did the \NOUN{Jeep} end up in my \NOUN{rooftop} \NOUN{office} again?

I \PAST{rubbed} my \NOUN{temples},
\PRESPART{trying} to ward off the \NOUN{headache} that \PAST{was} brewing.

It \PAST{wasn’t} the first \NOUN{time} something bizarre \PAST{had} happened in my \NOUN{life},
but I \PAST{was} starting to think I \PAST{was} losing my \NOUN{mind}.

One \NOUN{minute} I \PAST{was} sipping my \NOUN{morning} \NOUN{coffee}
and staring out at the \NOUN{city} \NOUN{skyline},
and the next, I \PAST{was} face to face with a \ATTR{bright blue} \NOUN{Jeep Wrangler},
\PASTPART{parked} smack in the middle of my \NOUN{office}.

I \PAST{looked} around,
\PRESPART{hoping} that maybe I had somehow missed the \NOM{construction} \NOUN{crew} that must have hauled it up here,
but the \NOUN{room} \PAST{was} quiet and still...
\\
\bottomrule
\end{tabular}

\caption{Qualitative comparison of human and LLM generations for narrative prompt number 295432 from the WritingPrompts dataset for the Zero-Shot setting. In the table \NOUN{Nouns}, \PAST{Past tense}, \ATTR{Attributive Adjectives}, \PRESPART{Present Participle Clauses}, \NOM{Nominalizations} and \PASTPART{Past Participle Clauses} are annotated.}
\label{tab:qualitative_example}
\end{table*}

\FloatBarrier


\begin{figure*}[p]
    \includegraphics[width=\textwidth]{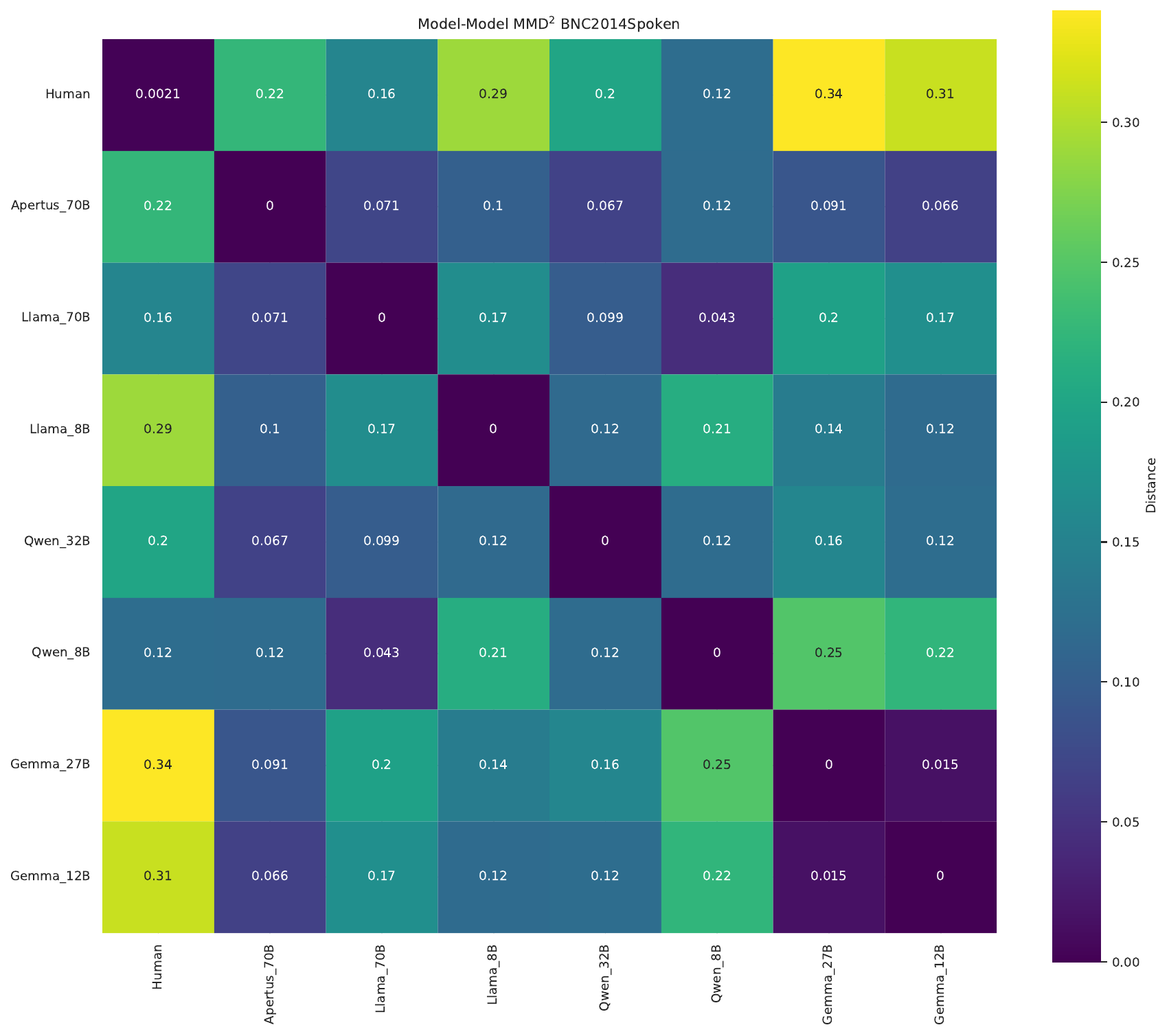}
    \caption{Observed MMD distance between different models for \textbf{BNC2014Spoken} in the Zero-Shot setting. The MMD between Human-Human is calculated as described in \ref{sec:MMD_stability}}
    \label{fig:Model_Model_MMD_BNC2014Spoken}
\end{figure*}

\begin{figure*}[p]
    \includegraphics[width=\textwidth]{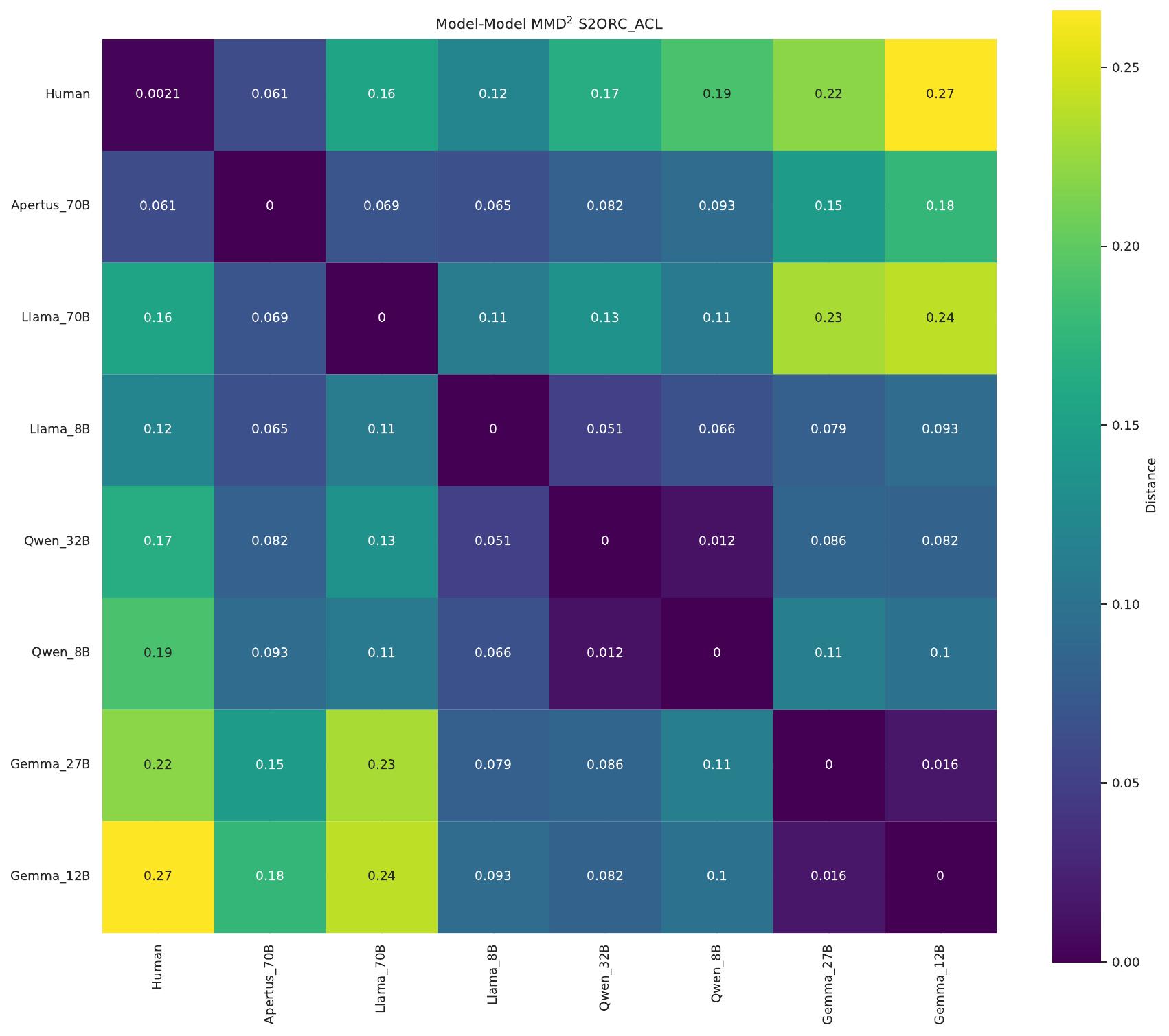}
    \caption{Observed MMD distance between different models for \textbf{S2ORC\_ACL} in the Zero-Shot setting. The MMD between Human-Human is calculated as described in \ref{sec:MMD_stability}}
    \label{fig:Model_Model_MMD_S2ORC_ACL}
\end{figure*}

\begin{figure*}[p]
    \includegraphics[width=\textwidth]{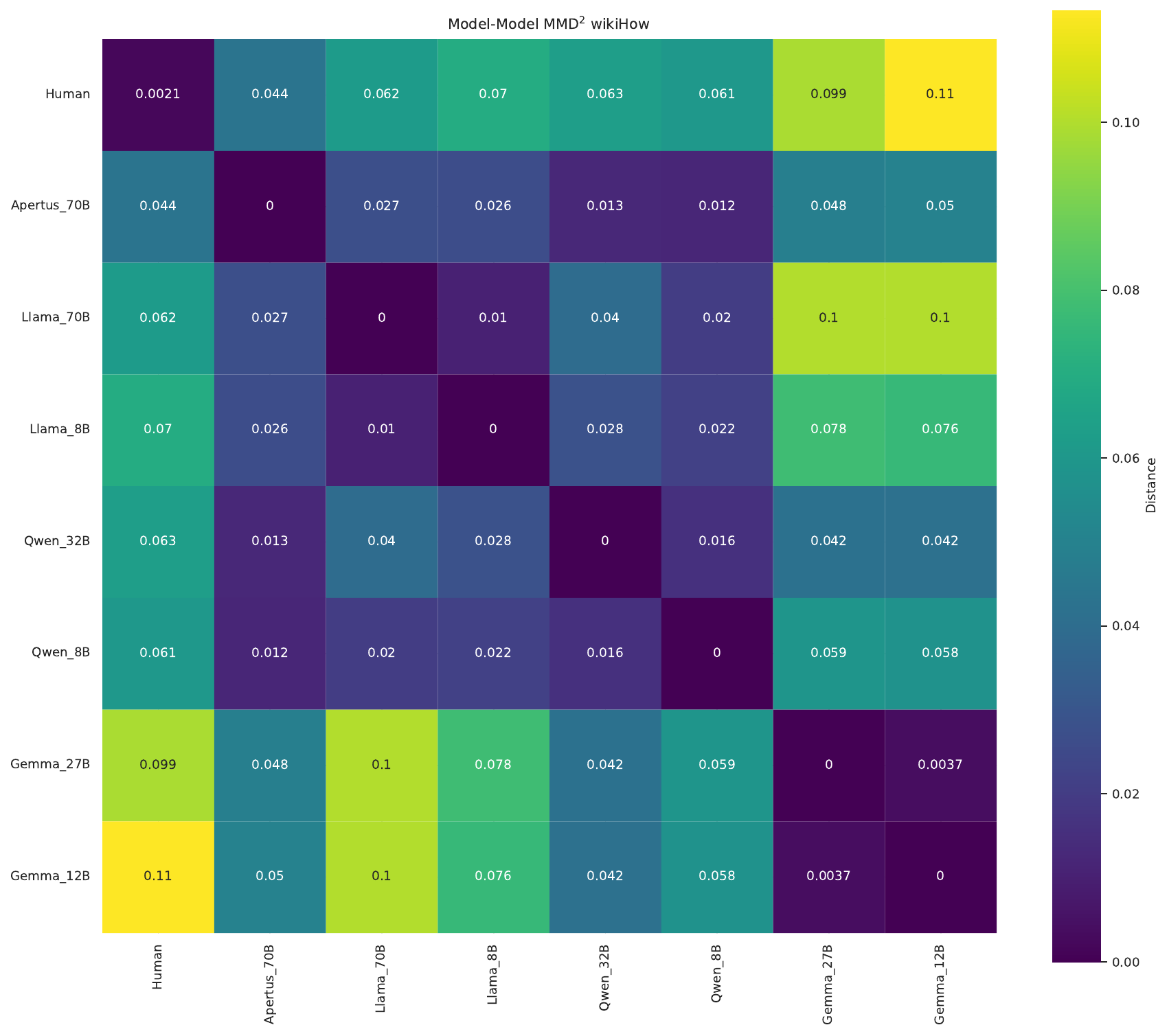}
    \caption{Observed MMD distance between different models for \textbf{wikiHow} in the Zero-Shot setting. The MMD between Human-Human is calculated as described in \ref{sec:MMD_stability}}
    \label{fig:Model_Model_MMD_wikiHow}
\end{figure*}

\begin{figure*}[p]
    \includegraphics[width=\textwidth]{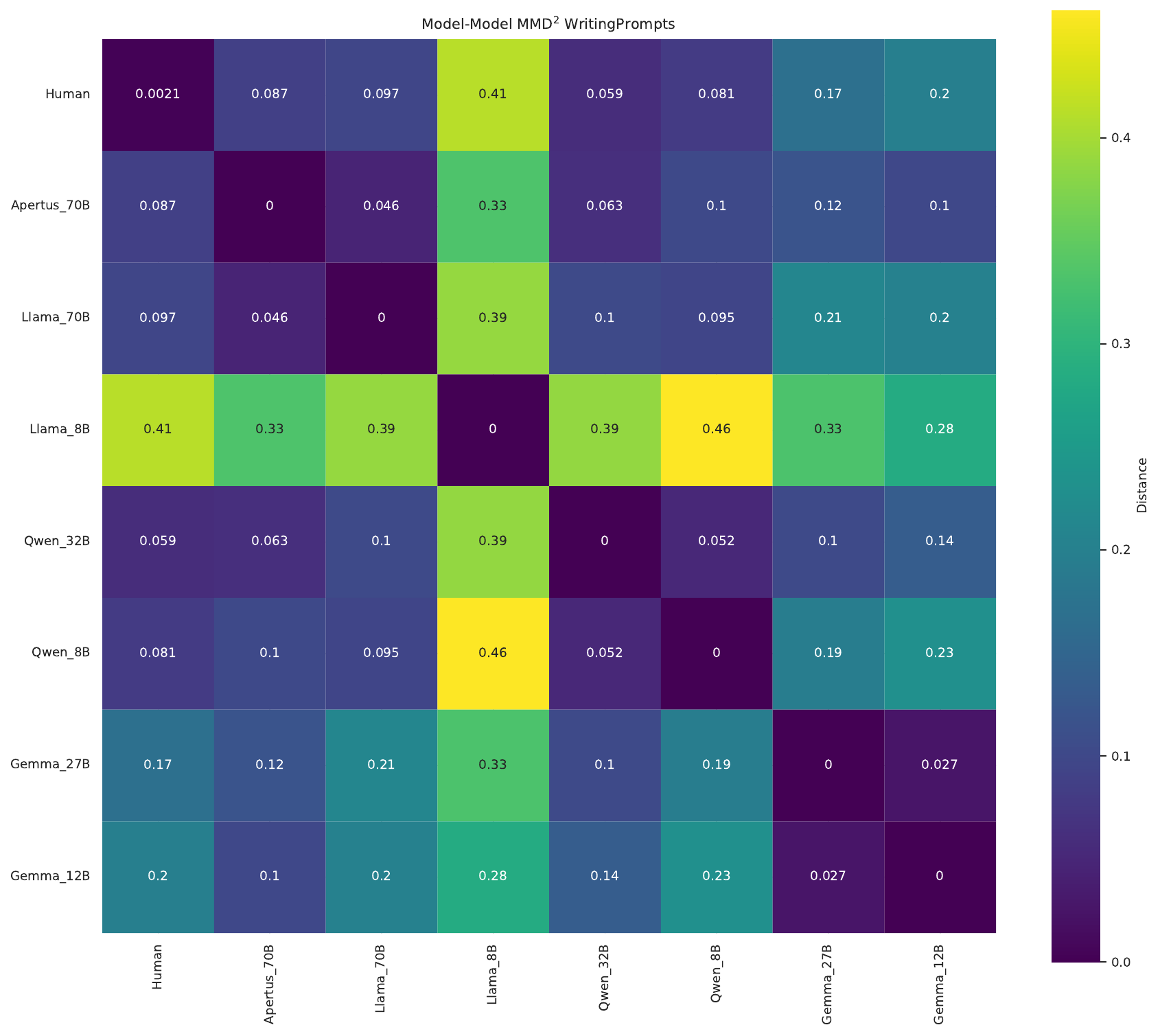}
    \caption{Observed MMD distance between different models for \textbf{WritingPrompts} in the Zero-Shot setting. The MMD between Human-Human is calculated as described in \ref{sec:MMD_stability}}
    \label{fig:Model_Model_MMD_WritingPrompts}
\end{figure*}
\begin{figure*}[p]
    \includegraphics[width=\textwidth]{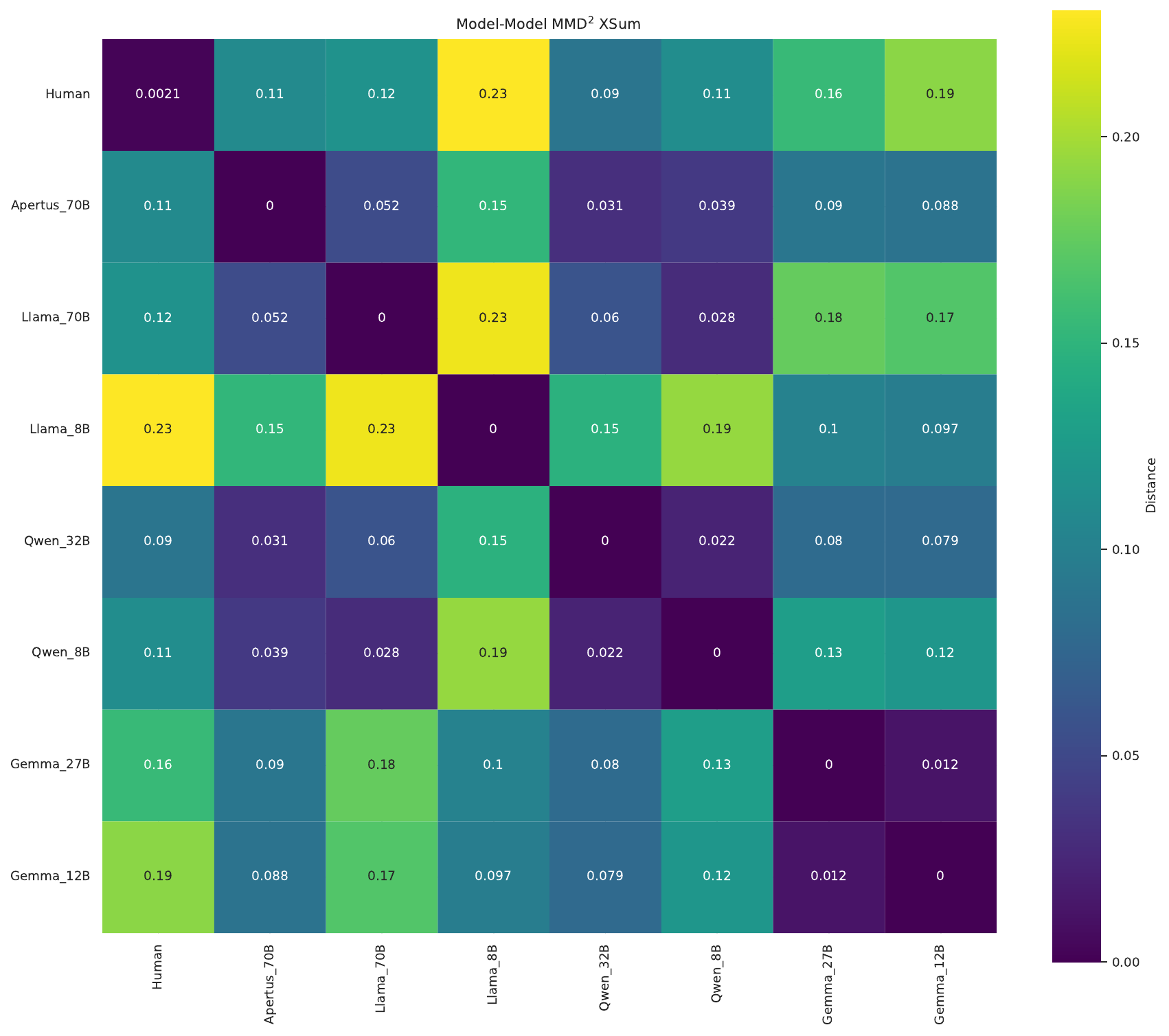}
    \caption{Observed MMD distance between different models for \textbf{XSum} in the Zero-Shot setting. The MMD between Human-Human is calculated as described in \ref{sec:MMD_stability}}
    \label{fig:Model_Model_MMD_XSum}
\end{figure*}

\onecolumn
\begin{table}[htbp]
\centering
\small 
\setlength{\tabcolsep}{4pt} 
\begin{tabular}{@{}l p{0.62\linewidth} r@{}}
\toprule
\textbf{Dimension} & \textbf{Feature} & \textbf{Loading} \\
\midrule

\multirow{24}{*}{{\textbf{Dimension 1}}} &
    private verbs & 0.96 \\
& THAT deletion & 0.91 \\
& contractions & 0.90 \\
& present tense verbs & 0.86 \\
& 2nd person pronouns & 0.86 \\
& DO as pro-verb & 0.82 \\
& analytic negation & 0.78 \\
& demonstrative pronouns & 0.76 \\
& general emphatics & 0.74 \\
& 1st person pronouns & 0.74 \\
& pronoun \textit{it} & 0.71 \\
& BE as main verb & 0.71 \\
& causative subordination & 0.66 \\
& discourse particles & 0.66 \\
& indefinite pronouns & 0.62 \\
& general hedges & 0.58 \\
& amplifiers & 0.56 \\
& sentence relatives & 0.55 \\
& WH questions & 0.52 \\
& possibility modals & 0.50 \\
& non-phrasal coordination & 0.48 \\
& WH clauses & 0.47 \\
& final prepositions & 0.43 \\
& nouns & -0.80 \\
& word length & -0.58 \\
& prepositions & -0.54 \\
& type/token ratio & -0.54 \\
& attributive adjectives & -0.47 \\

\midrule
\multirow{6}{*}{{\textbf{Dimension 2}}} &
    past tense verbs & 0.90 \\
& third person pronouns & 0.73 \\
& perfect aspect verbs & 0.48 \\
& public verbs & 0.43 \\
& synthetic negation & 0.40 \\
& present participial clauses & 0.39 \\

\midrule
\multirow{8}{*}{{\textbf{Dimension 3}}} &
    WH relative clauses on object positions & 0.63 \\
& pied piping constructions & 0.61 \\
& WH relative clauses on subject positions & 0.45 \\
& phrasal coordination & 0.36 \\
& nominalizations & 0.36 \\
& time adverbials & -0.60 \\
& place adverbials & -0.49 \\
& adverbs & -0.46 \\

\midrule
\multirow{6}{*}{{\textbf{Dimension 4}}} &
    infinitives & 0.76 \\
& prediction modals & 0.54 \\
& suasive verbs & 0.49 \\
& conditional subordination & 0.47 \\
& necessity modals & 0.46 \\
& split auxiliaries & 0.44 \\

\midrule
\multirow{6}{*}{{\textbf{Dimension 5}}} &
    conjuncts & 0.48 \\
& agentless passives & 0.43 \\
& past participial clauses & 0.42 \\
& BY-passives & 0.41 \\
& past participial WHIZ deletions & 0.40 \\
& other adverbial subordinators & 0.39 \\

\midrule
\multirow{4}{*}{{\textbf{Dimension 6}}} &
    THAT clauses as verb complements & 0.56 \\
& demonstratives & 0.55 \\
& That relative clause on object positions & 0.46 \\
& That clauses as adjective complements & 0.36 \\

\midrule
\multirow{1}{*}{{\textbf{Dimension 7}}} &
    SEEM / APPEAR & 0.35 \\
\bottomrule
\end{tabular}
\caption{Feature loadings from \citep[pp. ~89--90]{biber1988variation}}
\label{tab:biber1988-loadings}
\end{table}
\FloatBarrier
\twocolumn

\twocolumn

\FloatBarrier
\begin{figure*}
    \centering
    \includegraphics[width=\textwidth]{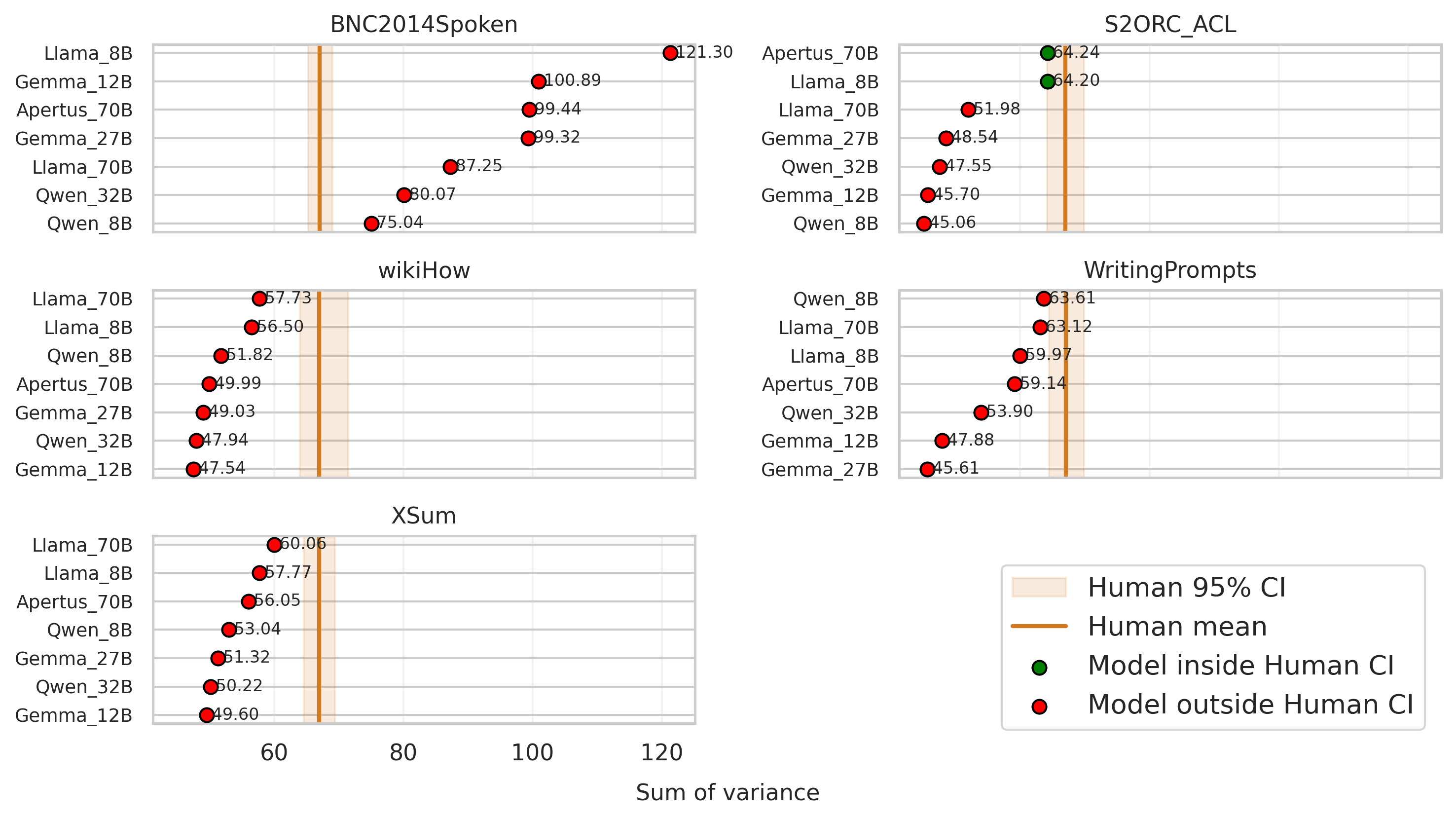}
    \caption{Sum of the variances of the 67 linguistic features after normalization on the corresponding full human distribution for each model in the Zero-Shot setting. The human mean is around 67, which is expected due to the normalization.}
    \label{fig:variances_models}
\end{figure*}


\end{document}